\def\ARXIVVERSION{}
\documentclass{article} 
\usepackage{preprint_style,times}

\usepackage{amsmath,amsfonts,bm}

\def\eqref#1{equation~\ref{#1}}

\def\1{\bm{1}}

\DeclareMathAlphabet{\mathsfit}{\encodingdefault}{\sfdefault}{m}{sl}
\SetMathAlphabet{\mathsfit}{bold}{\encodingdefault}{\sfdefault}{bx}{n}

\usepackage{amsmath}
\usepackage{amssymb}
\usepackage{booktabs}
\usepackage{array}
\usepackage{multirow}
\usepackage{float}
\usepackage{algorithm}
\usepackage{algorithmic}
\usepackage{graphicx}
\usepackage{placeins}
\usepackage{wrapfig}
\usepackage{hyperref}
\usepackage{url}

\title{Not All Variables Agree: Reliability-Aware Variable-Wise Gradient Surgery for \\Multivariate Time-Series Forecasting}

\author{%
  Jinwoo Park \\
  Department of Industrial Engineering\\
  Seoul National University\\
  Seoul, Republic of Korea \\
  \texttt{jinwoo\_park@snu.ac.kr} \\
  \And
  Hyeongwon Kang \\
  Department of Industrial \& Management Engineering\\
  Korea University\\
  Seoul, Republic of Korea \\
  \texttt{hyeongwon\_kang@korea.ac.kr} \\
  \And
  Pilsung Kang\thanks{Corresponding author} \\
  Department of Industrial Engineering\\
  Seoul National University\\
  Seoul, Republic of Korea \\
  \texttt{pilsung\_kang@snu.ac.kr}
}

\preprintfinalcopy
\begin{document}

\maketitle
\fancyhead[L]{Preprint}

\begin{abstract}
In data-driven training, multivariate time-series forecasting is usually optimized with a scalar loss averaged over samples, variables, and horizons. This averaging is convenient, but the optimizer sees only the aggregated gradient, which does not reveal whether the variable-wise contributions align or oppose one another. To quantify how often this disagreement arises, we measure the variable-wise gradients directly and find that \(30.6\%\) of their pairwise cosine similarities are negative on average across seven datasets. However, conflict and harm are not the same thing. Under shared training \(35\) of the \(64\) variables do worse than a full-input single-target oracle, and the harmed fraction is not reliably predicted by how often gradients conflict. We propose Per-Variable Surgery (PV-Surgery), an optimizer-side training strategy for backbones with cache-compatible layers. One backward pass builds variable-wise gradient proxies from output-side signals and keeps the pointwise forecasting loss. Reliability-aware selection targets layers whose proxy sums closely approximate their shared-gradient slices. Conditional pooling forms anchor and conflict pools without dropping variables. Common-direction surgery aligns variable or pooled gradients with their normalized mean and restores input norms to avoid reweighting. In experiments across five backbones, seven datasets, and four horizons, PV-Surgery lowers MSE by \(3.61\%\) and MAE by \(2.93\%\) on average. For multivariate forecasting, this indicates that the variable-wise structure hidden by mean-loss training is a usable optimization signal.
\end{abstract}

\section{Introduction}
\label{sec:introduction}

Multivariate time-series forecasting predicts the future values of multiple variables from a shared history of all variables \citep{lai2018modeling,lim2021time}. Forecasting architectures differ widely, but most are trained with a similar mean-loss objective. Prediction errors are averaged over samples, variables, and horizons to form a scalar loss, which is optimized through backpropagation \citep{nie2023time,liu2024itransformer,wang2024timexer}. This objective is convenient and model-agnostic, but it also collapses heterogeneous variable-wise learning signals into one shared update.

The three averages behind this scalarization do not play the same role. Averaging over samples estimates an expectation over the data distribution, whereas averaging over variables and horizons assumes that heterogeneous targets and future steps can be merged into one optimization signal without losing important structure. In practice, however, variables differ in scale, noise, predictability, dynamics, and cross-variable dependency \citep{lai2018modeling,liu2024itransformer,nochumsohn2025mtlinear}. Consequently, an update that helps one variable can be neutral or harmful for another, but the mean-loss gradient does not reveal which case holds \citep{nochumsohn2025mtlinear,liu2021conflict}.

More concretely, let \(L_d(\theta)\) denote the loss induced by variable \(d\), averaged over samples and horizons. Standard training minimizes \(L_{\mathrm{mean}}(\theta)=D^{-1}\sum_{d=1}^{D}L_d(\theta)\) and updates parameters with \(g_{\mathrm{mean}}=\nabla_{\theta}L_{\mathrm{mean}}(\theta)=D^{-1}\sum_{d=1}^{D}g_d\), where \(g_d=\nabla_{\theta}L_d(\theta)\). This aggregation is many-to-one, so different sets of variable-wise gradients can produce the same update. For \(D=2\), \(g_1=u+v\) and \(g_2=u-v\) produce \(g_{\mathrm{mean}}=u\), hiding the disagreement component \(v\). When \(\|v\|_2>\|u\|_2\), the inner product \(\langle g_1,g_2\rangle=\|u\|_2^2-\|v\|_2^2\) turns negative, so an update along \(-g_1\) reduces \(L_1\) but increases \(L_2\) to first order. Moreover, under any shared update \(\theta\leftarrow\theta-\eta g_{\mathrm{shared}}\) with step size \(\eta>0\), the first-order change of \(L_d\) is \(-\eta\langle g_d,g_{\mathrm{shared}}\rangle\), whose sign cannot be checked from the aggregate alone. The problem is therefore not that mean loss is invalid, but that its single gradient hides the variable-wise directions needed to distinguish disagreement from actual harm.

\begin{figure}[t]
\centering
\setlength{\tabcolsep}{1.5pt}
\begin{tabular}{@{}c@{\hspace{0.02\linewidth}}c@{}}
\includegraphics[height=0.245\linewidth]{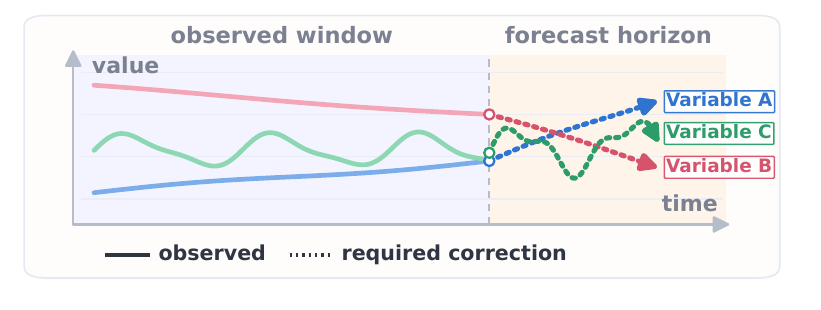}
&
\includegraphics[height=0.245\linewidth]{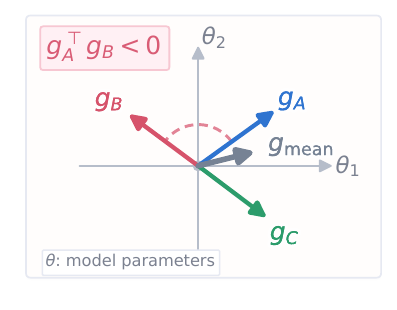}
\\[-1mm]
{\small (a) Variable-specific future corrections.}
&
{\small (b) Mean-loss gradient averaging.}
\end{tabular}
\caption{Conceptual illustration of variable-wise gradient conflict. (a) Different variables can require corrections in different directions over the same forecast horizon. (b) These corrections translate into variable-wise gradients that disagree in parameter space, while the mean-loss update shown here coincides with none of them.}
\label{fig:conceptual_conflict}
\end{figure}

Figure~\ref{fig:conceptual_conflict}(a) shows three variables requiring corrections in different directions over the same horizon, and Figure~\ref{fig:conceptual_conflict}(b) shows their parameter-space gradients, two of which oppose each other while their mean coincides with none of them. We call this \emph{variable-wise gradient conflict}, following the view that a negative inner product encodes incompatible descent directions \citep{yu2020gradient,nochumsohn2025mtlinear}. A natural reaction is to remove every conflict, as in gradient surgery \citep{yu2020gradient}. In multivariate forecasting, our diagnosis suggests otherwise. Conflicts are common but not uniformly harmful, so intervention should be selective, variable-aware, and anchored to the mean-loss update.

In this paper, we propose PV-Surgery, a reliability-aware optimizer-side framework. At each training step, one backward pass builds variable-wise gradient proxies from output-side signals. PV-Surgery then selects surgery layers based on the normalized error between each proxy sum and its matching reference-gradient slice, conditionally forms anchor and conflict pools without dropping variables, and aligns variable or pooled gradients with the normalized mean of their input directions before restoring their norms to avoid reweighting. PV-Surgery requires no changes to cache-compatible forecasting backbones or per-variable losses. We make the following three contributions:

\begin{itemize}
\item We analyze variable-wise conflict under standard mean-loss training using exact gradients. Although \(30.6\%\) of pairwise cosines are negative on average, conflict frequency is weakly related to harm. Shared training underperforms a full-input single-target oracle for \(35\) of \(64\) variables, while mean alignment does not reliably identify helpful partners.
\item We propose PV-Surgery, which reconstructs variable-wise gradient proxies in one backward pass, targets reconstruction-consistent layers, pools without dropping variables, and corrects directions while restoring surgery-input norms over the selected subspace.
\item We conduct experiments across five backbones, seven datasets, and four horizons, showing that PV-Surgery lowers MSE by \(3.61\%\) and MAE by \(2.93\%\) on average and that variable-wise structure hidden by mean-loss training is a usable optimization signal in these settings.
\end{itemize}

\section{Related Work}
\label{sec:related_work}

\subsection{Multivariate Time-Series Forecasting}
Multivariate forecasting has largely advanced through backbone design, spanning efficient attention \citep{zhou2021informer}, patch tokens \citep{nie2023time}, variate tokens \citep{liu2024itransformer}, exogenous cross-attention \citep{wang2024timexer}, series decomposition \citep{wu2021autoformer}, multi-scale convolution \citep{wang2023micn}, sample convolution \citep{liu2022scinet}, and linear baselines \citep{zeng2023transformers}. These approaches differ in architecture but retain the same mean-loss objective, whereas we intervene in the update it produces. MTLinear is the closest prior work, connecting multivariate forecasting to multi-task learning through variate gradient angle, correlation-based grouping, and gradient scaling for linear models \citep{nochumsohn2025mtlinear}. MTLinear fixes variable groups in advance and rescales gradient magnitudes, whereas we retain all variables, pool their gradients by directional compatibility at each step, and correct directions while restoring input norms across backbones.

\subsection{Forecasting Objectives and Variable-Wise Learning}
Forecasting objectives have expanded beyond point-wise MSE and MAE to include shape and time distortion losses \citep{leGuen2019dilate}, frequency-domain losses \citep{wang2024fredf}, transformation-invariant criteria \citep{lee2024tildeq}, patch-wise structural losses \citep{kudrat2025psloss}, and selective timestep masking \citep{fu2025selective}. These objectives change which errors are emphasized during training, but they still aggregate per-variable errors into one scalar. Reweighting the loss therefore does not reveal whether the corresponding gradients agree, cancel, or interfere.

\subsection{Gradient Conflict and Multi-Task Optimization}
Gradient conflict is studied in multi-task learning, where methods balance losses \citep{chen2018gradnorm}, seek Pareto-stationary descent directions \citep{sener2018multi}, project conflicting task gradients \citep{yu2020gradient}, maximize the worst-case improvement \citep{liu2021conflict}, or aggregate task gradients by bargaining \citep{navon2022multi}. Multivariate forecasting exposes its objectives differently because variables are components scalarized inside one loss rather than separate tasks, a setting that MTLinear connects to multi-task learning \citep{nochumsohn2025mtlinear}. We therefore construct variable-wise gradient proxies for these operators rather than applying them to the aggregate.

\section{Diagnosing Variable-Wise Gradient Conflict}
\label{sec:diagnosis}
\FloatBarrier

We analyze the frequency of variable-wise gradient conflict under standard multivariate forecasting training and its relation to the fraction of variables harmed under shared training. We also ask whether mean pairwise alignment reliably identifies helpful training partners. Together, these analyses assess whether pairwise cosine statistics provide a sound basis for deciding when to intervene.

\paragraph{Diagnostic setup.}
We train iTransformer with the standard mean-loss objective on all seven benchmarks, at prediction length 96 for the four ETT datasets, Weather, and Exchange, and at 24 for ILI. At analysis checkpoints we decompose the scalar objective into variable-induced losses \(L_d\), set \(g_d^{(t)}=\nabla_\theta L_d(\theta_t)\), and form the pairwise cosine matrix over nonzero gradients,
\begin{equation}
\label{eq:pairwise_cosine}
    C_{ij}^{(t)} =
    \frac{\langle g_i^{(t)}, g_j^{(t)} \rangle}
    {\|g_i^{(t)}\|_2 \|g_j^{(t)}\|_2}.
\end{equation}
A pair is conflicting at checkpoint \(t\) when \(C_{ij}^{(t)} < 0\). Checkpoint summaries discard the first half of recorded checkpoints as warmup because early updates are unstable, but this filter does not apply to separately trained oracle or partner-group results. The diagnostics expose variable-wise structure hidden by the baseline objective and are not used to tune PV-Surgery.

\begin{figure}[H]
\centering
\setlength{\tabcolsep}{1.5pt}
\begin{tabular}{@{}c@{\hspace{0.01\linewidth}}c@{\hspace{0.01\linewidth}}c@{}}
\includegraphics[width=0.360\linewidth]{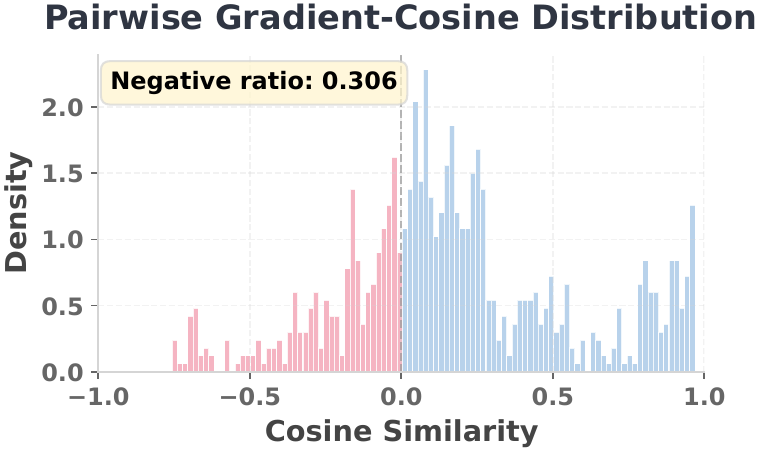}
&
\includegraphics[width=0.295\linewidth]{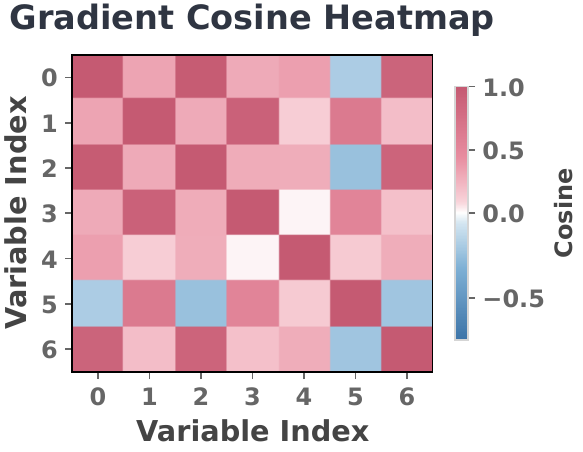}
&
\includegraphics[width=0.292\linewidth]{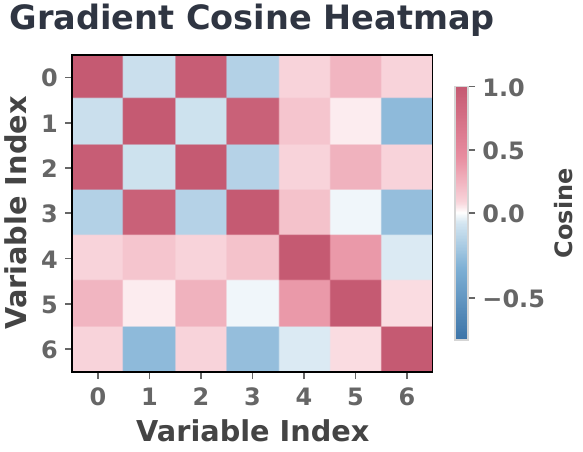}
\\[-1mm]
{\small (a) Post-warmup cosine distribution.}
&
{\small (b) Early checkpoint.}
&
{\small (c) Late checkpoint.}
\end{tabular}
\caption{A representative ETTh1 run from the variable-wise gradient-cosine diagnostic. (a) Distribution of the post-warmup pairwise cosines, of which \(30.6\%\) fall below zero for this run. (b) An early checkpoint, which lies inside the discarded warmup half and is shown only to expose the change over training. (c) A late checkpoint. Both heatmaps show that conflict is structured across variable pairs rather than being uniform, so one summary number would hide which pairs disagree.}
\label{fig:gradient_conflict_structure}
\end{figure}

\subsection{Variable-wise conflicts are frequent}

Across the seven datasets, \(30.6\%\) of the pairwise cosine similarities are negative on average, with per-dataset results in Appendix~\ref{app:gradient_conflict_all_datasets} and the ETTh1 case shown in Figure~\ref{fig:gradient_conflict_structure}. On the six datasets other than ILI, most post-warmup checkpoints contain at least one conflicting pair. The disagreement is also intermittent, since no pair is negative at every post-warmup checkpoint. Therefore, the standard scalar objective hides a substantial amount of variable-wise disagreement. However, this does not yet justify gradient surgery, since a negative cosine says only that two descent directions disagree at one checkpoint, not which variable is harmed or whether sharing is worse than separate training.

\subsection{Conflict and harm are not the same object}

To separate disagreement from harm, we compare the shared model with full-input single-target oracles, each optimized and evaluated on one target. We define the oracle gap as \mbox{\(\Delta_d^{\mathrm{oracle}}=\mathrm{MSE}_{\mathrm{shared},d}-\mathrm{MSE}_{\mathrm{oracle},d}\)}. A positive \(\Delta_d^{\mathrm{oracle}}\) means variable \(d\) is better predicted by its target-specific oracle. Across the seven datasets \(35\) of the \(64\) variables have a positive oracle gap, and Figure~\ref{fig:harm_subset_diagnostics}(a) reports the fraction for every dataset. At the dataset level, the negative-cosine ratio and the harmed fraction correlate at \(-0.29\) (\(p=0.53\)), so a larger conflict rate does not reliably coincide with a larger harmed fraction. ILI has the lowest conflict rate, whereas Weather has the highest harmed fraction (Table~\ref{tab:dataset_diagnostics}). A method that removes every negative cosine similarity would therefore risk destroying useful sharing for variables that improve under the shared model.

\subsection{Alignment alone does not reliably identify helpful partners}
\label{sec:partner_selection}

\begin{figure}[t]
\centering
\setlength{\tabcolsep}{1.5pt}
\begin{tabular}{@{}c@{\hspace{0.02\linewidth}}c@{}}
\includegraphics[width=0.480\linewidth]{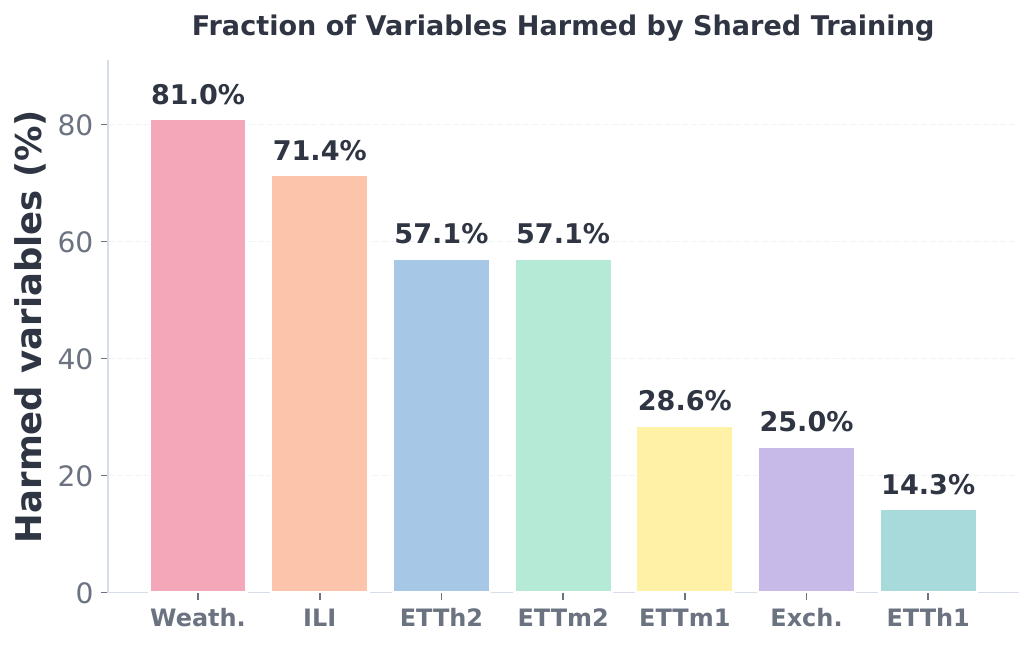}
&
\includegraphics[width=0.480\linewidth]{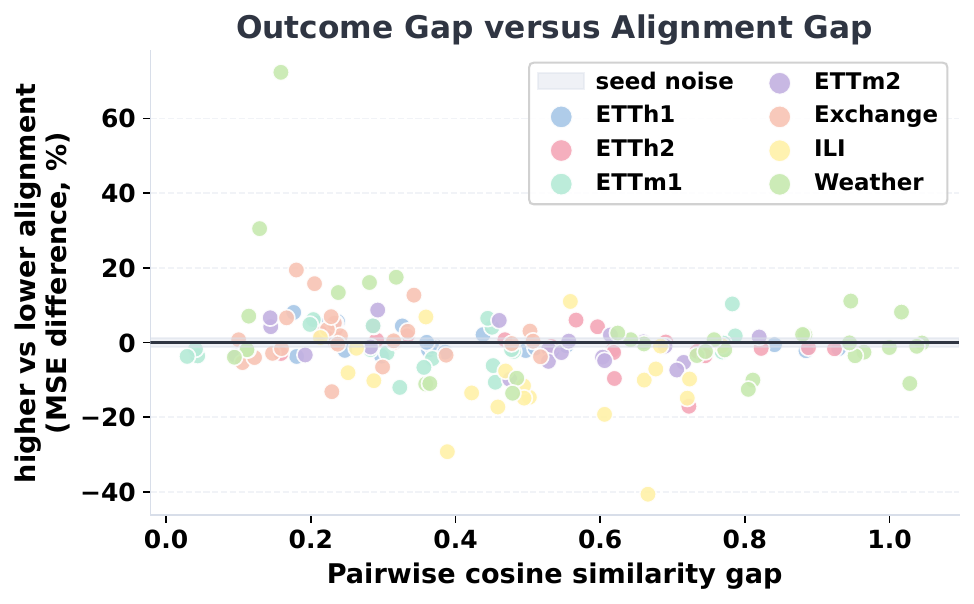}
\\[-1mm]
{\small (a) Harmed variables by dataset.}
&
{\small (b) Outcome gap versus alignment gap.}
\end{tabular}
\caption{(a) Fraction of variables for which shared training is worse than the target-specific oracle. (b) Across \(162\) target-and-size cells, the relative MSE difference between the two alignment groups versus the pairwise-alignment gap. Negative values favor higher alignment. The shaded \(\pm1.1\%\) band summarizes repeated-fit variation for fixed target-partner groups (Appendix~\ref{app:group_composition}).}
\label{fig:harm_subset_diagnostics}
\end{figure}
\FloatBarrier

For each reported target we rank the remaining variables by their post-warmup mean pairwise gradient cosine in the baseline model. For \(k\in\{2,3,4\}\), we train with its \(k-1\) highest-ranked or lowest-ranked partners, which form disjoint sets. Both conditions use matched replicate seeds, keep the full multivariate input, restrict the loss to the target and its partners, and are scored by held-out target MSE. Appendix~\ref{app:group_composition} gives the full protocol. Higher alignment wins \(100/162\) cells, and the all-cell median signed contrast is \(-1.4\%\), but the benchmark-averaged contrast is not distinguishable from zero. Figure~\ref{fig:harm_subset_diagnostics}(b) shows that a larger alignment gap does not reliably identify the better partner set, consistent with Appendix Figure~\ref{fig:group_gaps_all}, where ILI favors higher alignment at every group size but five of the other six datasets change sign across sizes. Group composition still matters, since the median absolute within-cell contrast is \(3.5\%\), over three times the \(1.1\%\) variation observed when the same target and partner set are retrained with different random seeds.

Together, the evidence shows that negative pairwise cosine similarities are common, but their frequency does not reveal which variables shared training harms. Partner composition can materially change a target's error, but mean gradient alignment does not reliably predict whether the change helps or hurts. The comparisons are associational rather than causal, and they motivate conditional, reliability-aware intervention rather than removing conflicts uniformly across variable subsets.

\section{Proposed Method}
\label{sec:method}

\begin{figure*}[t]
\centering
\includegraphics[width=\textwidth]{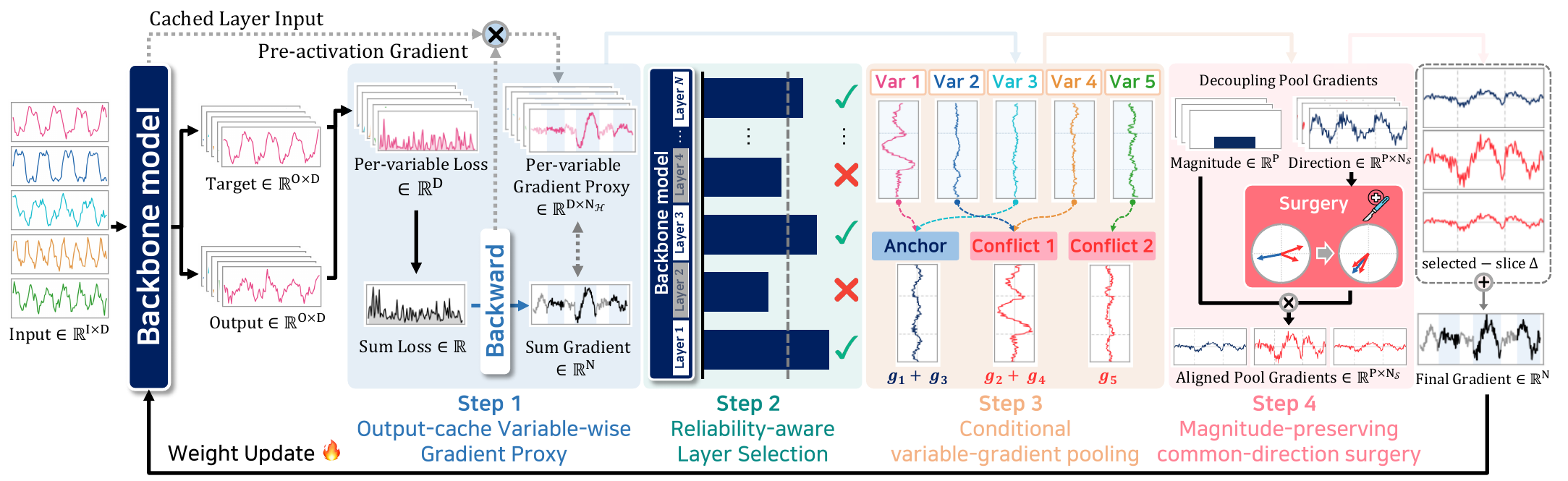}
\vspace{-4mm}
\caption{Overview of PV-Surgery. \textbf{Step 1: Output-cache variable-wise gradient proxy} reconstructs variable-wise proxy gradients from a single sum-loss backward pass and cached layer signals. \textbf{Step 2: Reliability-aware layer selection} selects slices by normalized proxy-sum error with output fallback. \textbf{Step 3: Conditional variable-gradient pooling} forms anchor and conflict pool gradients without dropping variables. \textbf{Step 4: Magnitude-preserving common-direction surgery} aligns variable or pooled gradients while restoring their pre-surgery norms to avoid reweighting.}
\label{fig:pv_surgery_overview}
\vspace{-2mm}
\end{figure*}

PV-Surgery is an optimizer-side gradient transformation that leaves the forecasting architecture, prediction target, and per-variable loss unchanged. Figure~\ref{fig:pv_surgery_overview} gives an overview of PV-Surgery, which consists of four stages that form a dependency chain. Computing exact per-variable gradients would require \(D\) costly backward passes, so PV-Surgery instead builds variable-wise gradient proxies from cached layer signals in one pass. Variable mixing can shift proxy sums from reference-gradient slices, so PV-Surgery selects layers by relative error and falls back to the output layer. Section~\ref{sec:partner_selection} shows that mean alignment does not reliably identify helpful partners, which is why variables are pooled conditionally at each step without dropping any of them. Finally, surgery aligns each variable or pooled gradient with the normalized mean direction, then restores its pre-surgery norm so the correction does not reweight the surgery inputs. Appendix~\ref{app:training_procedure} gives the training pseudocode.

\subsection{Problem setup}
For a batch of \(B\) windows, \(D\) variables, and horizon length \(H\), let \(\hat{Y},Y\in\mathbb{R}^{B\times H\times D}\) be the predicted and true future values and \(\varphi\) be the pointwise forecasting loss. The loss induced by variable \(d\) is
\begin{equation}
\label{eq:losses}
    L_d(\theta)=\frac{1}{BH}\sum_{b=1}^{B}\sum_{h=1}^{H}
    \varphi\!\left(\hat{Y}_{b,h,d},Y_{b,h,d}\right).
\end{equation}
Let \(L_{\Sigma}=\sum_{d=1}^{D}L_d\) and \(L_{\mathrm{mean}}=L_{\Sigma}/D\). With \(g_d=\nabla_\theta L_d\) and \(g_{\mathrm{mean}}=\nabla_\theta L_{\mathrm{mean}}\), PV-Surgery's exact single-backward reference gradient is
\begin{equation}
\label{eq:reference_gradient}
    g_0=\nabla_\theta L_{\Sigma}
    =\sum_{d=1}^{D}g_d
    =Dg_{\mathrm{mean}}.
\end{equation}
PV-Surgery backpropagates \(L_{\Sigma}\) once, avoiding \(D\) backward passes and a dense \(D\times|\theta|\) gradient tensor. At the output boundary, variable \(d\)'s cache row equals its \(g_d\) slice without \(1/D\) scaling because only that output slice enters \(L_d\), and the rows sum to the matching \(g_0\) slice without rescaling.

\subsection{Output-cache variable-wise gradient proxy}
One backward pass materializes only \(g_0\), so we must estimate variable-wise gradients from same-pass layer inputs and output-side backward signals. We hook cache-compatible linear layers \(\mathcal{H}\) to collect these quantities and convert them to a common variable-axis form by moving an explicit axis or reshaping batch-folded layouts. For a layer \(\ell\) with \(W_\ell\in\mathbb{R}^{F_{\mathrm{out}}\times F_{\mathrm{in}}}\), the caches keep a variable axis, \mbox{\(Z_\ell\in\mathbb{R}^{B\times D\times M_\ell\times F_{\mathrm{in}}}\)} and \mbox{\(E_\ell\in\mathbb{R}^{B\times D\times M_\ell\times F_{\mathrm{out}}}\)}, and flattening the batch and position axes gives \(Z_{\ell,d}\in\mathbb{R}^{B\,M_\ell\times F_{\mathrm{in}}}\) and \(E_{\ell,d}\in\mathbb{R}^{B\,M_\ell\times F_{\mathrm{out}}}\). The weight and bias gradients are then
\begin{equation}
\label{eq:cache_gradients}
    G^W_{\ell,d}=E_{\ell,d}^{\top}Z_{\ell,d},
    \qquad
    G^b_{\ell,d}=E_{\ell,d}^{\top}\mathbf{1}_{B\,M_\ell}.
\end{equation}
When a bias is present, it is concatenated with the flattened weight gradient.
\begin{equation}
\label{eq:cache_matrix}
	    G_\ell=\operatorname{stack}_{d=1}^{D}
	    \!\left(\operatorname{concat}\!\left(
	    \operatorname{vec}(G^W_{\ell,d}),G^b_{\ell,d}\right)\right)
	    \in\mathbb{R}^{D\times N_\ell}.
\end{equation}
Equation~\ref{eq:cache_gradients} is the per-variable chain-rule gradient at the output boundary and a proxy after variable mixing, motivating the reliability test. If a hooked layer lacks a variable-axis form for that batch, \(G_\ell=0_{D\times N_\ell}\). Here \(M_\ell\) counts positions sharing \(W_\ell\) within a variable and \(N_\ell\) cached parameters.

\subsection{Reliability-aware layer selection}
Reconstruction consistency varies by layer and backbone and can shift during training, so the method selects its layer set dynamically at each step rather than fixing it in advance. For every hooked layer \(\ell\in\mathcal{H}\), we compare the summed proxy with the matching slice of the single-backward reference.
\begin{equation}
\label{eq:layer_reliability}
    g_{\mathrm{proxy},\ell}=\sum_{d=1}^{D}G_\ell[d,\cdot],
    \qquad
    e_\ell=\frac{\|g_{\mathrm{proxy},\ell}-g_{0,\ell}\|_2}
    {\|g_{0,\ell}\|_2+\epsilon}.
\end{equation}
Reconstruction reliability \(r_\ell=(1+e_\ell)^{-1}\) scores proxy-sum agreement, while normalized slice norm \(q_\ell=\|g_{0,\ell}\|_2/(\|g_0\|_2+\epsilon)\) measures reference-slice magnitude. Let \(\mathcal H_{\mathrm{elig}}\) be the unprotected hooked layers. The median-reliability gate retains its non-output layers at or above median reliability and includes the output layer. For the resulting \(\mathcal H_{\mathrm{gate}}\), we set \(\mathcal A=\{\ell\in\mathcal H_{\mathrm{gate}}\mid q_\ell>0\}\) and compute
\begin{equation}
\label{eq:effective_coverage}
    \pi_\ell=\frac{q_\ell}{\sum_{j\in\mathcal{A}}q_j},
    \quad
    \mathcal{E}_{\mathrm{layer}}=-\sum_{\ell\in\mathcal{A}}
    \pi_\ell\log\pi_\ell,
    \quad
    K_{\mathrm{eff}}=\left\lceil e^{\mathcal{E}_{\mathrm{layer}}}\right\rceil.
\end{equation}
It takes the \(K_{\mathrm{eff}}\) largest-\(q_\ell\) layers and the output layer that pass the reconstruction-error check, falling back to the output layer if none do (Appendix~\ref{app:training_procedure}). Here \(e^{\mathcal{E}_{\mathrm{layer}}}\) is \(\pi\)'s exponentiated Shannon entropy, or order-one Hill number \citep{hill1973diversity}, and \(K_{\mathrm{eff}}\) is its ceiling. The selected set \(\mathcal{S}\) then gives
\begin{equation}
\label{eq:selected_gradients}
    G_{\mathcal{S}}=\operatorname{concat}_{\ell\in\mathcal{S}}G_\ell
    \in\mathbb{R}^{D\times N_{\mathcal{S}}},
    \qquad
    N_{\mathcal{S}}=\sum_{\ell\in\mathcal{S}}N_\ell
    \leq N_{\mathcal{H}}.
\end{equation}
Here \(N_{\mathcal{H}}=\sum_{\ell\in\mathcal{H}}N_\ell\) counts all cached parameters. Fixing the support before pooling or direction change keeps surgery out of layers where the cache decomposition is not reconstruction-consistent, and leaves later stages operating on \(N_{\mathcal{S}}\) rather than \(N_{\mathcal{H}}\) parameters.

\subsection{Conditional variable-gradient pooling}
Treating every variable gradient as a surgery objective would make direction surgery reconcile all pairs, although Section~\ref{sec:diagnosis} shows that a negative cosine alone does not reliably imply harmful sharing. We keep every variable but reduce the objective count. With \(G_{\mathrm{pre}}=G_{\mathcal{S}}\), each row contains one variable's proxy gradient over the selected parameter slices, and \(g_{0,\mathcal S}\) is the exact sum-loss gradient over those same slices:
\begin{equation}
\label{eq:pool_reference}
    g_{\mathrm{ref}}=g_{0,\mathcal{S}},
    \qquad
    a_d=\cos(G_{\mathrm{pre}}[d,\cdot],g_{\mathrm{ref}}).
\end{equation}
Pooling requires two finite, nonzero rows with a negative cosine, not just a row opposed to \(g_{\mathrm{ref}}\). Rows with \(a_d\geq0\) are aligned to the reference and the rest conflict with it. An anchor combines at least two aligned rows whose sum has nonnegative cosine with each. Each conflicting row greedily joins the pool whose current sum is most aligned with it, provided their cosine is positive. The partition \(M_{\mathrm{pool}}\in\{0,1\}^{P\times D}\) maps \(D\) rows to \(P\) objectives while preserving their aggregate gradient:
\begin{equation}
\label{eq:sum_pool}
    G_{\mathrm{pool}}=M_{\mathrm{pool}}G_{\mathrm{pre}}
    \in\mathbb{R}^{P\times N_{\mathcal{S}}},
    \qquad
    \sum_{k=1}^{P}G_{\mathrm{pool}}[k,\cdot]
    =\sum_{d=1}^{D}G_{\mathrm{pre}}[d,\cdot].
\end{equation}
Without pairwise conflict, the pooled branch skips direction surgery and remains \(G_p^{\mathrm{post}}=G_{\mathrm{pre}}\). The subsequent safe candidate-selection step may retain it over the corrected unpooled candidate.

\subsection{Magnitude-preserving common-direction surgery}
Pooling reduces the objective count, but pooled gradients can still disagree, so their directions must be adjusted and the unpooled and pooled branches remain candidates. For unpooled \(u\) and pooled \(p\), the inputs are \(G_{\mathrm{in}}^{(u)}=G_{\mathrm{pre}}\) and, when pooling is active, \(G_{\mathrm{in}}^{(p)}=G_{\mathrm{pool}}\). For \(c\in\{u,p\}\), let \(P_c\) be its objective count and decompose each nonzero objective into its magnitude and unit direction:
\begin{equation}
\label{eq:magnitude_direction}
    m_i^{(c)}=\|G_{\mathrm{in}}^{(c)}[i,\cdot]\|_2,
    \qquad
    \hat g_i^{(c)}=\frac{G_{\mathrm{in}}^{(c)}[i,\cdot]}{m_i^{(c)}}.
\end{equation}
Objectives with \(m_i^{(c)}=0\) take \(\hat g_i^{(c)}=0\). Let \(\bar g^{(c)}=P_c^{-1}\sum_i\hat g_i^{(c)}\). The operator aligns each objective with \(\bar g^{(c)}\) while preserving its input magnitude, and makes no change when \(\|\bar g^{(c)}\|_2=0\).
\begin{equation}
\label{eq:common_direction}
    \widetilde G_{\mathrm{in}}^{(c)}[i,\cdot]=
    \begin{cases}
    m_i^{(c)}\,\bar g^{(c)}/\|\bar g^{(c)}\|_2,
        & \|\bar g^{(c)}\|_2>0,\\
    G_{\mathrm{in}}^{(c)}[i,\cdot],
        & \|\bar g^{(c)}\|_2=0,
    \end{cases}
    \quad\text{and}\quad
    G_u^{\mathrm{post}}=\widetilde G_{\mathrm{in}}^{(u)}.
\end{equation}

\paragraph{Pool restoration and safe candidate selection.}
For active pooling, pool \(k\) distributes the change made by surgery equally across its member set \(\mathcal{M}_k\).
\begin{equation}
\label{eq:pool_restore}
    G_p^{\mathrm{post}}[d,\cdot]=G_{\mathrm{pre}}[d,\cdot]
    +\frac{\widetilde G_{\mathrm{in}}^{(p)}[k,\cdot]-G_{\mathrm{pool}}[k,\cdot]}
    {|\mathcal{M}_k|},
    \qquad d\in\mathcal{M}_k.
\end{equation}
This restores \(D\) variable rows while preserving each post-surgery pool sum, but not individual row norms. The descent-lexicographic selector compares \(G_u^{\mathrm{post}}\) with \(G_p^{\mathrm{post}}\). For candidate \(c\), let \(g_c=\sum_dG_c^{\mathrm{post}}[d,\cdot]\). For each finite, nonzero variable-wise gradient, the selector computes the following scores:
\begin{equation}
\label{eq:safe_selector}
    \gamma_{c,d}=\cos(G_{\mathrm{pre}}[d,\cdot],g_c),
    \qquad
    v_{c,d}=\mathbf{1}[\gamma_{c,d}<0].
\end{equation}
It first minimizes \(\sum_dv_{c,d}\), then compares the sorted cosine vectors from worst to best. Exact or numerically ambiguous ties retain the unpooled candidate. This ranks candidates and does not guarantee a descent-compatible update for every variable when both contain violations.

\paragraph{Final gradient assembly.}
For selected \(G_{\mathrm{post}}\), \(\Delta_{\mathcal S}\) is the change in the aggregate selected-subspace gradient:
\begin{equation}
\label{eq:surgery_delta}
    \Delta_{\mathcal{S}}=\sum_{d=1}^{D}G_{\mathrm{post}}[d,\cdot]
    -\sum_{d=1}^{D}G_{\mathrm{pre}}[d,\cdot].
\end{equation}
For each hooked layer \(\ell\), let \(\Omega_\ell\) be its flattened parameter indices, and let \(\Omega_{\mathcal{S}}=\bigcup_{\ell\in\mathcal{S}}\Omega_\ell\). The full-sum assembly inserts every hooked-slice proxy sum and then adds the selected-subspace correction.
\begin{equation}
\label{eq:gradient_assembly}
    g_{\mathrm{final}}\leftarrow g_0,
    \quad
    g_{\mathrm{final}}[\Omega_\ell]\leftarrow\sum_{d=1}^{D}G_\ell[d,\cdot]\ \ (\ell\in\mathcal{H}),
    \quad
    g_{\mathrm{final}}[\Omega_{\mathcal{S}}]\leftarrow g_{\mathrm{final}}[\Omega_{\mathcal{S}}]+\Delta_{\mathcal{S}}.
\end{equation}
Thus, non-hooked parameters retain \(g_0\). Unselected hooked slices use their variable-wise proxy sums, and selected slices add the corresponding \(\Delta_{\mathcal S}\) entries.

\section{Experiments}

\subsection{Experimental setup}
\paragraph{Datasets.}
We evaluate PV-Surgery on ETTh1, ETTh2, ETTm1, ETTm2, Weather, and Exchange using an input length of \(96\) and prediction lengths \(\{96,192,336,720\}\). For ILI, the input length is \(36\) and the prediction lengths are \(\{24,36,48,60\}\). All methods share chronological splits. Appendices~\ref{app:experimental_configuration} and \ref{app:dataset_diagnostics} give settings and dataset statistics.

\paragraph{Baselines.}
We compare standard MSE training and PV-Surgery on DLinear \citep{zeng2023transformers}, iTransformer \citep{liu2024itransformer}, MICN \citep{wang2023micn}, SCINet \citep{liu2022scinet}, and TimeXer \citep{wang2024timexer}. We additionally compare four forecasting-specific methods, TILDE-Q \citep{lee2024tildeq}, FreDF \citep{wang2024fredf}, PSLoss \citep{kudrat2025psloss}, and Selective Learning \citep{fu2025selective}. All use the common protocol of Appendix~\ref{app:experimental_configuration}, with iTransformer for this comparison.

\paragraph{Metrics.}
We report test mean squared error (MSE) and mean absolute error (MAE), both averaged over test windows, horizons, and variables. Exact definitions are provided in Appendix~\ref{app:experimental_configuration}.

\subsection{Main results}
\label{sec:main_results}

Tables~\ref{tab:backbone_pv} and \ref{tab:backbone_pv_weather_ili} report all \(140\) standard-MSE comparisons, with Weather and ILI in Appendix~\ref{app:backbone_weather_ili} due to the page limit. Separately, Table~\ref{tab:loss_baselines_itransformer} compares PV-Surgery with four forecasting-specific methods using iTransformer across five datasets. Appendix~\ref{app:loss_baselines_all_models} covers all backbones and benchmarks.

\begin{table}[H]
\vspace{-4mm}
\centering
\caption{Backbone comparison between standard MSE training and PV-Surgery on five of the seven benchmarks. Within each backbone and metric, the better value between MSE and +PV is bolded using unrounded scores. Weather and ILI use the same protocol and are reported in Table~\ref{tab:backbone_pv_weather_ili}.}
\label{tab:backbone_pv}
\vspace{1mm}
\setlength{\tabcolsep}{2.4pt}
\renewcommand{\arraystretch}{0.96}
\scriptsize
\resizebox{\textwidth}{!}{%
\begin{tabular}{@{}c|c|rrrr|rrrr|rrrr|rrrr|rrrr@{}}
\toprule
\multicolumn{2}{c|}{Models} & \multicolumn{4}{c|}{DLinear} & \multicolumn{4}{c|}{iTransformer} & \multicolumn{4}{c|}{MICN} & \multicolumn{4}{c|}{SCINet} & \multicolumn{4}{c}{TimeXer} \\
\midrule
\multicolumn{2}{c|}{Training setup} & \multicolumn{2}{c}{MSE} & \multicolumn{2}{c|}{+PV} & \multicolumn{2}{c}{MSE} & \multicolumn{2}{c|}{+PV} & \multicolumn{2}{c}{MSE} & \multicolumn{2}{c|}{+PV} & \multicolumn{2}{c}{MSE} & \multicolumn{2}{c|}{+PV} & \multicolumn{2}{c}{MSE} & \multicolumn{2}{c}{+PV} \\
\midrule
\multicolumn{2}{c|}{Metric} & \multicolumn{1}{c}{MSE} & \multicolumn{1}{c}{MAE} & \multicolumn{1}{c}{MSE} & \multicolumn{1}{c|}{MAE} & \multicolumn{1}{c}{MSE} & \multicolumn{1}{c}{MAE} & \multicolumn{1}{c}{MSE} & \multicolumn{1}{c|}{MAE} & \multicolumn{1}{c}{MSE} & \multicolumn{1}{c}{MAE} & \multicolumn{1}{c}{MSE} & \multicolumn{1}{c|}{MAE} & \multicolumn{1}{c}{MSE} & \multicolumn{1}{c}{MAE} & \multicolumn{1}{c}{MSE} & \multicolumn{1}{c|}{MAE} & \multicolumn{1}{c}{MSE} & \multicolumn{1}{c}{MAE} & \multicolumn{1}{c}{MSE} & \multicolumn{1}{c}{MAE} \\
\midrule
\multirow{5}{*}{\textbf{ETTh1}} & 96 & 0.462 & 0.444 & \textbf{0.461} & \textbf{0.440} & 0.454 & 0.447 & \textbf{0.451} & \textbf{0.442} & 0.552 & 0.530 & \textbf{0.492} & \textbf{0.497} & 0.490 & 0.464 & \textbf{0.473} & \textbf{0.449} & \textbf{0.460} & 0.454 & 0.461 & \textbf{0.450} \\
 & 192 & 0.514 & 0.476 & \textbf{0.514} & \textbf{0.471} & 0.517 & 0.485 & \textbf{0.508} & \textbf{0.476} & 0.594 & 0.553 & \textbf{0.561} & \textbf{0.543} & 0.545 & 0.495 & \textbf{0.530} & \textbf{0.483} & \textbf{0.511} & 0.485 & 0.518 & \textbf{0.485} \\
 & 336 & 0.559 & 0.506 & \textbf{0.556} & \textbf{0.501} & 0.561 & 0.513 & \textbf{0.550} & \textbf{0.504} & \textbf{0.688} & \textbf{0.623} & 0.700 & 0.630 & 0.590 & 0.521 & \textbf{0.574} & \textbf{0.510} & 0.564 & \textbf{0.511} & \textbf{0.564} & 0.515 \\
 & 720 & 0.651 & 0.573 & \textbf{0.641} & \textbf{0.561} & 0.671 & 0.583 & \textbf{0.661} & \textbf{0.575} & \textbf{0.753} & \textbf{0.671} & 0.798 & 0.700 & 0.700 & 0.588 & \textbf{0.687} & \textbf{0.581} & 0.693 & 0.590 & \textbf{0.691} & \textbf{0.581} \\
\cmidrule(lr){2-22}
 & Avg & 0.546 & 0.500 & \textbf{0.543} & \textbf{0.493} & 0.551 & 0.507 & \textbf{0.543} & \textbf{0.499} & 0.647 & 0.594 & \textbf{0.638} & \textbf{0.593} & 0.581 & 0.517 & \textbf{0.566} & \textbf{0.506} & \textbf{0.557} & 0.510 & 0.558 & \textbf{0.508} \\
\midrule
\multirow{5}{*}{\textbf{ETTh2}} & 96 & 0.239 & 0.329 & \textbf{0.229} & \textbf{0.313} & 0.239 & 0.323 & \textbf{0.229} & \textbf{0.314} & 0.238 & 0.333 & \textbf{0.230} & \textbf{0.320} & 0.254 & 0.334 & \textbf{0.237} & \textbf{0.317} & 0.233 & 0.319 & \textbf{0.231} & \textbf{0.316} \\
 & 192 & 0.329 & 0.393 & \textbf{0.290} & \textbf{0.357} & 0.303 & 0.366 & \textbf{0.287} & \textbf{0.355} & 0.315 & 0.388 & \textbf{0.291} & \textbf{0.369} & 0.311 & 0.370 & \textbf{0.297} & \textbf{0.358} & \textbf{0.294} & 0.360 & 0.295 & \textbf{0.358} \\
 & 336 & 0.437 & 0.462 & \textbf{0.341} & \textbf{0.396} & 0.356 & 0.401 & \textbf{0.334} & \textbf{0.388} & 0.452 & 0.472 & \textbf{0.352} & \textbf{0.412} & 0.356 & 0.402 & \textbf{0.346} & \textbf{0.391} & \textbf{0.341} & 0.394 & 0.342 & \textbf{0.390} \\
 & 720 & 0.650 & 0.582 & \textbf{0.470} & \textbf{0.484} & 0.457 & 0.460 & \textbf{0.446} & \textbf{0.452} & 0.672 & 0.583 & \textbf{0.513} & \textbf{0.520} & 0.470 & 0.472 & \textbf{0.447} & \textbf{0.454} & 0.449 & 0.457 & \textbf{0.446} & \textbf{0.451} \\
\cmidrule(lr){2-22}
 & Avg & 0.414 & 0.442 & \textbf{0.332} & \textbf{0.388} & 0.339 & 0.388 & \textbf{0.324} & \textbf{0.377} & 0.419 & 0.444 & \textbf{0.346} & \textbf{0.405} & 0.347 & 0.394 & \textbf{0.332} & \textbf{0.380} & 0.329 & 0.382 & \textbf{0.328} & \textbf{0.379} \\
\midrule
\multirow{5}{*}{\textbf{ETTm1}} & 96 & 0.388 & 0.395 & \textbf{0.383} & \textbf{0.389} & 0.407 & 0.415 & \textbf{0.402} & \textbf{0.404} & 0.445 & 0.472 & \textbf{0.430} & \textbf{0.460} & 0.448 & 0.428 & \textbf{0.421} & \textbf{0.414} & 0.432 & 0.421 & \textbf{0.396} & \textbf{0.403} \\
 & 192 & 0.447 & 0.426 & \textbf{0.445} & \textbf{0.421} & 0.480 & 0.457 & \textbf{0.461} & \textbf{0.439} & 0.495 & 0.511 & \textbf{0.470} & \textbf{0.487} & \textbf{0.488} & 0.453 & 0.490 & \textbf{0.446} & 0.477 & 0.453 & \textbf{0.455} & \textbf{0.429} \\
 & 336 & 0.507 & 0.459 & \textbf{0.505} & \textbf{0.454} & 0.520 & 0.477 & \textbf{0.517} & \textbf{0.473} & 0.567 & 0.570 & \textbf{0.508} & \textbf{0.513} & \textbf{0.548} & 0.483 & 0.551 & \textbf{0.475} & 0.521 & 0.479 & \textbf{0.505} & \textbf{0.462} \\
 & 720 & 0.575 & 0.502 & \textbf{0.573} & \textbf{0.499} & 0.618 & 0.531 & \textbf{0.575} & \textbf{0.510} & 0.627 & 0.596 & \textbf{0.576} & \textbf{0.562} & \textbf{0.614} & 0.521 & 0.628 & \textbf{0.516} & 0.580 & 0.515 & \textbf{0.564} & \textbf{0.503} \\
\cmidrule(lr){2-22}
 & Avg & 0.479 & 0.446 & \textbf{0.476} & \textbf{0.441} & 0.506 & 0.470 & \textbf{0.488} & \textbf{0.457} & 0.533 & 0.537 & \textbf{0.496} & \textbf{0.505} & 0.524 & 0.471 & \textbf{0.523} & \textbf{0.463} & 0.502 & 0.467 & \textbf{0.480} & \textbf{0.449} \\
\midrule
\multirow{5}{*}{\textbf{ETTm2}} & 96 & 0.160 & 0.268 & \textbf{0.155} & \textbf{0.253} & 0.157 & 0.258 & \textbf{0.150} & \textbf{0.247} & 0.166 & 0.276 & \textbf{0.146} & \textbf{0.248} & 0.155 & 0.253 & \textbf{0.150} & \textbf{0.246} & 0.151 & 0.249 & \textbf{0.149} & \textbf{0.247} \\
 & 192 & 0.209 & 0.308 & \textbf{0.197} & \textbf{0.287} & 0.205 & 0.295 & \textbf{0.195} & \textbf{0.283} & 0.202 & 0.303 & \textbf{0.190} & \textbf{0.285} & 0.202 & 0.290 & \textbf{0.193} & \textbf{0.278} & 0.200 & 0.287 & \textbf{0.193} & \textbf{0.282} \\
 & 336 & 0.262 & 0.350 & \textbf{0.239} & \textbf{0.319} & 0.252 & 0.328 & \textbf{0.240} & \textbf{0.317} & 0.266 & 0.355 & \textbf{0.231} & \textbf{0.316} & 0.248 & 0.322 & \textbf{0.236} & \textbf{0.309} & 0.239 & 0.316 & \textbf{0.235} & \textbf{0.312} \\
 & 720 & 0.348 & 0.408 & \textbf{0.306} & \textbf{0.365} & 0.322 & 0.373 & \textbf{0.310} & \textbf{0.365} & 0.389 & 0.434 & \textbf{0.331} & \textbf{0.389} & 0.321 & 0.368 & \textbf{0.308} & \textbf{0.357} & 0.312 & 0.367 & \textbf{0.307} & \textbf{0.360} \\
\cmidrule(lr){2-22}
 & Avg & 0.245 & 0.334 & \textbf{0.225} & \textbf{0.306} & 0.234 & 0.313 & \textbf{0.224} & \textbf{0.303} & 0.256 & 0.342 & \textbf{0.224} & \textbf{0.310} & 0.231 & 0.308 & \textbf{0.222} & \textbf{0.297} & 0.226 & 0.305 & \textbf{0.221} & \textbf{0.300} \\
\midrule
\multirow{5}{*}{\textbf{Exchange}} & 96 & 0.105 & 0.234 & \textbf{0.100} & \textbf{0.226} & 0.108 & 0.236 & \textbf{0.105} & \textbf{0.231} & 0.123 & 0.262 & \textbf{0.109} & \textbf{0.242} & 0.120 & 0.249 & \textbf{0.116} & \textbf{0.244} & \textbf{0.113} & 0.239 & 0.113 & \textbf{0.239} \\
 & 192 & 0.224 & 0.349 & \textbf{0.191} & \textbf{0.323} & 0.222 & 0.344 & \textbf{0.211} & \textbf{0.335} & 0.219 & 0.352 & \textbf{0.212} & \textbf{0.348} & 0.223 & 0.347 & \textbf{0.222} & \textbf{0.346} & 0.216 & 0.337 & \textbf{0.213} & \textbf{0.335} \\
 & 336 & 0.425 & 0.498 & \textbf{0.321} & \textbf{0.425} & 0.397 & 0.463 & \textbf{0.390} & \textbf{0.460} & 0.498 & 0.525 & \textbf{0.378} & \textbf{0.470} & \textbf{0.405} & \textbf{0.470} & 0.409 & 0.473 & 0.396 & 0.459 & \textbf{0.382} & \textbf{0.452} \\
 & 720 & \textbf{0.652} & \textbf{0.640} & 0.847 & 0.724 & 1.099 & 0.800 & \textbf{1.075} & \textbf{0.792} & 2.931 & 1.400 & \textbf{1.106} & \textbf{0.793} & \textbf{1.101} & \textbf{0.804} & 1.103 & 0.806 & 1.105 & 0.798 & \textbf{1.010} & \textbf{0.763} \\
\cmidrule(lr){2-22}
 & Avg & \textbf{0.352} & 0.430 & 0.365 & \textbf{0.424} & 0.457 & 0.461 & \textbf{0.446} & \textbf{0.454} & 0.943 & 0.635 & \textbf{0.451} & \textbf{0.463} & \textbf{0.462} & 0.468 & 0.463 & \textbf{0.467} & 0.457 & 0.458 & \textbf{0.429} & \textbf{0.447} \\
\bottomrule
\end{tabular}%
}
\vspace{-5mm}
\end{table}

PV-Surgery obtains lower MSE in \(111\) of the \(140\) settings and lower MAE in \(119\), both counted on unrounded scores, and averaging the relative change within each setting gives \(3.61\%\) in MSE and \(2.93\%\) in MAE. The improvement is largest on MICN, at \(8.50\%\) in MSE and \(5.67\%\) in MAE, and on DLinear, at \(5.38\%\) and \(5.01\%\). PV-Surgery therefore improves the majority of settings for every backbone and dataset, although the direction and size of the average change depend on both.

\begin{table}[H]
\vspace{-4mm}
\centering
\caption{Comparison of PV-Surgery with four forecasting-specific methods on iTransformer for five of the seven benchmarks. The MSE row denotes standard MSE training. Best results are bolded and second-best results are underlined using unrounded scores.}
\label{tab:loss_baselines_itransformer}
\vspace{1mm}
\setlength{\tabcolsep}{2.6pt}
\renewcommand{\arraystretch}{1.02}
\scriptsize
\resizebox{\textwidth}{!}{%
\begin{tabular}{@{}c|c|rrrr|rrrr|rrrr|rrrr|rrrr@{}}
\toprule
\multicolumn{2}{c|}{Dataset} & \multicolumn{4}{c|}{ETTh1} & \multicolumn{4}{c|}{ETTh2} & \multicolumn{4}{c|}{ETTm1} & \multicolumn{4}{c|}{ETTm2} & \multicolumn{4}{c}{Exchange} \\
\midrule
\multicolumn{2}{c|}{Forecast length} & 96 & 192 & 336 & \multicolumn{1}{c|}{720} & 96 & 192 & 336 & \multicolumn{1}{c|}{720} & 96 & 192 & 336 & \multicolumn{1}{c|}{720} & 96 & 192 & 336 & \multicolumn{1}{c|}{720} & 96 & 192 & 336 & \multicolumn{1}{c}{720} \\
\midrule
\multirow{2}{*}{\textbf{MSE}} & MSE & 0.454 & 0.517 & 0.561 & 0.671 & 0.239 & 0.303 & 0.356 & 0.457 & 0.407 & 0.480 & \underline{0.520} & 0.618 & 0.157 & 0.205 & 0.252 & 0.322 & 0.108 & 0.222 & 0.397 & \underline{1.099} \\
 & MAE & 0.447 & 0.485 & 0.513 & 0.583 & 0.323 & 0.366 & 0.401 & 0.460 & 0.415 & 0.457 & 0.477 & 0.531 & 0.258 & 0.295 & 0.328 & 0.373 & 0.236 & 0.344 & 0.463 & \underline{0.800} \\
\midrule
\multirow{2}{*}{\textbf{TILDE-Q}} & MSE & 0.469 & 0.528 & 0.564 & 0.669 & 0.233 & \underline{0.287} & 0.354 & 0.473 & \textbf{0.402} & 0.465 & 0.537 & 0.622 & 0.156 & 0.200 & 0.247 & 0.311 & 0.111 & 0.219 & 0.428 & 1.109 \\
 & MAE & 0.455 & 0.489 & 0.510 & 0.577 & 0.315 & \underline{0.353} & 0.400 & 0.466 & \underline{0.404} & 0.440 & 0.483 & 0.534 & 0.251 & 0.285 & 0.320 & 0.362 & 0.239 & 0.341 & 0.481 & 0.806 \\
\midrule
\multirow{2}{*}{\textbf{FreDF}} & MSE & \textbf{0.451} & \underline{0.512} & 0.560 & \underline{0.662} & 0.235 & 0.295 & 0.350 & 0.450 & 0.410 & 0.475 & 0.553 & 0.603 & 0.154 & 0.197 & 0.245 & \textbf{0.309} & 0.112 & 0.223 & 0.462 & 1.132 \\
 & MAE & \underline{0.445} & \underline{0.478} & 0.508 & 0.576 & 0.317 & 0.360 & 0.397 & 0.455 & 0.408 & 0.452 & 0.492 & 0.524 & 0.250 & 0.284 & 0.319 & \underline{0.362} & 0.241 & 0.344 & 0.502 & 0.819 \\
\midrule
\multirow{2}{*}{\textbf{PSLoss}} & MSE & 0.458 & 0.513 & \underline{0.558} & 0.665 & 0.234 & 0.291 & 0.360 & 0.473 & 0.429 & 0.499 & 0.538 & 0.626 & \underline{0.153} & \underline{0.196} & \textbf{0.239} & 0.310 & 0.113 & 0.213 & 0.413 & 1.146 \\
 & MAE & 0.448 & 0.480 & 0.509 & 0.577 & 0.316 & 0.355 & 0.406 & 0.469 & 0.421 & 0.459 & 0.482 & 0.535 & \underline{0.247} & \textbf{0.281} & \textbf{0.314} & \textbf{0.361} & 0.241 & 0.338 & 0.476 & 0.825 \\
\midrule
\multirow{2}{*}{\textbf{SL}} & MSE & 0.465 & 0.527 & 0.560 & 0.665 & \textbf{0.229} & \textbf{0.284} & \underline{0.348} & \textbf{0.438} & 0.404 & \textbf{0.458} & 0.521 & \underline{0.590} & 0.153 & 0.199 & 0.243 & 0.312 & \underline{0.108} & \textbf{0.211} & \underline{0.393} & 1.228 \\
 & MAE & 0.449 & 0.485 & \textbf{0.501} & \textbf{0.568} & \textbf{0.312} & \textbf{0.349} & \underline{0.393} & \textbf{0.449} & \textbf{0.400} & \textbf{0.432} & \textbf{0.465} & \textbf{0.505} & 0.250 & 0.285 & \underline{0.317} & 0.362 & \underline{0.232} & \textbf{0.332} & \underline{0.460} & 0.856 \\
\midrule
\multirow{2}{*}{\textbf{PV(Ours)}} & MSE & \underline{0.451} & \textbf{0.508} & \textbf{0.550} & \textbf{0.661} & \underline{0.229} & 0.287 & \textbf{0.334} & \underline{0.446} & \underline{0.402} & \underline{0.461} & \textbf{0.517} & \textbf{0.575} & \textbf{0.150} & \textbf{0.195} & \underline{0.240} & \underline{0.310} & \textbf{0.105} & \underline{0.211} & \textbf{0.390} & \textbf{1.075} \\
 & MAE & \textbf{0.442} & \textbf{0.476} & \underline{0.504} & \underline{0.575} & \underline{0.314} & 0.355 & \textbf{0.388} & \underline{0.452} & 0.404 & \underline{0.439} & \underline{0.473} & \underline{0.510} & \textbf{0.247} & \underline{0.283} & \underline{0.317} & 0.365 & \textbf{0.231} & \underline{0.335} & \textbf{0.460} & \textbf{0.792} \\
\bottomrule
\end{tabular}%
}
\vspace{-5mm}
\end{table}

Across the five datasets in Table~\ref{tab:loss_baselines_itransformer}, PV-Surgery ranks first or second in \(19/20\) MSE cells and \(17/20\) MAE cells. No method is best throughout. Some competing methods perform better in individual ETT settings, while PV-Surgery is particularly competitive at longer horizons and on Exchange. This pattern supports the diagnostic premise that intervention should preserve useful sharing rather than treat every disagreement as harmful. PV-Surgery acts on the update rather than the loss and can therefore complement these methods instead of replacing them.

\subsection{Ablation study}
\label{sec:ablation}

The full configuration gives the lowest MSE and MAE. Removing the output-cache proxy or conflict pooling causes the largest degradations. Because the output-cache ablation also disables cache-dependent layer selection, pooling, and candidate comparison, its large performance drop reflects several coupled components being removed together. Removing conflict pooling degrades both metrics much more than removing anchor pooling, consistent with its role in grouping variables into fewer, directionally coherent objectives when conflicts are frequent. Without layer selection, surgery runs on every hooked layer, lowering accuracy and raising mean wall-clock training time from \(332.45\) to \(642.98\) seconds. Appendix~\ref{app:ablation_all_datasets} reports dataset-level results.

\begin{table}[H]
\vspace{-5mm}
\centering
\caption{Core component ablation on MICN--Exchange, averaged over four horizons.
\(\Delta\) is the relative change against the full recipe.
Appendix~\ref{app:ablation_all_datasets} reports the same table for every dataset.}
\label{tab:core_ablation}
\vspace{1mm}
\footnotesize
\renewcommand{\arraystretch}{0.5}
\begin{tabular*}{\linewidth}{@{\extracolsep{\fill}}cl>{\fontsize{7.75}{9}\selectfont}r>{\scriptsize}r>{\fontsize{7.75}{9}\selectfont}r>{\scriptsize}r>{\fontsize{7.75}{9}\selectfont}c@{}}
\toprule
\multirow{2}{*}{\textbf{Category}} & \multirow{2}{*}{\textbf{Module}}
  & \multicolumn{2}{c}{\textbf{MSE}} & \multicolumn{2}{c}{\textbf{MAE}}
  & \multirow{2}{*}{\textbf{Time (s)}} \\
\cmidrule(lr){3-4}\cmidrule(lr){5-6}
& & \textbf{value} & \(\boldsymbol{\Delta\%}\) & \textbf{value} & \(\boldsymbol{\Delta\%}\) & \\
\midrule
Gradient proxy & w/o output-cache proxy & 0.937 & \(+107.7\) & 0.629 & \(+35.7\) & 338.92 \\
\cmidrule(lr){1-7}
Layer selection & w/o layer selection & 0.538 & \(+19.1\) & 0.482 & \(+4.0\) & 642.98 \\
\cmidrule(lr){1-7}
\multirow{3}{*}{Variable pooling}
  & w/o conflict pooling & 0.858 & \(+90.0\) & 0.607 & \(+31.0\) & 253.42 \\
  & w/o anchor pooling   & 0.506 & \(+12.1\) & 0.478 & \(+3.2\)  & 272.47 \\
  & w/o variable pooling & 0.497 & \(+10.0\) & 0.474 & \(+2.2\)  & 277.40 \\
\cmidrule(lr){1-7}
\multirow{2}{*}{Direction surgery}
  & w/o magnitude decoupling & 0.512 & \(+13.5\) & 0.493 & \(+6.3\) & 333.05 \\
  & w/o safe candidate selection & 0.478 & \(+6.0\)  & 0.478 & \(+3.1\) & 312.27 \\
\midrule
\multicolumn{2}{c}{\textbf{PV-Surgery (full)}} & \textbf{0.451} & -- & \textbf{0.463} & -- & 332.45 \\
\bottomrule
\end{tabular*}
\vspace{-5mm}
\end{table}

\ifdefined\ARXIVVERSION
\clearpage
\fi
\subsection{Mechanism evidence}
\label{sec:mechanism}

\begin{wrapfigure}[14]{r}{0.42\linewidth}
\vspace{-6mm}
\centering
\includegraphics[width=\linewidth]{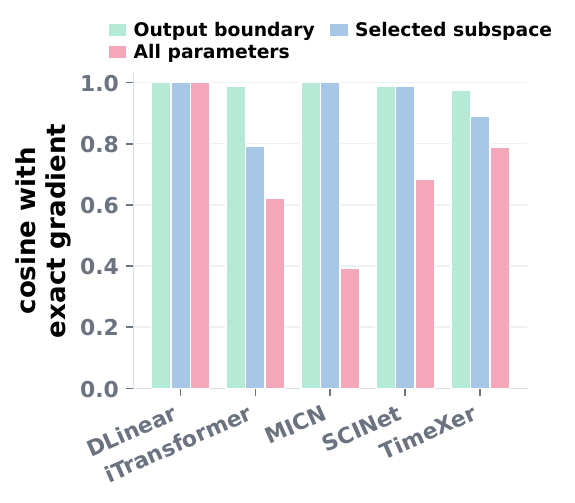}
\vspace{-6mm}
\caption{Cached-to-exact gradient cosine by backbone. Selected-slice cosine excludes zero proxies.}
\label{fig:mechanism_evidence}
\vspace{-4mm}
\end{wrapfigure}
We directly test the mechanism by comparing exact per-variable gradients with cache-reconstructed proxies across the main-result settings (Figure~\ref{fig:mechanism_evidence}). Proxies are nearly exact at the output boundary. Fidelity declines after variable mixing, but nonzero selected-slice proxies remain close. Appendix~\ref{app:mechanism} details the protocol and tests whether surgery improves first-order update behavior through conflict mass, which measures pairwise disagreement among reconstructed proxy rows, and descent violations, which count variable objectives pushed uphill by the update. PV-Surgery lowers violations in most runs without increasing conflict mass. The appendix also analyzes oracle-gap reduction, where \(23\) of \(29\) harmed variables close part of their oracle gap (Figure~\ref{fig:oracle_gap}). Appendix~\ref{app:qualitative} gives qualitative results.

\section{Conclusion}
\label{sec:conclusion}
Multivariate forecasters optimize a loss averaged across variables, leaving the effect of the shared update on each variable unclear. We find that pairwise gradient conflict is common and most variables underperform a full-input single-target oracle. But conflict frequency is weakly related to harm, and mean alignment does not reliably identify helpful partners. PV-Surgery addresses this problem with a single-backward optimizer-side transformation. The method constructs variable-wise gradient proxies, restricts surgery to a reconstruction-consistent subspace, pools variables conditionally, and corrects directions while preserving surgery-input norms. Across matched settings, it reduces MSE by \(3.61\%\) and MAE by \(2.93\%\) on average, with every module contributing to both metrics. Mechanism analyses show that selected proxies track variable-wise gradients and forecasts for harmed variables often improve. Together, the results support selective, reliability-aware intervention over uniform conflict removal.
PV-Surgery has two main limitations. The method raises median per-epoch cost to \(1.54\times\) the mean-loss baseline (Appendix~\ref{app:training_time}), and gains can weaken when proxy fidelity is low, selected layers mix variables, or variable relationships differ across datasets. Future work should reduce this overhead and improve proxy construction under stronger variable mixing and more diverse cross-variable structures without sacrificing useful sharing.

\subsection*{AI use statement}
In this work, we used generative AI tools to edit prose, translate draft passages for internal review, provide feedback on the method and experiments, assist with parts of the method implementation, analysis code, and plotting code, and help interpret experimental results. We did not use these tools to originate the research question or central research idea. We reviewed all AI-assisted text and code and take responsibility for the final text, claims, and artifacts.

\subsection*{Ethics statement}
The experiments use established public forecasting benchmarks and do not collect personal data or make individual-level decisions. Nevertheless, benchmark accuracy does not establish reliability under distribution shift or safety in operational use. 

\subsection*{Reproducibility statement}
Section~\ref{sec:method} and Algorithm~\ref{alg:pv_surgery} specify the loss scaling, cache construction, layer-selection rule, pooling and restoration operations, candidate comparison, and final-gradient assembly. Appendix~\ref{app:experimental_configuration} reports the forecasting protocol and optimization settings. Tables~\ref{tab:backbone_pv} and \ref{tab:backbone_pv_weather_ili} cover every one of the \(5\times7\times4=140\) main comparisons, while Table~\ref{tab:core_ablation} and Appendix~\ref{app:ablation_all_datasets} jointly cover all 140 matched ablation settings per variant. Except for the multi-seed analyses in Appendix~\ref{app:group_composition}, all reported runs use the single seed \(42\), as specified in Appendix~\ref{app:experimental_configuration}.

\bibliography{references}
\bibliographystyle{preprint_references}

\appendix
\section{Dataset Diagnostics}
\label{app:dataset_diagnostics}

Table~\ref{tab:dataset_diagnostics} summarizes the datasets and the diagnostic quantities used to contextualize PV-Surgery. In addition to common dataset metadata, we report conflict-oriented indicators tied to the optimization issue studied in this paper. The negative cosine ratio measures how often variable-wise gradients disagree, the mean cosine measures average alignment, and the harmed-variable fraction is the share of variables for which shared training performs worse than the full-input single-target oracle. Diagnostics are computed with iTransformer at prediction length \(96\), except for ILI where prediction length \(24\) is used.

\begin{table}[H]
\centering
\caption{Dataset statistics and conflict-oriented diagnostics. \(T\) is the number of timestamps and \(D\) is the number of target variables. Neg. cos. is the percentage of negative pairwise variable-gradient cosines and Mean cos. is their average. Both follow the protocol of Section~\ref{sec:diagnosis} and are computed after dropping the warmup half of the recorded checkpoints. Harmed vars. is the percentage of variables for which shared training is worse than the full-input single-target oracle on the held-out test split.}
\label{tab:dataset_diagnostics}
\vspace{1mm}
\setlength{\tabcolsep}{3.0pt}
\scriptsize
\resizebox{0.78\textwidth}{!}{%
\begin{tabular}{@{}llccccc@{}}
\toprule
Dataset & Frequency & \(T\) & \(D\) & Neg. cos. (\%) & Mean cos. & Harmed vars. (\%) \\
\midrule
ETTh1 & 1 hour & 17,420 & 7 & 30.6 & 0.178 & 14.3 \\
ETTh2 & 1 hour & 17,420 & 7 & 30.4 & 0.281 & 57.1 \\
ETTm1 & 15 min & 69,680 & 7 & 30.8 & 0.172 & 28.6 \\
ETTm2 & 15 min & 69,680 & 7 & 29.9 & 0.246 & 57.1 \\
Weather & 10 min & 52,696 & 21 & 37.5 & 0.133 & 81.0 \\
Exchange & 1 day & 7,588 & 8 & 39.3 & 0.148 & 25.0 \\
ILI & 1 week & 966 & 7 & 15.9 & 0.508 & 71.4 \\
\bottomrule
\end{tabular}%
}
\end{table}

\section{Gradient-Conflict Figures for All Datasets}
\label{app:gradient_conflict_all_datasets}

Figure~\ref{fig:gradient_conflict_structure} presents ETTh1 in the main text. Figures~\ref{fig:gradient_conflict_etth2_appendix}--\ref{fig:gradient_conflict_ili_appendix} use the same three-panel layout, consisting of the post-warmup cosine distribution and heatmaps from an early and a late checkpoint, for the remaining six benchmarks. ETTh1, ETTh2, ETTm1, ETTm2, Weather, and Exchange use prediction length \(96\), and ILI uses \(24\).

\paragraph{Conflict is intermittent.} No variable pair carries a negative cosine at every post-warmup checkpoint in any of the seven benchmarks. The most persistent pair is on ETTh2 and is negative at \(72\%\) of its checkpoints, while the median pair ranges from \(11\%\) to \(43\%\) across the seven benchmarks (Table~\ref{tab:conflict_persistence}). When each pair is averaged across its post-warmup checkpoints, the resulting cosines span \([-0.12,0.98]\). Thus, even the lowest-alignment pairs are only weakly opposed on average, and this average hides changes in sign over training. A static pair-level rule could therefore act after the relation has changed. To account for this behavior, the method in Section~\ref{sec:method} instead recomputes the relevant gradient relations from the current batch and reevaluates layer support at every step.

\begin{table}[H]
\centering
\caption{Persistence of variable-pair conflict over training. The Checkpoints column gives the number of post-warmup
analysis points, and the Pairs column gives the number of variable pairs in each benchmark. Persistence is the share of those
checkpoints at which a pair has a negative cosine, reported for the most persistent pair and for the
median pair. No entry reaches \(100\%\), so no pair conflicts throughout.}
\label{tab:conflict_persistence}
\vspace{1mm}
\setlength{\tabcolsep}{3.0pt}
\scriptsize
\resizebox{0.54\textwidth}{!}{%
\begin{tabular}{@{}lcccc@{}}
\toprule
\multirow{2}{*}{Dataset} & \multirow{2}{*}{Checkpoints} & \multirow{2}{*}{Pairs}
  & \multicolumn{2}{c}{Persistence (\%)} \\
\cmidrule(l){4-5}
 & & & Most persistent & Median \\
\midrule
ETTh1 & 46 & 21 & 50.0 & 28.3 \\
ETTh2 & 39 & 21 & 71.8 & 20.5 \\
ETTm1 & 48 & 21 & 50.0 & 35.4 \\
ETTm2 & 37 & 21 & 48.6 & 40.5 \\
Weather & 44 & 210 & 68.2 & 40.9 \\
Exchange & 38 & 28 & 55.3 & 43.4 \\
ILI & 150 & 21 & 37.3 & 11.3 \\
\bottomrule
\end{tabular}%
}
\end{table}

\newcommand{\gradientdatasetfigure}[5]{%
\begin{figure}[p]
\centering
\setlength{\tabcolsep}{1.5pt}
\begin{tabular}{@{}c@{\hspace{0.01\linewidth}}c@{\hspace{0.01\linewidth}}c@{}}
\includegraphics[width=0.360\linewidth]{#2/cosine_similarity_distribution.pdf}
&
\includegraphics[width=0.295\linewidth]{#2/cosine_similarity_heatmap_early.pdf}
&
\includegraphics[width=0.292\linewidth]{#2/cosine_similarity_heatmap_late.pdf}
\\[-1mm]
{\small (a) Post-warmup cosine distribution.}
&
{\small (b) Early checkpoint.}
&
{\small (c) Late checkpoint.}
\end{tabular}
\caption{A representative #1 run from the variable-wise gradient-cosine diagnostic at prediction length #4. (a) Distribution of the post-warmup pairwise cosines, of which \(#5\%\) fall below zero for this run. (b) An early checkpoint, which lies inside the discarded warmup half and is shown only to expose the change over training. (c) A late checkpoint. Both heatmaps show that conflict is structured across variable pairs rather than being uniform, so one summary number would hide which pairs disagree.}
\label{#3}
\end{figure}
}

\gradientdatasetfigure
{ETTh2}
{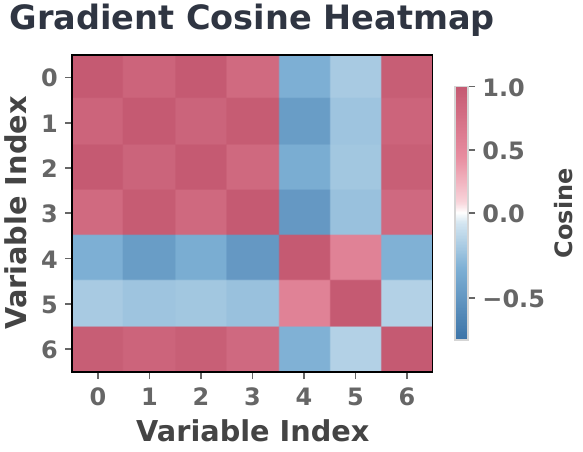}
{fig:gradient_conflict_etth2_appendix}
{96}
{30.4}

\gradientdatasetfigure
{ETTm1}
{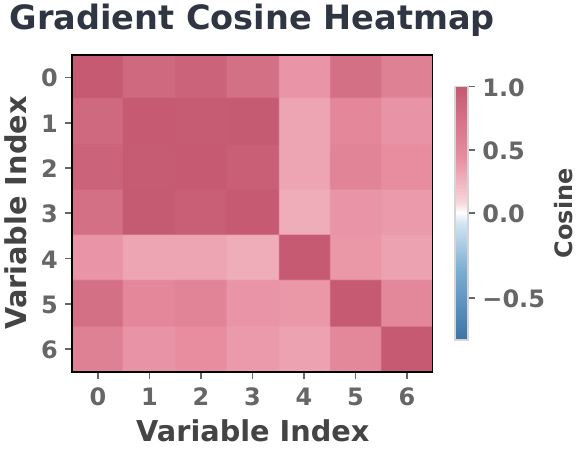}
{fig:gradient_conflict_ettm1_appendix}
{96}
{30.8}

\gradientdatasetfigure
{ETTm2}
{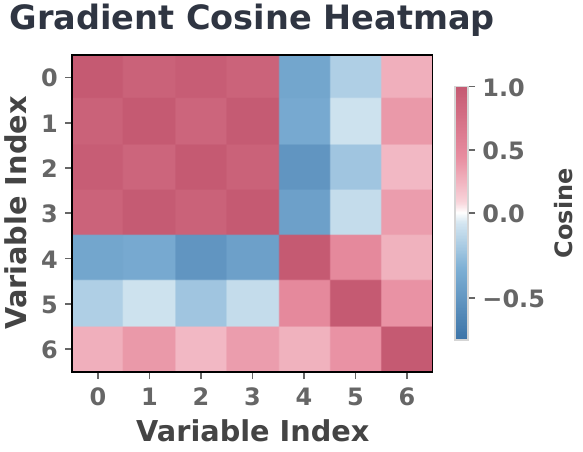}
{fig:gradient_conflict_ettm2_appendix}
{96}
{29.9}

\gradientdatasetfigure
{Weather}
{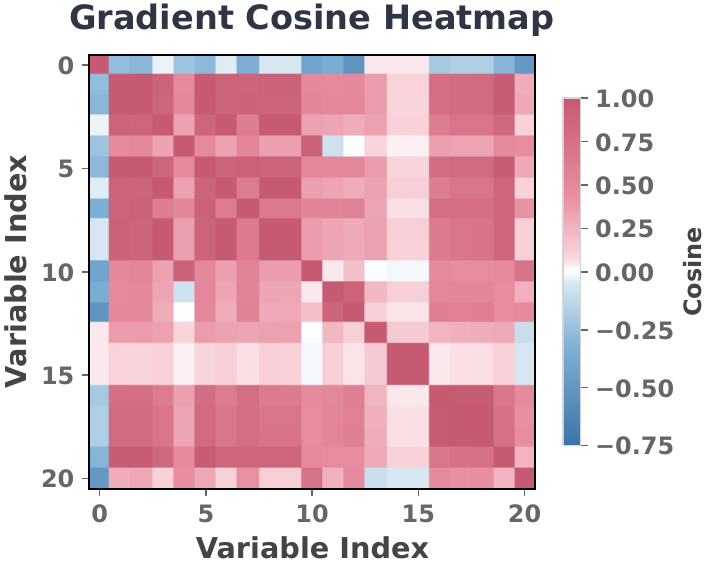}
{fig:gradient_conflict_weather_appendix}
{96}
{37.5}

\gradientdatasetfigure
{Exchange}
{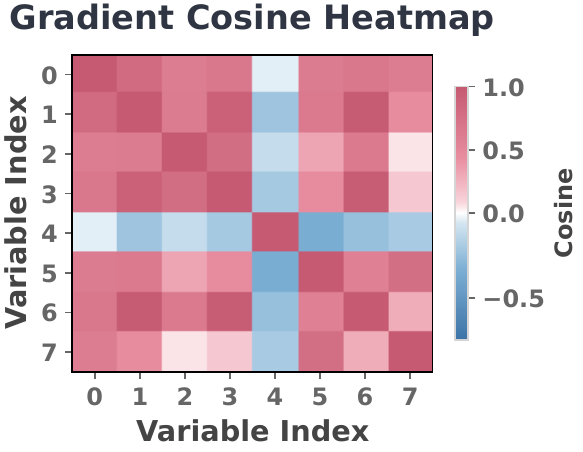}
{fig:gradient_conflict_exchange_appendix}
{96}
{39.3}

\gradientdatasetfigure
{ILI}
{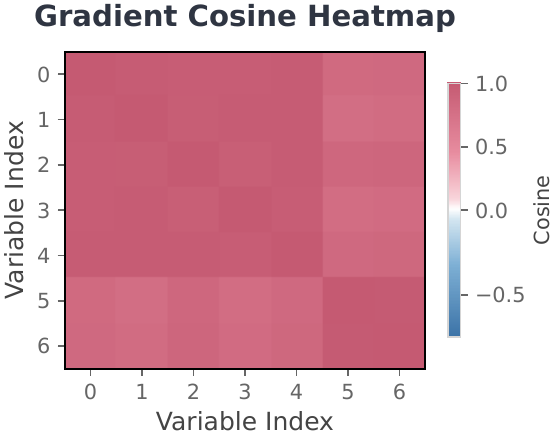}
{fig:gradient_conflict_ili_appendix}
{24}
{15.9}

\let\gradientdatasetfigure\relax

\section{Group-Composition Diagnostics for All Datasets}
\label{app:group_composition}

\paragraph{Group-composition protocol.}
For each reported target variable, we rank the remaining variables by their post-warmup mean pairwise gradient cosine in the baseline run. For each total group size \(k\in\{2,3,4\}\), we train one model with the \(k-1\) highest-ranked partners and another with the \(k-1\) lowest-ranked partners. The two partner sets are disjoint in every reported cell. Each model still receives every variable as input. Only the supervised loss is restricted to the target and its selected partners, and evaluation uses the target's held-out per-variable test MSE. The \(162\) target-and-size cells cover every target in the four ETT benchmarks, Exchange, and ILI, plus \(11\) of Weather's \(21\) targets. Higher- and lower-alignment conditions use matched replicate seeds, three per cell, which totals \(972\) group-training runs. We report the seed-averaged signed contrast \(100(\mathrm{MSE}_{\mathrm{high}}-\mathrm{MSE}_{\mathrm{low}})/\mathrm{MSE}_{\mathrm{low}}\). Negative values favor the higher-alignment set. A cell counts as a higher-alignment win when its signed contrast is negative. The median signed contrast retains each cell's direction, whereas the median absolute within-cell contrast removes the sign and measures the size of the partner-set difference. For every fixed dataset, target, group size, and partner condition, we express the standard deviation of test MSE across the three seeds as a percentage of their mean. To express seed variation on the scale of a two-run difference, we multiply this coefficient of variation by \(\sqrt{2}\), which is the standard deviation of the difference under an equal-variance independence approximation. We use it as a descriptive reference rather than as a paired uncertainty estimate. The median over the \(324\) fixed partner conditions is \(1.1\%\).

\begin{figure}[h]
\centering
\includegraphics[width=0.95\linewidth]{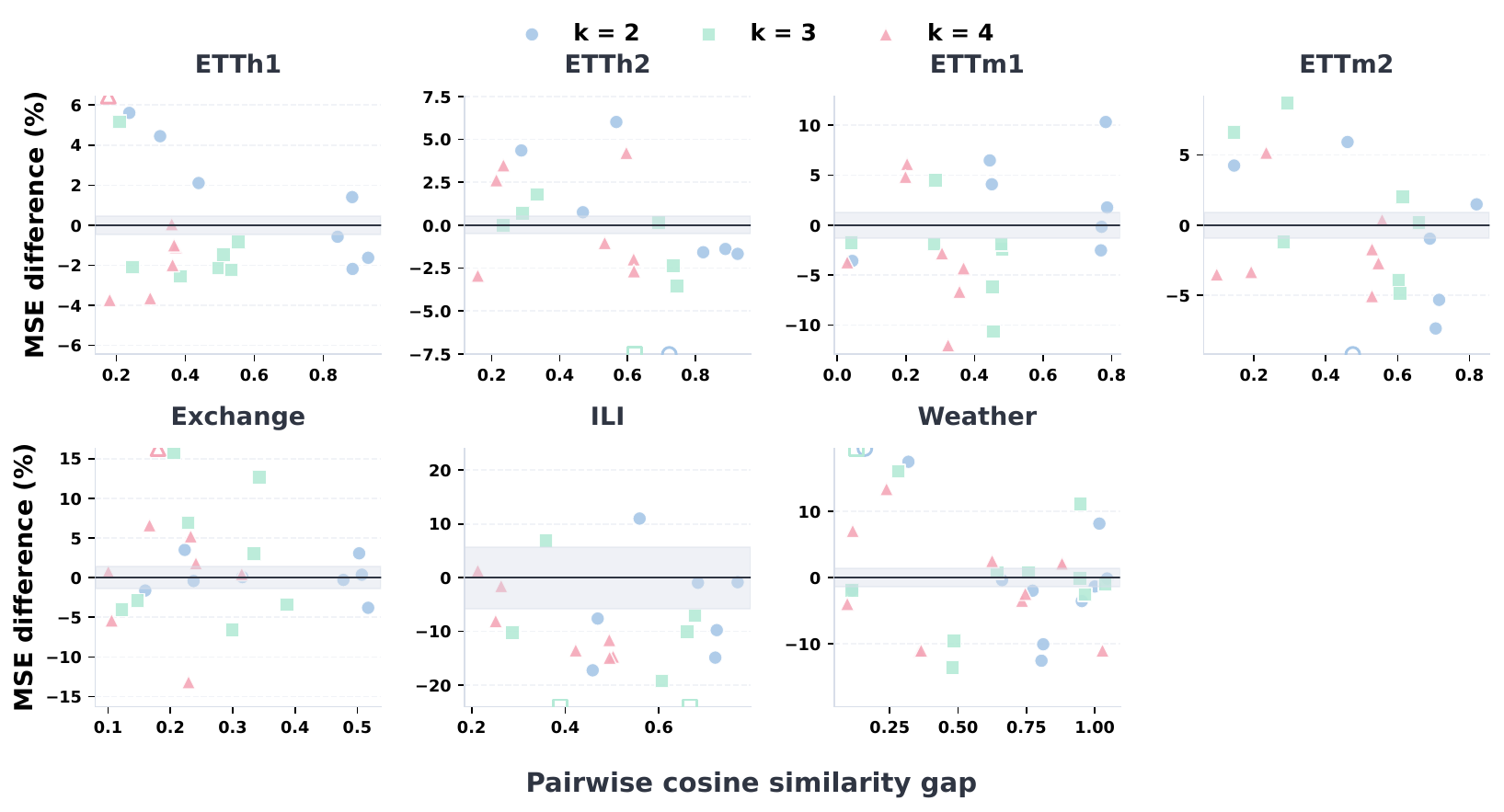}
\caption{Figure~\ref{fig:harm_subset_diagnostics}(b) split by dataset, with the three group sizes marked
separately. Each plot holds the \(21\) to \(33\) target-and-size cells of one benchmark.}
\label{fig:group_gap_scatter_by_dataset}
\end{figure}

\begin{figure}[h]
\centering
\includegraphics[width=0.95\linewidth]{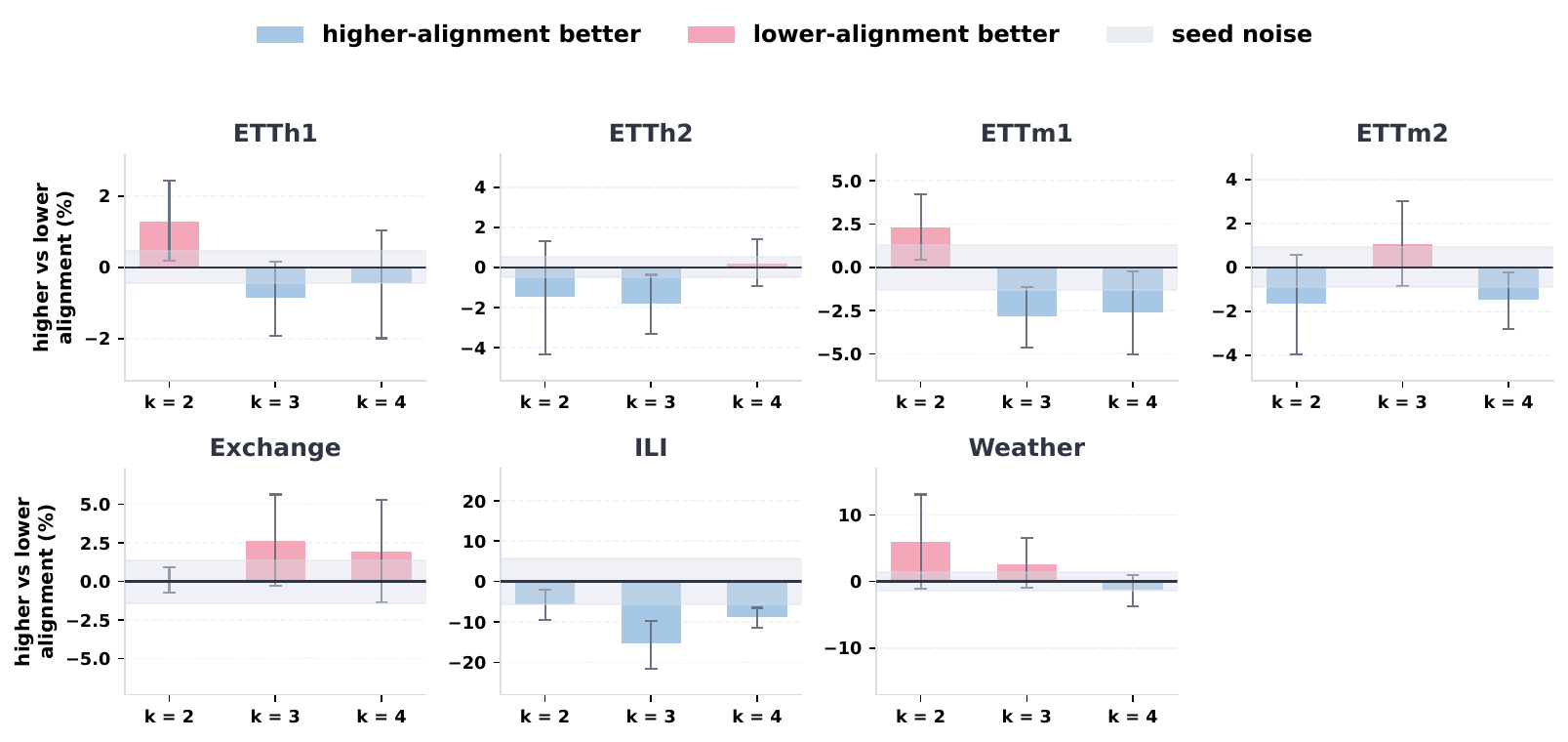}
\caption{Percentage difference between the higher- and lower-alignment partner groups' test MSE, taken
relative to the lower-alignment group and averaged over the targets of each benchmark. Negative means the
higher-alignment group did better. Vertical ranges differ across panels. The grey band is
that benchmark's seed noise and the error bars are the standard error across its targets.}
\label{fig:group_gaps_all}
\end{figure}

\begin{figure}[h]
\centering
\includegraphics[width=0.7\linewidth]{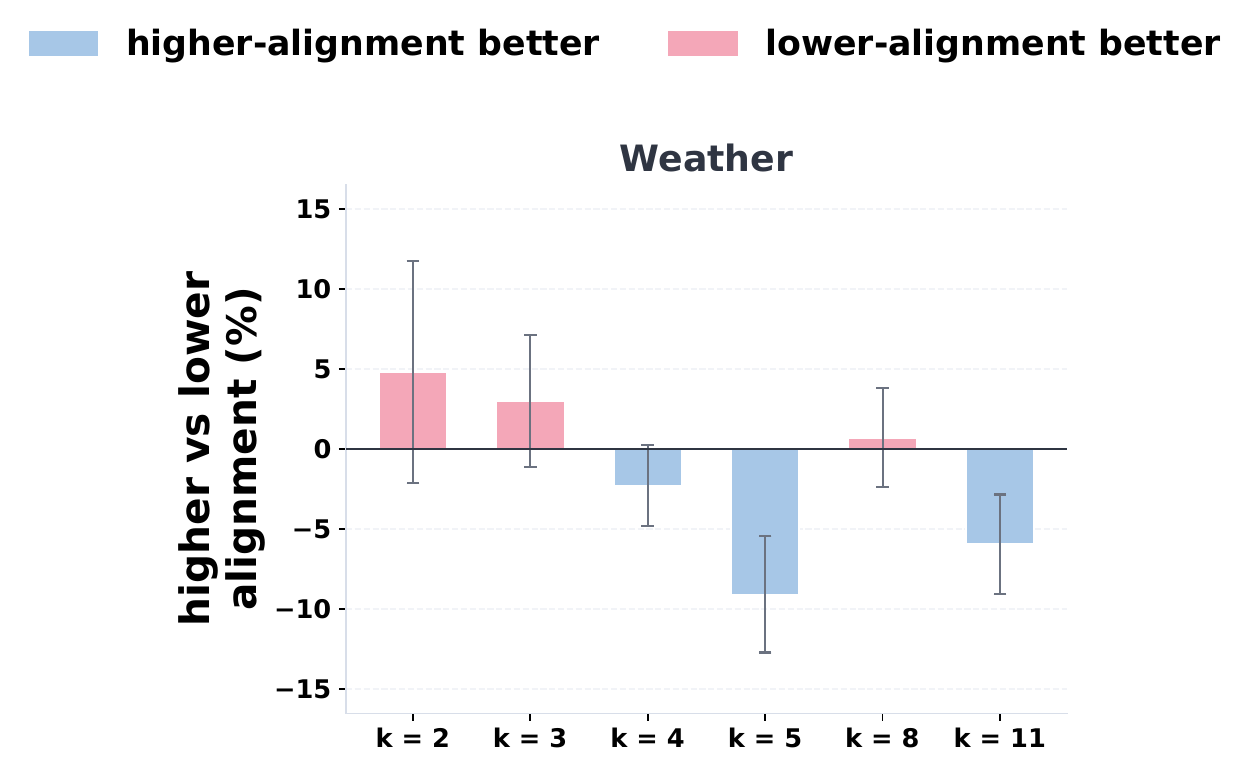}
\caption{Weather at the six tested group sizes \(k\in\{2,3,4,5,8,11\}\), using seed \(42\). Bars show the mean higher- versus lower-alignment MSE contrast across \(11\) targets, and error bars show the standard error across targets. Negative values favor the higher-alignment group. The contrast does not change monotonically with group size.}
\label{fig:weather_large_k}
\end{figure}

Figure~\ref{fig:group_gap_scatter_by_dataset} splits Figure~\ref{fig:harm_subset_diagnostics}(b) by
dataset, and Figure~\ref{fig:group_gaps_all} reports the higher- minus lower-alignment contrast for every
benchmark and group size. Higher-alignment sets win \(100\) of the \(162\) cells, and the median signed contrast across cells is \(-1.4\%\). A two-sided one-sample \(t\)-test over the seven benchmark-level mean contrasts does not reject a zero mean (\(p=0.45\)), although five of the seven means are negative. Excluding ILI shifts the pooled contrast from \(-0.9\%\) to \(+0.4\%\). The median absolute within-cell contrast is \(3.5\%\), which exceeds the \(1.1\%\) descriptive seed-variation reference above, so partner choice can materially change target error even though mean alignment does not reliably identify the better set. Across all \(162\) cells, Spearman's correlation between the alignment gap and MSE difference is \(-0.19\). The arithmetic mean of the seven within-benchmark correlations is \(-0.16\), and a two-sided one-sample \(t\)-test of those correlations against zero gives \(p=0.07\).
Figure~\ref{fig:weather_large_k} reports the seed-\(42\) Weather extension at the six tested group sizes \(k\in\{2,3,4,5,8,11\}\), including the disjoint-set limit \(k=11\) for \(D=21\).

\FloatBarrier

\section{PV-Surgery Training Procedure}
\label{app:training_procedure}

Algorithm~\ref{alg:pv_surgery} makes the four stages in Figure~\ref{fig:pv_surgery_overview} explicit for one training batch. The reference gradient and output-cache signals come from the same single backward pass on the summed variable loss. Because no rescaling is applied to this sum loss, its raw reference gradient satisfies \(\|g_0\|_2=D\|g_{\mathrm{mean}}\|_2\). Baseline and PV-Surgery runs use the same AdamW hyperparameters. The remaining operations transform cached gradient matrices and assemble the selected-slice correction. Let \(\mathcal H_{\mathrm{run}}\subseteq\mathcal H\) denote the layers whose proxy rows materialize for the current batch. We construct their observed proxy matrices and set \(G_\ell=0_{D\times N_\ell}\) for \(\ell\in\mathcal H\setminus\mathcal H_{\mathrm{run}}\). Layer scoring and final assembly use this zero-extended collection over \(\mathcal H\), while parameters outside the hooked set retain their reference-gradient values.

\begin{algorithm}[H]
\caption{PV-Surgery Training}
\label{alg:pv_surgery}
\footnotesize
\begin{algorithmic}
\STATE \textbf{Input:} batch \((X,Y)\), \(D\) variables, backbone \(f_\theta\),
       cache-compatible hooked layers \(\mathcal{H}\)
\STATE \textbf{Output:} optimizer gradient \(g_{\mathrm{final}}\)
\STATE \textit{// Step 1: Output-cache variable-wise gradient proxy}
\STATE \(\hat Y\leftarrow f_\theta(X)\)
       \COMMENT{hooks cache \(\{Z_\ell\}_{\ell\in\mathcal H}\)}
\STATE \(L_\Sigma\leftarrow\sum_{d=1}^{D}L_d(\hat Y_{:,:,d},Y_{:,:,d})\)
       \quad \(g_0\leftarrow\operatorname{Backward}(L_\Sigma)\)
       \COMMENT{Eqs.~\ref{eq:losses},~\ref{eq:reference_gradient}}
\STATE \(\mathcal{H}_{\mathrm{run}}\leftarrow
       \{\ell\in\mathcal{H}:Z_\ell\text{ and }E_\ell\text{ materialize for this batch}\}\)
\STATE \(G_\ell[d,\cdot]\leftarrow
       \operatorname{concat}\!\left(\operatorname{vec}(E_{\ell,d}^{\top}Z_{\ell,d}),
       E_{\ell,d}^{\top}\mathbf{1}\right)\) for \(\ell\in\mathcal H_{\mathrm{run}}\)
       \COMMENT{Eqs.~\ref{eq:cache_gradients},~\ref{eq:cache_matrix}}
\STATE \(G_\ell\leftarrow0_{D\times N_\ell}\) for
       \(\ell\in\mathcal H\setminus\mathcal H_{\mathrm{run}}\)
\STATE \textit{// Step 2: Reliability-aware layer selection}
\STATE \((e_\ell,r_\ell,q_\ell)\leftarrow
       \operatorname{LayerScores}(G_\ell,g_{0,\ell},g_0)\) for \(\ell\in\mathcal H\)
       \COMMENT{Eq.~\ref{eq:layer_reliability}}
\STATE \(\mathcal H_{\mathrm{elig}}\leftarrow
       \{\ell\in\mathcal H:\ell\text{ is not protected}\}\)
\STATE \(\rho\leftarrow\operatorname{median}
       \{r_\ell:\ell\in\mathcal H_{\mathrm{elig}}\}\)
\STATE \(\mathcal H_{\mathrm{gate}}\leftarrow
       \{\ell\in\mathcal H_{\mathrm{elig}}:
       r_\ell\geq\rho\text{ or }\ell=\ell_{\mathrm{out}}\}\)
\STATE \(\mathcal A\leftarrow
       \{\ell\in\mathcal H_{\mathrm{gate}}:q_\ell>0\}\)
\IF{\(\mathcal A\neq\varnothing\)}
    \STATE \(\pi_\ell\leftarrow q_\ell/\sum_{j\in\mathcal A}q_j\)
    \quad \(K_{\mathrm{eff}}\leftarrow
    \lceil\exp(-\sum_{\ell\in\mathcal A}\pi_\ell\log\pi_\ell)\rceil\)
    \STATE \(\mathcal T\leftarrow
    \operatorname{Top}_{K_{\mathrm{eff}}}(\mathcal A,q)
    \cup(\mathcal A\cap\{\ell_{\mathrm{out}}\})\)
    \STATE \(\mathcal S\leftarrow\{\ell\in\mathcal T:e_\ell\leq1\}\)
    \COMMENT{Eq.~\ref{eq:effective_coverage}}
\ELSE
    \STATE \(\mathcal S\leftarrow\varnothing\)
\ENDIF
\STATE \(\mathcal S\leftarrow
       \operatorname{OutputFallback}(\mathcal S,\mathcal H_{\mathrm{elig}},\ell_{\mathrm{out}})\)
\STATE \(G_{\mathcal S}\leftarrow
       \operatorname{concat}_{\ell\in\mathcal S}G_\ell\)
       \COMMENT{Eq.~\ref{eq:selected_gradients}}
\STATE \textit{// Step 3: Conditional variable-gradient pooling}
\STATE \(G_{\mathrm{pre}}\leftarrow G_{\mathcal{S}}\)
       \quad \(g_{\mathrm{ref}}\leftarrow g_{0,\mathcal S}\)
\STATE \(\mathcal V\leftarrow\{d:G_{\mathrm{pre}}[d]\text{ is finite and }
       \|G_{\mathrm{pre}}[d]\|_2>\epsilon\}\)
\STATE \((a,M_{\mathrm{pool}},\mathrm{active})\leftarrow
       \operatorname{ConditionalAnchorConflict}
       (G_{\mathrm{pre}},g_{\mathrm{ref}};\operatorname{valid}=\mathcal V,
       \operatorname{order}=(a_d,d))\)
       \COMMENT{Eq.~\ref{eq:pool_reference}}
\IF{\(\mathrm{active}\)}
    \STATE \(G_{\mathrm{pool}}\leftarrow M_{\mathrm{pool}}G_{\mathrm{pre}}\)
    \COMMENT{Eq.~\ref{eq:sum_pool}}
\ENDIF
\STATE \textit{// Step 4: Magnitude-preserving common-direction surgery}
\STATE \(G_{\mathrm{in}}^{(u)}\leftarrow G_{\mathrm{pre}}\), and
       \(G_{\mathrm{in}}^{(p)}\leftarrow G_{\mathrm{pool}}\) when pooling is active
\FOR{each available candidate \(c\in\{u,p\}\)}
    \STATE \(\widetilde G_{\mathrm{in}}^{(c)}\leftarrow
    \operatorname{CommonDirection}_{\mathrm{preserve}}
    (G_{\mathrm{in}}^{(c)})\)
    \COMMENT{Eqs.~\ref{eq:magnitude_direction},~\ref{eq:common_direction}}
\ENDFOR
\STATE \(G_u^{\mathrm{post}}\leftarrow\widetilde G_{\mathrm{in}}^{(u)}\)
\IF{\(\mathrm{active}\)}
    \STATE \(G_p^{\mathrm{post}}\leftarrow
    \operatorname{PoolRestore}(G_{\mathrm{pre}},M_{\mathrm{pool}},
    \widetilde G_{\mathrm{in}}^{(p)})\)
    \COMMENT{Eq.~\ref{eq:pool_restore}}
\ELSE
    \STATE \(G_p^{\mathrm{post}}\leftarrow G_{\mathrm{pre}}\)
\ENDIF
\STATE \(G_{\mathrm{post}}\leftarrow
       \operatorname{DescentLexicographicSelect}^{u\text{-tie}}
       (G_{\mathrm{pre}},G_u^{\mathrm{post}},G_p^{\mathrm{post}};\mathcal V)\)
       \COMMENT{Eq.~\ref{eq:safe_selector}}
\STATE \(\Delta_{\mathcal{S}}\leftarrow
       \sum_dG_{\mathrm{post}}[d,\cdot]-\sum_dG_{\mathrm{pre}}[d,\cdot]\)
       \COMMENT{Eq.~\ref{eq:surgery_delta}}
\STATE \(g_{\mathrm{final}}\leftarrow
       \operatorname{Assemble}(g_0,\{G_\ell\}_{\ell\in\mathcal{H}},
       \mathcal{S},\Delta_{\mathcal{S}})\)
       \COMMENT{Eq.~\ref{eq:gradient_assembly}}
\end{algorithmic}
\end{algorithm}

\paragraph{Pooling and candidate-selection conventions.}
Cosine and pairwise-conflict tests use only rows that are finite and have norm greater than the numerical tolerance. Cosines within a machine-scale tolerance of zero are treated as zero. A zero or non-finite row is assigned \(a_d=0\), remains a singleton objective, and does not activate pooling. A zero or non-finite reference yields the identity pooling plan. When pooling is active, conflict rows are processed in ascending reference-alignment order, with the variable index breaking ties. Each row joins the existing conflict pool with the largest cosine only when that cosine is positive. Otherwise, it starts a new pool. Candidate scoring uses the same valid pre-surgery rows. A candidate with a zero or non-finite aggregate is invalid. If only one candidate is valid, it is selected. If both are invalid or numerically tied, the unpooled candidate is retained.

\paragraph{Numerical tolerance for common-direction surgery.}
Equation~\ref{eq:common_direction} defines the ideal map using an exact zero test. In the implementation, we use the first branch only when \(\|\bar g^{(c)}\|_2>\epsilon\) and otherwise retain the input, where \(\epsilon>0\) is a small numerical tolerance. This avoids normalizing a near-zero mean direction.

\paragraph{Proxy-supported hooked-block updates.}
PV-Surgery treats the hooked parameter blocks as the region in which an update should remain attributable to variable-indexed signals. When \(\ell\in\mathcal H_{\mathrm{run}}\), every hooked-block contribution is constructed from observed proxy rows and can therefore pass through the same reconstruction, pooling, and direction checks as the other variable-wise inputs. When \(\ell\notin\mathcal H_{\mathrm{run}}\), the single-backward slice \(g_{0,\ell}\) is still available, but it contains only the aggregate of the unresolved variable contributions. Substituting that slice would reintroduce an update whose variable-wise sources cannot be inspected by the method. We therefore impose a blockwise evidence-support constraint that permits a nonzero hooked-block data-gradient contribution only when its variable-indexed rows materialize. Under this constraint, zero is the only admissible current data-gradient contribution on an unsupported block and is also the minimum-norm choice. This is a conservative abstention rule rather than a claim that a zero data-gradient contribution is a better descent direction than \(g_{0,\ell}\). Parameters outside \(\mathcal H\) are not assigned to the variable-aware region and continue to use \(g_0\). A zero-extended block remains zero even if it enters \(\mathcal S\), because its columns are zero for every surgery input and the norm-preserving direction operator cannot create support on those columns.

This zero data-gradient contribution is injected as an explicit zero loss-gradient tensor. It is therefore not the same as removing the parameters from AdamW or setting their gradients to \texttt{None}. AdamW still advances their optimizer state and applies decoupled weight decay. If a block has accumulated momentum from earlier nonzero updates, that state can continue to affect later parameter values while it decays. The rule should consequently be understood as withholding the current data-gradient update on an unsupported hooked block, not as permanently freezing the block.

\paragraph{Layer-selection thresholds.}
Although the cache supports explicit variable axes and supported batch-folded layouts, the configured protection rule excludes patch and value embedding parameters from direction-surgery selection to avoid changing the shared input representation before downstream variable mixing. If registered as hooked blocks, these parameters still follow the proxy-sum assembly rule above. Excluding them from \(\mathcal S\) prevents direction-surgery correction on those slices but does not restore \(g_0\). Their data-gradient contribution is the sum of the variable-indexed proxy rows when those rows materialize for the current batch, and zero when no such variable-axis proxy materializes. Among the remaining candidates the reliability gate keeps every non-output layer whose \(r_\ell\) reaches the median over the eligible hooked layers, and the output layer is exempt from this gate. Equation~\ref{eq:effective_coverage} is evaluated on \(q_\ell\), which also orders the layers before the \(K_{\mathrm{eff}}\) cut. A layer that survives the cut is eligible when its relative reconstruction error satisfies \(e_\ell\le1\), equivalently \(r_\ell\ge1/2\). This means that its proxy-sum error does not exceed \(\|g_{0,\ell}\|_2+\epsilon\), which marks, up to numerical tolerance, the boundary at which the reconstruction error begins to exceed the magnitude of the gradient being reconstructed. If nothing is left, PV-Surgery falls back to the output layer.

\paragraph{Properties of effective coverage.}
The unrounded quantity \(e^{\mathcal{E}_{\mathrm{layer}}}\) in Equation~\ref{eq:effective_coverage} is the exponential of the Shannon entropy of \(\pi\) and therefore the order-one Hill number \citep{hill1973diversity} of the normalized slice-norm distribution. Because \(\pi_\ell=q_\ell/\sum_{j\in\mathcal A}q_j\), it represents the fraction of the total gated slice norm carried by layer \(\ell\). The entropy \(\mathcal{E}_{\mathrm{layer}}\) measures how evenly these fractions are distributed, and exponentiation converts this log-scale measure into an effective number of contributing layers. This number is one when a single layer carries all of the slice norm and \(m\) when the norm is uniform over \(m\) layers. An intermediate value is the number of equally contributing layers that would produce the same concentration as the observed distribution. For example, \(\pi=(0.8,0.1,0.1)\) gives \(e^{\mathcal{E}_{\mathrm{layer}}}\approx1.89\) and therefore \(K_{\mathrm{eff}}=2\) after taking the ceiling. The selector uses this ceiling as an entropy-derived cutoff for the largest-norm layers, while the reliability and reconstruction criteria still determine which layers are eligible. The Hill number is invariant to a common rescaling of the layer slice norms, so a backbone whose layer gradients are uniformly larger does not receive a wider support. It is recomputed at each step, allowing the support to contract when one layer dominates and to widen when the slice-norm distribution becomes diffuse. A layer whose share vanishes contributes \(\pi_\ell\log\pi_\ell\to0\) and has a vanishing effect on the effective count. The choice of order also matters. Order zero counts every gated layer equally, while order two emphasizes dominant layers more strongly. Order one retains the Shannon weighting in which each entropy contribution is weighted by its normalized slice-norm share \(\pi_\ell\). We avoid a fixed top-\(k\) because the number of hooked layers differs by backbone, from one to seventeen in our experiments, so a new backbone would otherwise need its own value.

\section{Detailed Experimental Configuration}
\label{app:experimental_configuration}

\paragraph{Forecasting protocol.}
For ETTh1, ETTh2, ETTm1, ETTm2, Weather, and Exchange, we use an input length of \(96\), a decoder-label length of \(48\), and prediction lengths \(\{96,192,336,720\}\). ILI uses input and decoder-label lengths of \(36\) and \(18\). Its main, ablation, mechanism, and objective-comparison runs use prediction lengths \(\{24,36,48,60\}\), while the diagnostic, oracle, and partner-group analyses in Section~\ref{sec:diagnosis} use \(24\). Every method uses the chronological train, validation, and test splits provided by the forecasting framework. Training windows are not shuffled because Selective Learning tracks residual statistics across consecutive windows \citep{fu2025selective}.

\paragraph{Optimization.}
Unless otherwise noted for the multi-seed analyses in Appendix~\ref{app:group_composition}, each reported run uses seed \(42\). Training uses a batch size of \(64\) for at most \(30\) epochs and stops after seven consecutive validation epochs without improvement. We use AdamW with a fixed learning rate of \(10^{-4}\), weight decay \(5\times10^{-4}\), and a gradient-norm limit of \(1\).
\paragraph{Specialized-objective settings.}
In our objective comparison, TILDE-Q uses \(\alpha=0.5\) and \(\gamma=0\), which removes the amplitude term. FreDF weights the temporal MSE and frequency-domain terms by \(0.5\) each. Its frequency MAE is the mean modulus of the complex RFFT difference. PSLoss uses \(\lambda=3\), limits the adaptive patch length to \(24\), and enables Gradient-based Dynamic Weighting with automatic target-layer selection and an equal-weight fallback. Selective Learning uses \((r_u,r_a)=(0.3,0.3)\), \((0.1,0.6)\), \((0.2,0.2)\), \((0.2,0.5)\), \((0.1,0.2)\), \((\mathrm{off},0.9)\), and \((0.1,0.1)\) for ETTh1, ETTh2, ETTm1, ETTm2, Weather, Exchange, and ILI, respectively. Its anomaly mask uses a pretrained DLinear estimator for each dataset and horizon. TILDE-Q, FreDF, and PSLoss select checkpoints using their configured validation objectives, whereas Selective Learning uses validation MSE. Early stopping monitors validation MSE with patience \(7\) for all four methods.
\paragraph{Diagnostic logging.}
The shared-model diagnostics in Section~\ref{sec:diagnosis} record gradients every \(\max(1,\lfloor30S/200\rfloor)\) optimizer steps, where \(S\) is the number of steps in one epoch. This interval targets approximately \(200\) snapshots over a complete \(30\)-epoch run. Early stopping reduces the total in most benchmarks, while ILI has an interval of one step because each epoch is short. Before removing the warmup half, the runs contain \(74\) to \(300\) snapshots. The post-warmup counts reported in Appendix~\ref{app:gradient_conflict_all_datasets} are therefore \(37\) to \(150\).

\paragraph{Backbone hyperparameters.}
DLinear uses a moving-average window of \(25\) and shared rather than variable-specific trend and seasonal heads. iTransformer uses \(d_{\mathrm{model}}=512\), \(d_{\mathrm{ff}}=512\), \(8\) attention heads, \(3\) encoder layers, GELU activations, and dropout \(0.1\). TimeXer uses \(d_{\mathrm{model}}=256\), \(d_{\mathrm{ff}}=512\), \(8\) attention heads, \(2\) encoder layers, patch length \(16\), GELU activations, dropout \(0.1\), and input normalization. MICN uses \(d_{\mathrm{model}}=512\), \(d_{\mathrm{ff}}=2048\), \(8\) heads, \(1\) decoder layer, convolution kernels \(\{12,16\}\), and dropout \(0.05\). SCINet uses one stack and dropout \(0.1\). Standard MSE training and PV-Surgery use the same backbone settings.

\paragraph{Hardware.}
We run the experiments on NVIDIA TITAN RTX GPUs with \(24\) GB of memory.

\paragraph{Metric definitions.}
We compute MSE and MAE on standardized forecasts and targets without applying the inverse transformation. Let \(N_{\mathrm{test}}\) denote the number of test windows. The reported metrics are
\begin{equation}
\label{eq:test_metrics}
\begin{aligned}
    \mathrm{MSE}&=\frac{1}{N_{\mathrm{test}}HD}
    \sum_{n=1}^{N_{\mathrm{test}}}\sum_{h=1}^{H}\sum_{d=1}^{D}
    (\hat Y_{n,h,d}-Y_{n,h,d})^2,\\
    \mathrm{MAE}&=\frac{1}{N_{\mathrm{test}}HD}
    \sum_{n=1}^{N_{\mathrm{test}}}\sum_{h=1}^{H}\sum_{d=1}^{D}
    |\hat Y_{n,h,d}-Y_{n,h,d}|.
\end{aligned}
\end{equation}

\section{Backbone Results for Weather and ILI}
\label{app:backbone_weather_ili}

The page limit restricts Table~\ref{tab:backbone_pv} to five of the seven benchmarks. Table~\ref{tab:backbone_pv_weather_ili} reports the omitted Weather and ILI results, completing the same comparison over five backbones, seven datasets, and four prediction lengths. Each MSE and +PV pair uses the same data split, backbone configuration, and optimization protocol.

\begin{table}[H]
\centering
\caption{Backbone comparison between standard MSE training and PV-Surgery on Weather and ILI.
Within each backbone and metric, the better value between MSE and +PV is bolded using unrounded
scores. The protocol and horizons follow Table~\ref{tab:backbone_pv}.}
\label{tab:backbone_pv_weather_ili}
\setlength{\tabcolsep}{2.4pt}
\renewcommand{\arraystretch}{0.96}
\scriptsize
\resizebox{\textwidth}{!}{%
\begin{tabular}{@{}c|c|rrrr|rrrr|rrrr|rrrr|rrrr@{}}
\toprule
\multicolumn{2}{c|}{Models} & \multicolumn{4}{c|}{DLinear} & \multicolumn{4}{c|}{iTransformer} & \multicolumn{4}{c|}{MICN} & \multicolumn{4}{c|}{SCINet} & \multicolumn{4}{c}{TimeXer} \\
\midrule
\multicolumn{2}{c|}{Training setup} & \multicolumn{2}{c}{MSE} & \multicolumn{2}{c|}{+PV} & \multicolumn{2}{c}{MSE} & \multicolumn{2}{c|}{+PV} & \multicolumn{2}{c}{MSE} & \multicolumn{2}{c|}{+PV} & \multicolumn{2}{c}{MSE} & \multicolumn{2}{c|}{+PV} & \multicolumn{2}{c}{MSE} & \multicolumn{2}{c}{+PV} \\
\midrule
\multicolumn{2}{c|}{Metric} & \multicolumn{1}{c}{MSE} & \multicolumn{1}{c}{MAE} & \multicolumn{1}{c}{MSE} & \multicolumn{1}{c|}{MAE} & \multicolumn{1}{c}{MSE} & \multicolumn{1}{c}{MAE} & \multicolumn{1}{c}{MSE} & \multicolumn{1}{c|}{MAE} & \multicolumn{1}{c}{MSE} & \multicolumn{1}{c}{MAE} & \multicolumn{1}{c}{MSE} & \multicolumn{1}{c|}{MAE} & \multicolumn{1}{c}{MSE} & \multicolumn{1}{c}{MAE} & \multicolumn{1}{c}{MSE} & \multicolumn{1}{c|}{MAE} & \multicolumn{1}{c}{MSE} & \multicolumn{1}{c}{MAE} & \multicolumn{1}{c}{MSE} & \multicolumn{1}{c}{MAE} \\
\midrule
\multirow{5}{*}{\textbf{Weather}} & 96 & \textbf{0.203} & 0.266 & 0.203 & \textbf{0.245} & 0.183 & 0.222 & \textbf{0.182} & \textbf{0.222} & 0.398 & 0.442 & \textbf{0.318} & \textbf{0.383} & \textbf{0.166} & \textbf{0.212} & 0.170 & 0.215 & \textbf{0.164} & \textbf{0.208} & 0.180 & 0.223 \\
 & 192 & 0.241 & 0.302 & \textbf{0.241} & \textbf{0.292} & 0.231 & \textbf{0.263} & 0.231 & 0.265 & \textbf{0.403} & \textbf{0.447} & 0.435 & 0.460 & 0.218 & 0.259 & \textbf{0.218} & \textbf{0.257} & \textbf{0.208} & \textbf{0.251} & 0.233 & 0.269 \\
 & 336 & 0.286 & 0.338 & \textbf{0.283} & \textbf{0.325} & 0.285 & 0.300 & \textbf{0.281} & \textbf{0.299} & \textbf{0.380} & \textbf{0.422} & 0.436 & 0.468 & 0.280 & 0.303 & \textbf{0.278} & \textbf{0.300} & \textbf{0.261} & \textbf{0.290} & 0.283 & 0.302 \\
 & 720 & 0.349 & 0.385 & \textbf{0.345} & \textbf{0.368} & 0.358 & 0.349 & \textbf{0.353} & \textbf{0.345} & 0.496 & 0.501 & \textbf{0.479} & \textbf{0.493} & 0.361 & 0.356 & \textbf{0.357} & \textbf{0.352} & \textbf{0.340} & \textbf{0.342} & 0.356 & 0.350 \\
\cmidrule(lr){2-22}
 & Avg & 0.270 & 0.323 & \textbf{0.268} & \textbf{0.308} & 0.264 & 0.283 & \textbf{0.262} & \textbf{0.283} & 0.419 & 0.453 & \textbf{0.417} & \textbf{0.451} & 0.256 & 0.282 & \textbf{0.256} & \textbf{0.281} & \textbf{0.243} & \textbf{0.273} & 0.263 & 0.286 \\
\midrule
\multirow{5}{*}{\textbf{ILI}} & 24 & 3.828 & 1.464 & \textbf{3.590} & \textbf{1.414} & \textbf{2.018} & \textbf{0.878} & 2.190 & 0.905 & \textbf{3.169} & \textbf{1.237} & 3.190 & 1.243 & 3.946 & 1.405 & \textbf{3.945} & \textbf{1.400} & 2.798 & \textbf{0.990} & \textbf{2.723} & 1.047 \\
 & 36 & 3.801 & 1.435 & \textbf{3.490} & \textbf{1.358} & \textbf{2.173} & 0.945 & 2.293 & \textbf{0.943} & 2.996 & 1.228 & \textbf{2.985} & \textbf{1.220} & 3.970 & 1.417 & \textbf{3.778} & \textbf{1.369} & \textbf{2.204} & \textbf{0.931} & 2.619 & 1.056 \\
 & 48 & 3.808 & 1.420 & \textbf{3.446} & \textbf{1.330} & \textbf{2.099} & \textbf{0.940} & 2.283 & 0.941 & 3.269 & 1.284 & \textbf{3.228} & \textbf{1.272} & 3.977 & 1.434 & \textbf{3.698} & \textbf{1.371} & \textbf{2.250} & \textbf{0.954} & 2.542 & 1.041 \\
 & 60 & 4.175 & 1.446 & \textbf{3.760} & \textbf{1.359} & 2.579 & 1.107 & \textbf{2.136} & \textbf{0.932} & \textbf{3.273} & \textbf{1.267} & 3.364 & 1.283 & 3.961 & 1.433 & \textbf{3.639} & \textbf{1.368} & 2.572 & 1.058 & \textbf{2.406} & \textbf{1.009} \\
\cmidrule(lr){2-22}
 & Avg & 3.903 & 1.441 & \textbf{3.571} & \textbf{1.365} & \textbf{2.217} & 0.967 & 2.225 & \textbf{0.930} & \textbf{3.177} & \textbf{1.254} & 3.191 & 1.254 & 3.963 & 1.422 & \textbf{3.765} & \textbf{1.377} & \textbf{2.456} & \textbf{0.983} & 2.572 & 1.038 \\
\bottomrule
\end{tabular}%
}
\end{table}

\FloatBarrier

\section{Loss-Baseline Results for All Backbones}
\label{app:loss_baselines_all_models}

The page limit restricts Table~\ref{tab:loss_baselines_itransformer} to iTransformer on five of the seven benchmarks. Tables~\ref{tab:loss_baselines_dlinear}--\ref{tab:loss_baselines_timexer} report all seven benchmarks separately for each backbone, completing \(280\) backbone, dataset, prediction-length, and metric cells.

No method is best in every cell. Counting a tie for each tied method, PV-Surgery gives the lowest value in \(91\) of the \(280\) cells and Selective Learning in \(77\). These tables compare the methods as alternatives and do not evaluate combined objectives. The specialized methods change the forecasting loss, whereas PV-Surgery retains pointwise MSE and modifies the optimizer update. The two mechanisms can therefore be combined in principle, but their joint effect is not evaluated here.


\begin{table}[p]
\centering
\caption{Loss-baseline comparison on DLinear for all seven benchmarks. The MSE row denotes standard MSE training. Best results are bolded and second-best results are underlined using unrounded scores.}
\label{tab:loss_baselines_dlinear}
\setlength{\tabcolsep}{2.6pt}
\renewcommand{\arraystretch}{1.02}
\scriptsize
\resizebox{\textwidth}{!}{%
\resizebox{\linewidth}{!}{%
\begin{tabular}{@{}c|c|rrrr|rrrr|rrrr|rrrr|rrrr|rrrr|rrrr@{}}
\toprule
\multicolumn{2}{c|}{Dataset} & \multicolumn{4}{c|}{ETTh1} & \multicolumn{4}{c|}{ETTh2} & \multicolumn{4}{c|}{ETTm1} & \multicolumn{4}{c|}{ETTm2} & \multicolumn{4}{c|}{Weather} & \multicolumn{4}{c|}{Exchange} & \multicolumn{4}{c}{ILI} \\
\midrule
\multicolumn{2}{c|}{Forecast length} & 96 & 192 & 336 & \multicolumn{1}{c|}{720} & 96 & 192 & 336 & \multicolumn{1}{c|}{720} & 96 & 192 & 336 & \multicolumn{1}{c|}{720} & 96 & 192 & 336 & \multicolumn{1}{c|}{720} & 96 & 192 & 336 & \multicolumn{1}{c|}{720} & 96 & 192 & 336 & \multicolumn{1}{c|}{720} & 24 & 36 & 48 & \multicolumn{1}{c}{60} \\
\midrule
\multirow{2}{*}{\textbf{MSE}} & MSE & 0.462 & 0.514 & 0.559 & 0.651 & 0.239 & 0.329 & 0.437 & 0.650 & 0.388 & 0.447 & 0.507 & 0.575 & 0.160 & 0.209 & 0.262 & 0.348 & 0.203 & 0.241 & 0.286 & 0.349 & \underline{0.105} & 0.224 & 0.425 & \textbf{0.652} & 3.828 & 3.801 & 3.808 & 4.175 \\
 & MAE & 0.444 & 0.476 & 0.506 & 0.573 & 0.329 & 0.393 & 0.462 & 0.582 & 0.395 & 0.426 & 0.459 & 0.502 & 0.268 & 0.308 & 0.350 & 0.408 & 0.266 & 0.302 & 0.338 & 0.385 & \underline{0.234} & 0.349 & 0.498 & \textbf{0.640} & 1.464 & 1.435 & 1.420 & 1.446 \\
\midrule
\multirow{2}{*}{\textbf{TILDE-Q}} & MSE & 0.467 & 0.516 & 0.558 & 0.647 & \underline{0.233} & \underline{0.293} & \underline{0.354} & \underline{0.524} & \textbf{0.383} & \textbf{0.445} & 0.510 & 0.576 & 0.158 & 0.203 & 0.246 & 0.317 & \underline{0.200} & \underline{0.238} & 0.282 & \underline{0.343} & 0.116 & 0.213 & 0.432 & 0.888 & 4.014 & \underline{3.685} & \underline{3.538} & 3.777 \\
 & MAE & 0.443 & 0.472 & \textbf{0.501} & 0.567 & \underline{0.317} & \underline{0.361} & \underline{0.408} & \underline{0.514} & \textbf{0.387} & \textbf{0.420} & 0.455 & \textbf{0.496} & 0.258 & 0.294 & 0.325 & \underline{0.375} & \underline{0.242} & \underline{0.281} & 0.321 & 0.371 & 0.250 & 0.348 & 0.509 & 0.760 & 1.529 & \underline{1.410} & \underline{1.352} & \underline{1.365} \\
\midrule
\multirow{2}{*}{\textbf{FreDF}} & MSE & \textbf{0.459} & \textbf{0.511} & \textbf{0.555} & \underline{0.644} & 0.235 & 0.308 & 0.391 & 0.591 & 0.386 & 0.446 & 0.506 & 0.574 & 0.157 & \underline{0.202} & 0.248 & 0.319 & 0.200 & \textbf{0.238} & \underline{0.281} & 0.344 & 0.107 & \underline{0.208} & 0.430 & 0.790 & \underline{3.810} & 3.734 & 3.797 & 4.146 \\
 & MAE & \textbf{0.439} & \underline{0.471} & \underline{0.501} & \underline{0.566} & 0.323 & 0.378 & 0.437 & 0.550 & 0.392 & 0.424 & 0.458 & 0.501 & 0.260 & 0.295 & 0.331 & 0.380 & 0.249 & 0.286 & \underline{0.320} & 0.370 & 0.239 & \underline{0.345} & 0.497 & 0.700 & 1.466 & 1.422 & 1.427 & 1.456 \\
\midrule
\multirow{2}{*}{\textbf{PSLoss}} & MSE & 0.462 & \underline{0.513} & 0.557 & 0.648 & 0.238 & 0.308 & 0.390 & 0.626 & 0.385 & 0.445 & \underline{0.505} & \underline{0.573} & \underline{0.156} & \textbf{0.197} & \underline{0.242} & \underline{0.317} & 0.202 & \underline{0.238} & \textbf{0.281} & \textbf{0.341} & 0.120 & 0.226 & \underline{0.388} & 0.700 & 3.961 & 3.775 & 3.577 & \underline{3.763} \\
 & MAE & 0.441 & 0.474 & 0.504 & 0.570 & 0.325 & 0.376 & 0.433 & 0.560 & 0.391 & \underline{0.421} & \underline{0.455} & \underline{0.497} & \underline{0.254} & \textbf{0.286} & \underline{0.322} & 0.377 & \textbf{0.242} & \textbf{0.277} & \textbf{0.311} & \textbf{0.359} & 0.256 & 0.358 & \underline{0.478} & 0.655 & 1.503 & 1.434 & 1.362 & 1.368 \\
\midrule
\multirow{2}{*}{\textbf{SL}} & MSE & \underline{0.460} & 0.514 & 0.558 & 0.650 & 0.244 & 0.323 & 0.395 & 0.623 & 0.392 & 0.452 & 0.511 & 0.578 & 0.160 & 0.216 & 0.275 & 0.378 & \textbf{0.200} & 0.240 & 0.288 & 0.348 & 0.105 & 0.219 & 0.406 & \underline{0.675} & 3.872 & 3.819 & 3.753 & 4.158 \\
 & MAE & 0.442 & 0.476 & 0.506 & 0.574 & 0.333 & 0.389 & 0.437 & 0.563 & 0.397 & 0.428 & 0.460 & 0.505 & 0.268 & 0.315 & 0.361 & 0.428 & 0.259 & 0.301 & 0.341 & 0.386 & 0.236 & 0.348 & 0.487 & \underline{0.649} & \underline{1.459} & 1.425 & 1.396 & 1.437 \\
\midrule
\multirow{2}{*}{\textbf{PV(Ours)}} & MSE & 0.461 & 0.514 & \underline{0.556} & \textbf{0.641} & \textbf{0.229} & \textbf{0.290} & \textbf{0.341} & \textbf{0.470} & \underline{0.383} & \underline{0.445} & \textbf{0.505} & \textbf{0.573} & \textbf{0.155} & \textbf{0.197} & \textbf{0.239} & \textbf{0.306} & 0.203 & 0.241 & 0.283 & 0.345 & \textbf{0.100} & \textbf{0.191} & \textbf{0.321} & 0.847 & \textbf{3.590} & \textbf{3.490} & \textbf{3.446} & \textbf{3.760} \\
 & MAE & \underline{0.440} & \textbf{0.471} & \textbf{0.501} & \textbf{0.561} & \textbf{0.313} & \textbf{0.357} & \textbf{0.396} & \textbf{0.484} & \underline{0.389} & \underline{0.421} & \textbf{0.454} & 0.499 & \textbf{0.253} & \underline{0.287} & \textbf{0.319} & \textbf{0.365} & 0.245 & 0.292 & 0.325 & \underline{0.368} & \textbf{0.226} & \textbf{0.323} & \textbf{0.425} & 0.724 & \textbf{1.414} & \textbf{1.358} & \textbf{1.330} & \textbf{1.359} \\
\bottomrule
\end{tabular}%
}
}
\end{table}

\begin{table}[p]
\centering
\caption{Loss-baseline comparison on iTransformer for all seven benchmarks. The MSE row denotes standard MSE training. Best results are bolded and second-best results are underlined using unrounded scores.}
\label{tab:loss_baselines_itransformer_full}
\setlength{\tabcolsep}{2.6pt}
\renewcommand{\arraystretch}{1.02}
\scriptsize
\resizebox{\textwidth}{!}{%
\resizebox{\linewidth}{!}{%
\begin{tabular}{@{}c|c|rrrr|rrrr|rrrr|rrrr|rrrr|rrrr|rrrr@{}}
\toprule
\multicolumn{2}{c|}{Dataset} & \multicolumn{4}{c|}{ETTh1} & \multicolumn{4}{c|}{ETTh2} & \multicolumn{4}{c|}{ETTm1} & \multicolumn{4}{c|}{ETTm2} & \multicolumn{4}{c|}{Weather} & \multicolumn{4}{c|}{Exchange} & \multicolumn{4}{c}{ILI} \\
\midrule
\multicolumn{2}{c|}{Forecast length} & 96 & 192 & 336 & \multicolumn{1}{c|}{720} & 96 & 192 & 336 & \multicolumn{1}{c|}{720} & 96 & 192 & 336 & \multicolumn{1}{c|}{720} & 96 & 192 & 336 & \multicolumn{1}{c|}{720} & 96 & 192 & 336 & \multicolumn{1}{c|}{720} & 96 & 192 & 336 & \multicolumn{1}{c|}{720} & 24 & 36 & 48 & \multicolumn{1}{c}{60} \\
\midrule
\multirow{2}{*}{\textbf{MSE}} & MSE & 0.454 & 0.517 & 0.561 & 0.671 & 0.239 & 0.303 & 0.356 & 0.457 & 0.407 & 0.480 & \underline{0.520} & 0.618 & 0.157 & 0.205 & 0.252 & 0.322 & 0.183 & 0.231 & 0.285 & 0.358 & 0.108 & 0.222 & 0.397 & \underline{1.099} & 2.018 & 2.173 & \textbf{2.099} & 2.579 \\
 & MAE & 0.447 & 0.485 & 0.513 & 0.583 & 0.323 & 0.366 & 0.401 & 0.460 & 0.415 & 0.457 & 0.477 & 0.531 & 0.258 & 0.295 & 0.328 & 0.373 & 0.222 & 0.263 & 0.300 & 0.349 & 0.236 & 0.344 & 0.463 & \underline{0.800} & \underline{0.878} & 0.945 & \underline{0.940} & 1.107 \\
\midrule
\multirow{2}{*}{\textbf{TILDE-Q}} & MSE & 0.469 & 0.528 & 0.564 & 0.669 & 0.233 & \underline{0.287} & 0.354 & 0.473 & \textbf{0.402} & 0.465 & 0.537 & 0.622 & 0.156 & 0.200 & 0.247 & 0.311 & 0.180 & 0.231 & 0.283 & \textbf{0.350} & 0.111 & 0.219 & 0.428 & 1.109 & \underline{1.969} & \underline{2.035} & \underline{2.143} & 2.548 \\
 & MAE & 0.455 & 0.489 & 0.510 & 0.577 & 0.315 & \underline{0.353} & 0.400 & 0.466 & \underline{0.404} & 0.440 & 0.483 & 0.534 & 0.251 & 0.285 & 0.320 & 0.362 & 0.217 & 0.261 & 0.297 & \textbf{0.342} & 0.239 & 0.341 & 0.481 & 0.806 & 0.928 & 0.931 & \textbf{0.934} & 1.095 \\
\midrule
\multirow{2}{*}{\textbf{FreDF}} & MSE & \textbf{0.451} & \underline{0.512} & 0.560 & \underline{0.662} & 0.235 & 0.295 & 0.350 & 0.450 & 0.410 & 0.475 & 0.553 & 0.603 & 0.154 & 0.197 & 0.245 & \textbf{0.309} & \underline{0.178} & \underline{0.227} & 0.283 & 0.354 & 0.112 & 0.223 & 0.462 & 1.132 & 2.047 & 2.079 & 2.272 & 2.507 \\
 & MAE & \underline{0.445} & \underline{0.478} & 0.508 & 0.576 & 0.317 & 0.360 & 0.397 & 0.455 & 0.408 & 0.452 & 0.492 & 0.524 & 0.250 & 0.284 & 0.319 & \underline{0.362} & \underline{0.216} & \underline{0.257} & \underline{0.297} & 0.346 & 0.241 & 0.344 & 0.502 & 0.819 & \textbf{0.872} & \textbf{0.905} & 0.984 & 1.081 \\
\midrule
\multirow{2}{*}{\textbf{PSLoss}} & MSE & 0.458 & 0.513 & \underline{0.558} & 0.665 & 0.234 & 0.291 & 0.360 & 0.473 & 0.429 & 0.499 & 0.538 & 0.626 & \underline{0.153} & \underline{0.196} & \textbf{0.239} & 0.310 & \textbf{0.177} & \textbf{0.225} & \textbf{0.278} & \underline{0.352} & 0.113 & 0.213 & 0.413 & 1.146 & \textbf{1.880} & \textbf{2.025} & 2.300 & 2.562 \\
 & MAE & 0.448 & 0.480 & 0.509 & 0.577 & 0.316 & 0.355 & 0.406 & 0.469 & 0.421 & 0.459 & 0.482 & 0.535 & \underline{0.247} & \textbf{0.281} & \textbf{0.314} & \textbf{0.361} & \textbf{0.214} & \textbf{0.254} & \textbf{0.293} & \underline{0.342} & 0.241 & 0.338 & 0.476 & 0.825 & 0.899 & \underline{0.927} & 1.003 & 1.119 \\
\midrule
\multirow{2}{*}{\textbf{SL}} & MSE & 0.465 & 0.527 & 0.560 & 0.665 & \textbf{0.229} & \textbf{0.284} & \underline{0.348} & \textbf{0.438} & 0.404 & \textbf{0.458} & 0.521 & \underline{0.590} & 0.153 & 0.199 & 0.243 & 0.312 & 0.182 & 0.236 & \underline{0.280} & 0.356 & \underline{0.108} & \textbf{0.211} & \underline{0.393} & 1.228 & 2.324 & 2.312 & 2.324 & \underline{2.348} \\
 & MAE & 0.449 & 0.485 & \textbf{0.501} & \textbf{0.568} & \textbf{0.312} & \textbf{0.349} & \underline{0.393} & \textbf{0.449} & \textbf{0.400} & \textbf{0.432} & \textbf{0.465} & \textbf{0.505} & 0.250 & 0.285 & \underline{0.317} & 0.362 & 0.219 & 0.264 & 0.297 & 0.346 & \underline{0.232} & \textbf{0.332} & \underline{0.460} & 0.856 & 0.972 & 0.985 & 0.982 & \underline{1.018} \\
\midrule
\multirow{2}{*}{\textbf{PV(Ours)}} & MSE & \underline{0.451} & \textbf{0.508} & \textbf{0.550} & \textbf{0.661} & \underline{0.229} & 0.287 & \textbf{0.334} & \underline{0.446} & \underline{0.402} & \underline{0.461} & \textbf{0.517} & \textbf{0.575} & \textbf{0.150} & \textbf{0.195} & \underline{0.240} & \underline{0.310} & 0.182 & 0.231 & 0.281 & 0.353 & \textbf{0.105} & \underline{0.211} & \textbf{0.390} & \textbf{1.075} & 2.190 & 2.293 & 2.283 & \textbf{2.136} \\
 & MAE & \textbf{0.442} & \textbf{0.476} & \underline{0.504} & \underline{0.575} & \underline{0.314} & 0.355 & \textbf{0.388} & \underline{0.452} & 0.404 & \underline{0.439} & \underline{0.473} & \underline{0.510} & \textbf{0.247} & \underline{0.283} & \underline{0.317} & 0.365 & 0.222 & 0.265 & 0.299 & 0.345 & \textbf{0.231} & \underline{0.335} & \textbf{0.460} & \textbf{0.792} & 0.905 & 0.943 & 0.941 & \textbf{0.932} \\
\bottomrule
\end{tabular}%
}
}
\end{table}

\begin{table}[p]
\centering
\caption{Loss-baseline comparison on MICN for all seven benchmarks. The MSE row denotes standard MSE training. Best results are bolded and second-best results are underlined using unrounded scores.}
\label{tab:loss_baselines_micn}
\setlength{\tabcolsep}{2.6pt}
\renewcommand{\arraystretch}{1.02}
\scriptsize
\resizebox{\textwidth}{!}{%
\resizebox{\linewidth}{!}{%
\begin{tabular}{@{}c|c|rrrr|rrrr|rrrr|rrrr|rrrr|rrrr|rrrr@{}}
\toprule
\multicolumn{2}{c|}{Dataset} & \multicolumn{4}{c|}{ETTh1} & \multicolumn{4}{c|}{ETTh2} & \multicolumn{4}{c|}{ETTm1} & \multicolumn{4}{c|}{ETTm2} & \multicolumn{4}{c|}{Weather} & \multicolumn{4}{c|}{Exchange} & \multicolumn{4}{c}{ILI} \\
\midrule
\multicolumn{2}{c|}{Forecast length} & 96 & 192 & 336 & \multicolumn{1}{c|}{720} & 96 & 192 & 336 & \multicolumn{1}{c|}{720} & 96 & 192 & 336 & \multicolumn{1}{c|}{720} & 96 & 192 & 336 & \multicolumn{1}{c|}{720} & 96 & 192 & 336 & \multicolumn{1}{c|}{720} & 96 & 192 & 336 & \multicolumn{1}{c|}{720} & 24 & 36 & 48 & \multicolumn{1}{c}{60} \\
\midrule
\multirow{2}{*}{\textbf{MSE}} & MSE & 0.552 & 0.594 & 0.688 & 0.753 & 0.238 & 0.315 & 0.452 & 0.672 & 0.445 & 0.495 & 0.567 & 0.627 & 0.166 & 0.202 & 0.266 & 0.389 & 0.398 & 0.403 & 0.380 & 0.496 & 0.123 & 0.219 & 0.498 & 2.931 & \textbf{3.169} & 2.996 & 3.269 & \textbf{3.273} \\
 & MAE & 0.530 & 0.553 & 0.623 & 0.671 & 0.333 & 0.388 & 0.472 & 0.583 & 0.472 & 0.511 & 0.570 & 0.596 & 0.276 & 0.303 & 0.355 & 0.434 & 0.442 & 0.447 & 0.422 & 0.501 & 0.262 & 0.352 & 0.525 & 1.400 & \textbf{1.237} & 1.228 & 1.284 & \textbf{1.267} \\
\midrule
\multirow{2}{*}{\textbf{TILDE-Q}} & MSE & \textbf{0.480} & \textbf{0.529} & \textbf{0.556} & 0.916 & 0.237 & \underline{0.300} & 0.403 & 0.656 & \underline{0.408} & \underline{0.449} & \underline{0.505} & \textbf{0.548} & 0.153 & 0.200 & 0.241 & 0.308 & 0.299 & 0.400 & 0.506 & 0.506 & 0.117 & 0.215 & 0.419 & 3.289 & 4.109 & 3.027 & 3.145 & 3.475 \\
 & MAE & \underline{0.485} & 0.518 & \textbf{0.534} & 0.747 & \underline{0.325} & 0.371 & 0.445 & 0.579 & \underline{0.434} & \underline{0.464} & \underline{0.501} & \textbf{0.533} & 0.257 & 0.295 & 0.324 & 0.373 & 0.364 & 0.431 & 0.501 & 0.501 & 0.252 & 0.353 & 0.498 & 1.426 & 1.477 & 1.245 & 1.265 & 1.303 \\
\midrule
\multirow{2}{*}{\textbf{FreDF}} & MSE & 0.501 & 0.548 & 0.579 & \textbf{0.699} & 0.241 & 0.326 & 0.396 & 0.735 & 0.423 & 0.449 & 0.516 & \underline{0.570} & \underline{0.148} & 0.191 & 0.236 & \textbf{0.301} & 0.325 & \underline{0.322} & \textbf{0.340} & \underline{0.451} & 0.119 & \textbf{0.210} & \underline{0.416} & 3.067 & 3.585 & 2.999 & \textbf{3.054} & \underline{3.286} \\
 & MAE & 0.504 & 0.533 & 0.552 & \textbf{0.630} & 0.332 & 0.396 & 0.446 & 0.618 & 0.454 & 0.471 & 0.510 & \underline{0.554} & \underline{0.252} & 0.286 & 0.322 & \textbf{0.368} & 0.386 & \underline{0.384} & \textbf{0.388} & \underline{0.469} & 0.254 & \underline{0.350} & 0.495 & 1.426 & 1.325 & 1.236 & \textbf{1.228} & 1.274 \\
\midrule
\multirow{2}{*}{\textbf{PSLoss}} & MSE & 0.492 & \textbf{0.529} & \underline{0.558} & \underline{0.753} & \underline{0.236} & 0.309 & 0.402 & 0.630 & 0.428 & 0.475 & 0.516 & 0.924 & 0.149 & \textbf{0.187} & \underline{0.233} & \underline{0.302} & \underline{0.282} & 0.337 & 0.423 & 0.480 & 0.129 & 0.243 & 0.446 & 1.649 & 3.231 & 2.991 & \underline{3.141} & 3.347 \\
 & MAE & 0.493 & \underline{0.517} & \underline{0.534} & 0.666 & 0.328 & 0.381 & 0.450 & 0.566 & 0.450 & 0.483 & 0.505 & 0.761 & 0.256 & \textbf{0.283} & \underline{0.320} & \underline{0.371} & \underline{0.357} & 0.394 & 0.453 & 0.486 & 0.268 & 0.373 & 0.505 & 0.939 & 1.255 & 1.230 & 1.250 & \underline{1.271} \\
\midrule
\multirow{2}{*}{\textbf{SL}} & MSE & \underline{0.489} & \underline{0.532} & 0.624 & 0.769 & 0.236 & 0.301 & \underline{0.385} & \underline{0.615} & \textbf{0.397} & \textbf{0.444} & \textbf{0.503} & 0.611 & 0.153 & 0.209 & 0.258 & 0.384 & \textbf{0.247} & \textbf{0.321} & \underline{0.376} & \textbf{0.435} & \textbf{0.108} & 0.220 & 0.448 & \underline{1.353} & 3.317 & \textbf{2.975} & 3.185 & 3.586 \\
 & MAE & \textbf{0.466} & \textbf{0.494} & 0.586 & \underline{0.659} & 0.326 & \underline{0.371} & \underline{0.431} & \underline{0.565} & \textbf{0.408} & \textbf{0.442} & \textbf{0.468} & 0.555 & 0.257 & 0.308 & 0.345 & 0.434 & \textbf{0.322} & \textbf{0.378} & \underline{0.416} & \textbf{0.464} & \textbf{0.240} & 0.351 & \underline{0.494} & \underline{0.854} & 1.251 & \textbf{1.208} & \underline{1.243} & 1.310 \\
\midrule
\multirow{2}{*}{\textbf{PV(Ours)}} & MSE & 0.492 & 0.561 & 0.700 & 0.798 & \textbf{0.230} & \textbf{0.291} & \textbf{0.352} & \textbf{0.513} & 0.430 & 0.470 & 0.508 & 0.576 & \textbf{0.146} & \underline{0.190} & \textbf{0.231} & 0.331 & 0.318 & 0.435 & 0.436 & 0.479 & \underline{0.109} & \underline{0.212} & \textbf{0.378} & \textbf{1.106} & \underline{3.190} & \underline{2.985} & 3.228 & 3.364 \\
 & MAE & 0.497 & 0.543 & 0.630 & 0.700 & \textbf{0.320} & \textbf{0.369} & \textbf{0.412} & \textbf{0.520} & 0.460 & 0.487 & 0.513 & 0.562 & \textbf{0.248} & \underline{0.285} & \textbf{0.316} & 0.389 & 0.383 & 0.460 & 0.468 & 0.493 & \underline{0.242} & \textbf{0.348} & \textbf{0.470} & \textbf{0.793} & \underline{1.243} & \underline{1.220} & 1.272 & 1.283 \\
\bottomrule
\end{tabular}%
}
}
\end{table}

\begin{table}[p]
\centering
\caption{Loss-baseline comparison on SCINet for all seven benchmarks. The MSE row denotes standard MSE training. Best results are bolded and second-best results are underlined using unrounded scores.}
\label{tab:loss_baselines_scinet}
\setlength{\tabcolsep}{2.6pt}
\renewcommand{\arraystretch}{1.02}
\scriptsize
\resizebox{\textwidth}{!}{%
\resizebox{\linewidth}{!}{%
\begin{tabular}{@{}c|c|rrrr|rrrr|rrrr|rrrr|rrrr|rrrr|rrrr@{}}
\toprule
\multicolumn{2}{c|}{Dataset} & \multicolumn{4}{c|}{ETTh1} & \multicolumn{4}{c|}{ETTh2} & \multicolumn{4}{c|}{ETTm1} & \multicolumn{4}{c|}{ETTm2} & \multicolumn{4}{c|}{Weather} & \multicolumn{4}{c|}{Exchange} & \multicolumn{4}{c}{ILI} \\
\midrule
\multicolumn{2}{c|}{Forecast length} & 96 & 192 & 336 & \multicolumn{1}{c|}{720} & 96 & 192 & 336 & \multicolumn{1}{c|}{720} & 96 & 192 & 336 & \multicolumn{1}{c|}{720} & 96 & 192 & 336 & \multicolumn{1}{c|}{720} & 96 & 192 & 336 & \multicolumn{1}{c|}{720} & 96 & 192 & 336 & \multicolumn{1}{c|}{720} & 24 & 36 & 48 & \multicolumn{1}{c}{60} \\
\midrule
\multirow{2}{*}{\textbf{MSE}} & MSE & 0.490 & 0.545 & 0.590 & 0.700 & 0.254 & 0.311 & 0.356 & 0.470 & 0.448 & 0.488 & 0.548 & 0.614 & 0.155 & 0.202 & 0.248 & 0.321 & 0.166 & 0.218 & 0.280 & 0.361 & 0.120 & 0.223 & 0.405 & 1.101 & 3.946 & 3.970 & 3.977 & 3.961 \\
 & MAE & 0.464 & 0.495 & 0.521 & 0.588 & 0.334 & 0.370 & 0.402 & 0.472 & 0.428 & 0.453 & 0.483 & 0.521 & 0.253 & 0.290 & 0.322 & 0.368 & 0.212 & 0.259 & 0.303 & 0.356 & 0.249 & 0.347 & 0.470 & 0.804 & 1.405 & 1.417 & 1.434 & 1.433 \\
\midrule
\multirow{2}{*}{\textbf{TILDE-Q}} & MSE & 0.485 & 0.540 & 0.585 & 0.696 & 0.238 & 0.301 & 0.353 & 0.454 & 0.441 & \underline{0.475} & 0.534 & 0.603 & \underline{0.148} & 0.194 & 0.240 & 0.312 & 0.162 & 0.217 & 0.277 & 0.352 & 0.116 & 0.218 & 0.399 & \underline{1.092} & 4.101 & 4.126 & 3.969 & 3.921 \\
 & MAE & 0.457 & 0.488 & 0.514 & 0.583 & 0.320 & 0.362 & 0.397 & 0.457 & 0.414 & \underline{0.437} & \underline{0.468} & 0.511 & \underline{0.243} & 0.279 & 0.313 & 0.359 & 0.208 & 0.256 & 0.298 & \underline{0.346} & 0.246 & 0.342 & 0.464 & 0.804 & 1.463 & 1.448 & 1.432 & 1.428 \\
\midrule
\multirow{2}{*}{\textbf{FreDF}} & MSE & \textbf{0.470} & \textbf{0.526} & \textbf{0.571} & \textbf{0.680} & \textbf{0.234} & \textbf{0.293} & \textbf{0.345} & \textbf{0.447} & 0.444 & 0.481 & \underline{0.531} & \underline{0.589} & \textbf{0.146} & \textbf{0.192} & \textbf{0.236} & \textbf{0.307} & 0.161 & 0.213 & \underline{0.271} & \underline{0.352} & \underline{0.112} & \underline{0.216} & \underline{0.393} & 1.093 & \textbf{3.713} & \textbf{3.699} & 3.842 & 3.939 \\
 & MAE & \textbf{0.447} & \textbf{0.479} & \textbf{0.505} & \textbf{0.575} & \textbf{0.316} & \textbf{0.356} & \underline{0.392} & \underline{0.454} & 0.419 & 0.445 & 0.471 & \underline{0.504} & \textbf{0.243} & \underline{0.278} & \textbf{0.309} & \textbf{0.356} & 0.206 & 0.253 & 0.295 & 0.347 & \underline{0.242} & \underline{0.340} & \underline{0.460} & \underline{0.802} & \textbf{1.359} & \textbf{1.365} & 1.419 & 1.434 \\
\midrule
\multirow{2}{*}{\textbf{PSLoss}} & MSE & 0.477 & 0.533 & 0.578 & 0.684 & 0.240 & 0.299 & 0.346 & 0.449 & 0.450 & 0.480 & 0.542 & 0.607 & 0.148 & 0.194 & 0.237 & 0.309 & \underline{0.161} & \textbf{0.211} & 0.271 & \textbf{0.351} & 0.122 & 0.225 & 0.401 & \textbf{1.086} & 3.951 & 3.799 & \textbf{3.646} & \textbf{3.553} \\
 & MAE & 0.453 & 0.485 & 0.512 & 0.580 & 0.320 & 0.359 & 0.393 & 0.457 & 0.422 & 0.440 & 0.472 & 0.509 & 0.246 & 0.281 & 0.311 & 0.358 & \underline{0.205} & \underline{0.250} & \underline{0.293} & \textbf{0.345} & 0.253 & 0.347 & 0.466 & \textbf{0.797} & \underline{1.399} & \underline{1.368} & \textbf{1.359} & \textbf{1.355} \\
\midrule
\multirow{2}{*}{\textbf{SL}} & MSE & 0.476 & 0.533 & 0.577 & \underline{0.683} & \underline{0.235} & \underline{0.297} & 0.348 & 0.453 & \textbf{0.389} & \textbf{0.449} & \textbf{0.513} & \textbf{0.585} & 0.152 & 0.196 & 0.241 & 0.314 & \textbf{0.160} & \underline{0.212} & \textbf{0.270} & 0.353 & \textbf{0.111} & \textbf{0.212} & \textbf{0.385} & 1.181 & 3.980 & 3.959 & 3.924 & 3.888 \\
 & MAE & \underline{0.449} & \underline{0.482} & \underline{0.507} & \underline{0.576} & 0.318 & 0.359 & 0.394 & 0.457 & \textbf{0.391} & \textbf{0.422} & \textbf{0.455} & \textbf{0.495} & 0.250 & 0.282 & 0.316 & 0.361 & \textbf{0.203} & \textbf{0.250} & \textbf{0.293} & 0.348 & \textbf{0.238} & \textbf{0.336} & \textbf{0.457} & 0.836 & 1.406 & 1.413 & 1.423 & 1.420 \\
\midrule
\multirow{2}{*}{\textbf{PV(Ours)}} & MSE & \underline{0.473} & \underline{0.530} & \underline{0.574} & 0.687 & 0.237 & 0.297 & \underline{0.346} & \underline{0.447} & \underline{0.421} & 0.490 & 0.551 & 0.628 & 0.150 & \underline{0.193} & \underline{0.236} & \underline{0.308} & 0.170 & 0.218 & 0.278 & 0.357 & 0.116 & 0.222 & 0.409 & 1.103 & \underline{3.945} & \underline{3.778} & \underline{3.698} & \underline{3.639} \\
 & MAE & 0.449 & 0.483 & 0.510 & 0.581 & \underline{0.317} & \underline{0.358} & \textbf{0.391} & \textbf{0.454} & \underline{0.414} & 0.446 & 0.475 & 0.516 & 0.246 & \textbf{0.278} & \underline{0.309} & \underline{0.357} & 0.215 & 0.257 & 0.300 & 0.352 & 0.244 & 0.346 & 0.473 & 0.806 & 1.400 & 1.369 & \underline{1.371} & \underline{1.368} \\
\bottomrule
\end{tabular}%
}
}
\end{table}

\begin{table}[p]
\centering
\caption{Loss-baseline comparison on TimeXer for all seven benchmarks. The MSE row denotes standard MSE training. Best results are bolded and second-best results are underlined using unrounded scores.}
\label{tab:loss_baselines_timexer}
\setlength{\tabcolsep}{2.6pt}
\renewcommand{\arraystretch}{1.02}
\scriptsize
\resizebox{\textwidth}{!}{%
\resizebox{\linewidth}{!}{%
\begin{tabular}{@{}c|c|rrrr|rrrr|rrrr|rrrr|rrrr|rrrr|rrrr@{}}
\toprule
\multicolumn{2}{c|}{Dataset} & \multicolumn{4}{c|}{ETTh1} & \multicolumn{4}{c|}{ETTh2} & \multicolumn{4}{c|}{ETTm1} & \multicolumn{4}{c|}{ETTm2} & \multicolumn{4}{c|}{Weather} & \multicolumn{4}{c|}{Exchange} & \multicolumn{4}{c}{ILI} \\
\midrule
\multicolumn{2}{c|}{Forecast length} & 96 & 192 & 336 & \multicolumn{1}{c|}{720} & 96 & 192 & 336 & \multicolumn{1}{c|}{720} & 96 & 192 & 336 & \multicolumn{1}{c|}{720} & 96 & 192 & 336 & \multicolumn{1}{c|}{720} & 96 & 192 & 336 & \multicolumn{1}{c|}{720} & 96 & 192 & 336 & \multicolumn{1}{c|}{720} & 24 & 36 & 48 & \multicolumn{1}{c}{60} \\
\midrule
\multirow{2}{*}{\textbf{MSE}} & MSE & \textbf{0.460} & \textbf{0.511} & \underline{0.564} & 0.693 & 0.233 & \underline{0.294} & 0.341 & 0.449 & 0.432 & 0.477 & 0.521 & 0.580 & 0.151 & 0.200 & 0.239 & 0.312 & 0.164 & \textbf{0.208} & \textbf{0.261} & \textbf{0.340} & \underline{0.113} & \underline{0.216} & \underline{0.396} & \underline{1.105} & 2.798 & \textbf{2.204} & \textbf{2.250} & 2.572 \\
 & MAE & 0.454 & 0.485 & \underline{0.511} & 0.590 & 0.319 & 0.360 & 0.394 & 0.457 & 0.421 & 0.453 & 0.479 & 0.515 & 0.249 & 0.287 & 0.316 & 0.367 & \underline{0.208} & \underline{0.251} & \underline{0.290} & \underline{0.342} & 0.239 & \underline{0.337} & \underline{0.459} & \underline{0.798} & \underline{0.990} & \textbf{0.931} & \textbf{0.954} & 1.058 \\
\midrule
\multirow{2}{*}{\textbf{TILDE-Q}} & MSE & 0.474 & 0.525 & 0.574 & \underline{0.693} & 0.238 & 0.295 & \underline{0.338} & \textbf{0.436} & 0.417 & 0.460 & \underline{0.517} & \underline{0.566} & \textbf{0.148} & \underline{0.194} & \underline{0.237} & \textbf{0.305} & \underline{0.163} & 0.216 & 0.275 & 0.352 & 0.122 & 0.227 & 0.456 & 1.171 & \underline{2.615} & 2.503 & 2.520 & 2.547 \\
 & MAE & 0.455 & 0.488 & 0.514 & 0.592 & 0.323 & 0.359 & \underline{0.389} & \textbf{0.448} & \underline{0.407} & 0.437 & 0.471 & 0.510 & \textbf{0.244} & \textbf{0.280} & \textbf{0.312} & \textbf{0.357} & 0.209 & 0.258 & 0.298 & 0.348 & 0.250 & 0.348 & 0.502 & 0.836 & 1.035 & 1.022 & 1.028 & 1.040 \\
\midrule
\multirow{2}{*}{\textbf{FreDF}} & MSE & 0.475 & 0.556 & 0.658 & 0.802 & 0.243 & 0.300 & 0.348 & 0.448 & 0.452 & 0.510 & 0.575 & 0.661 & 0.155 & 0.202 & 0.244 & 0.316 & 0.165 & 0.224 & 0.281 & 0.361 & 0.133 & 0.256 & 0.478 & 1.226 & 2.659 & 2.526 & 2.640 & 2.658 \\
 & MAE & 0.461 & 0.510 & 0.562 & 0.642 & 0.330 & 0.366 & 0.398 & 0.456 & 0.440 & 0.471 & 0.501 & 0.552 & 0.257 & 0.292 & 0.320 & 0.367 & 0.216 & 0.270 & 0.307 & 0.360 & 0.262 & 0.373 & 0.515 & 0.857 & 1.010 & 1.031 & 1.085 & 1.090 \\
\midrule
\multirow{2}{*}{\textbf{PSLoss}} & MSE & 0.468 & 0.522 & 0.579 & 0.699 & 0.237 & 0.295 & 0.342 & \underline{0.438} & 0.417 & 0.465 & 0.517 & 0.572 & 0.152 & 0.196 & 0.238 & 0.308 & 0.165 & 0.219 & 0.282 & 0.357 & 0.137 & 0.254 & 0.452 & 1.192 & 2.710 & \underline{2.316} & 2.671 & 2.541 \\
 & MAE & 0.459 & 0.494 & 0.520 & 0.602 & 0.321 & 0.360 & 0.391 & \underline{0.448} & 0.416 & 0.445 & 0.477 & 0.512 & 0.252 & 0.285 & 0.315 & \underline{0.360} & 0.216 & 0.262 & 0.304 & 0.352 & 0.268 & 0.372 & 0.499 & 0.844 & 1.044 & \underline{0.973} & 1.090 & 1.081 \\
\midrule
\multirow{2}{*}{\textbf{SL}} & MSE & 0.475 & 0.529 & 0.572 & 0.703 & \textbf{0.226} & \textbf{0.283} & \textbf{0.329} & 0.438 & \underline{0.407} & \textbf{0.455} & 0.522 & 0.600 & 0.153 & 0.198 & 0.241 & 0.322 & \textbf{0.162} & \underline{0.211} & \underline{0.263} & \underline{0.342} & \textbf{0.108} & 0.222 & 0.412 & 1.208 & \textbf{2.539} & 2.584 & \underline{2.450} & \textbf{2.392} \\
 & MAE & \textbf{0.448} & \textbf{0.480} & \textbf{0.507} & \underline{0.586} & \textbf{0.311} & \textbf{0.350} & \textbf{0.380} & 0.454 & 0.407 & \textbf{0.425} & \textbf{0.460} & \textbf{0.502} & 0.252 & 0.284 & 0.315 & 0.368 & \textbf{0.204} & \textbf{0.249} & \textbf{0.288} & \textbf{0.341} & \textbf{0.229} & 0.340 & 0.472 & 0.845 & \textbf{0.986} & 1.038 & \underline{1.018} & \textbf{1.001} \\
\midrule
\multirow{2}{*}{\textbf{PV(Ours)}} & MSE & \underline{0.461} & \underline{0.518} & \textbf{0.564} & \textbf{0.691} & \underline{0.231} & 0.295 & 0.342 & 0.446 & \textbf{0.396} & \underline{0.455} & \textbf{0.505} & \textbf{0.564} & \underline{0.149} & \textbf{0.193} & \textbf{0.235} & \underline{0.307} & 0.180 & 0.233 & 0.283 & 0.356 & 0.113 & \textbf{0.213} & \textbf{0.382} & \textbf{1.010} & 2.723 & 2.619 & 2.542 & \underline{2.406} \\
 & MAE & \underline{0.450} & \underline{0.485} & 0.515 & \textbf{0.581} & \underline{0.316} & \underline{0.358} & 0.390 & 0.451 & \textbf{0.403} & \underline{0.429} & \underline{0.462} & \underline{0.503} & \underline{0.247} & \underline{0.282} & \underline{0.312} & \underline{0.360} & 0.223 & 0.269 & 0.302 & 0.350 & \underline{0.239} & \textbf{0.335} & \textbf{0.452} & \textbf{0.763} & 1.047 & 1.056 & 1.041 & \underline{1.009} \\
\bottomrule
\end{tabular}%
}
}
\end{table}

\section{Dataset-Wise Component Ablations}
\label{app:ablation_all_datasets}

Table~\ref{tab:core_ablation} reports the component ablation for MICN on Exchange. Tables~\ref{tab:core_ablation_etth1}--\ref{tab:core_ablation_ili} give the corresponding result for each dataset. Every entry averages the five backbones and four prediction lengths, yielding \(20\) matched settings per dataset. Relative changes are computed against the full configuration using the unrounded averages.

For every removal, the mean of the seven dataset-level relative MSE changes is positive, so each module contributes on average. Removing a component improves MSE on some datasets, but the largest such gain is \(0.9\%\). Every ablated variant is also at least \(2.1\%\) worse than the full configuration on another dataset. Overall, the full configuration has lower MSE in \(37\) of the \(49\) dataset-variant comparisons. It is not uniformly best, but retaining all modules avoids the larger and more frequent degradations and provides the most stable performance across datasets.


\begin{table}[p]
\centering
\caption{Core component ablation on ETTh1, averaged over 20 matched backbone--prediction-length settings. \(\Delta\) is the relative change against the full configuration. Bold indicates the best score and underlining the second best using unrounded averages.}
\label{tab:core_ablation_etth1}
\vspace{1mm}
\footnotesize
\renewcommand{\arraystretch}{0.5}
\begin{tabular*}{\linewidth}{@{\extracolsep{\fill}}cl>{\fontsize{8}{9.6}\selectfont}r>{\scriptsize}r>{\fontsize{8}{9.6}\selectfont}r>{\scriptsize}r>{\fontsize{8}{9.6}\selectfont}c@{}}
\toprule
\multirow{2}{*}{\textbf{Category}} & \multirow{2}{*}{\textbf{Module}} & \multicolumn{2}{c}{\textbf{MSE}} & \multicolumn{2}{c}{\textbf{MAE}} & \multirow{2}{*}{\textbf{Time (s)}} \\
\cmidrule(lr){3-4}\cmidrule(lr){5-6}
 & & \textbf{value} & \(\boldsymbol{\Delta\%}\) & \textbf{value} & \(\boldsymbol{\Delta\%}\) & \\
\midrule
Gradient proxy & w/o output-cache proxy\textsuperscript{a} & 0.574 & \(+0.8\) & 0.525 & \(+0.9\) & 192.97 \\
\cmidrule(lr){1-7}
Layer selection & w/o layer selection & 0.569 & \(-0.2\) & 0.520 & \(-0.0\) & 288.11 \\
\cmidrule(lr){1-7}
\multirow{3}{*}{Variable pooling} & w/o conflict pooling & 0.576 & \(+1.1\) & 0.524 & \(+0.8\) & 197.07 \\
 & w/o anchor pooling & 0.568 & \(-0.2\) & 0.519 & \(-0.0\) & 213.68 \\
 & w/o variable pooling & \textbf{0.568} & \(-0.3\) & \underline{0.519} & \(-0.1\) & 198.92 \\
\cmidrule(lr){1-7}
\multirow{2}{*}{Direction surgery} & w/o magnitude decoupling & \underline{0.568} & \(-0.3\) & \textbf{0.519} & \(-0.2\) & 225.96 \\
 & w/o safe candidate selection & 0.578 & \(+1.5\) & 0.525 & \(+0.9\) & 194.06 \\
\midrule
\multicolumn{2}{c}{\textbf{PV-Surgery (full)}} & 0.570 & -- & 0.520 & -- & 213.65 \\
\bottomrule
\end{tabular*}
\par\vspace{0.1em}
{\raggedright\scriptsize\textsuperscript{a} Removing the output-cache proxy also disables cache-dependent layer selection, pooling, and candidate comparison.\par}
\end{table}

\begin{table}[p]
\centering
\caption{Core component ablation on ETTh2, averaged over 20 matched backbone--prediction-length settings. \(\Delta\) is the relative change against the full configuration. Bold indicates the best score and underlining the second best using unrounded averages.}
\label{tab:core_ablation_etth2}
\vspace{1mm}
\footnotesize
\renewcommand{\arraystretch}{0.5}
\begin{tabular*}{\linewidth}{@{\extracolsep{\fill}}cl>{\fontsize{8}{9.6}\selectfont}r>{\scriptsize}r>{\fontsize{8}{9.6}\selectfont}r>{\scriptsize}r>{\fontsize{8}{9.6}\selectfont}c@{}}
\toprule
\multirow{2}{*}{\textbf{Category}} & \multirow{2}{*}{\textbf{Module}} & \multicolumn{2}{c}{\textbf{MSE}} & \multicolumn{2}{c}{\textbf{MAE}} & \multirow{2}{*}{\textbf{Time (s)}} \\
\cmidrule(lr){3-4}\cmidrule(lr){5-6}
 & & \textbf{value} & \(\boldsymbol{\Delta\%}\) & \textbf{value} & \(\boldsymbol{\Delta\%}\) & \\
\midrule
Gradient proxy & w/o output-cache proxy\textsuperscript{a} & 0.367 & \(+10.5\) & 0.407 & \(+5.6\) & 238.37 \\
\cmidrule(lr){1-7}
Layer selection & w/o layer selection & 0.343 & \(+3.0\) & 0.391 & \(+1.5\) & 338.14 \\
\cmidrule(lr){1-7}
\multirow{3}{*}{Variable pooling} & w/o conflict pooling & 0.340 & \(+2.2\) & 0.390 & \(+1.2\) & 265.51 \\
 & w/o anchor pooling & 0.344 & \(+3.4\) & 0.393 & \(+1.9\) & 252.61 \\
 & w/o variable pooling & 0.343 & \(+3.0\) & 0.392 & \(+1.7\) & 239.77 \\
\cmidrule(lr){1-7}
\multirow{2}{*}{Direction surgery} & w/o magnitude decoupling & 0.335 & \(+0.6\) & 0.387 & \(+0.3\) & 256.50 \\
 & w/o safe candidate selection & \underline{0.333} & \(+0.2\) & \underline{0.386} & \(+0.0\) & 258.56 \\
\midrule
\multicolumn{2}{c}{\textbf{PV-Surgery (full)}} & \textbf{0.333} & -- & \textbf{0.386} & -- & 274.73 \\
\bottomrule
\end{tabular*}
\par\vspace{0.1em}
{\raggedright\scriptsize\textsuperscript{a} Removing the output-cache proxy also disables cache-dependent layer selection, pooling, and candidate comparison.\par}
\end{table}

\begin{table}[p]
\centering
\caption{Core component ablation on ETTm1, averaged over 20 matched backbone--prediction-length settings. \(\Delta\) is the relative change against the full configuration. Bold indicates the best score and underlining the second best using unrounded averages.}
\label{tab:core_ablation_ettm1}
\vspace{1mm}
\footnotesize
\renewcommand{\arraystretch}{0.5}
\begin{tabular*}{\linewidth}{@{\extracolsep{\fill}}cl>{\fontsize{8}{9.6}\selectfont}r>{\scriptsize}r>{\fontsize{8}{9.6}\selectfont}r>{\scriptsize}r>{\fontsize{8}{9.6}\selectfont}c@{}}
\toprule
\multirow{2}{*}{\textbf{Category}} & \multirow{2}{*}{\textbf{Module}} & \multicolumn{2}{c}{\textbf{MSE}} & \multicolumn{2}{c}{\textbf{MAE}} & \multirow{2}{*}{\textbf{Time (s)}} \\
\cmidrule(lr){3-4}\cmidrule(lr){5-6}
 & & \textbf{value} & \(\boldsymbol{\Delta\%}\) & \textbf{value} & \(\boldsymbol{\Delta\%}\) & \\
\midrule
Gradient proxy & w/o output-cache proxy\textsuperscript{a} & 0.511 & \(+3.7\) & 0.474 & \(+2.3\) & 792.11 \\
\cmidrule(lr){1-7}
Layer selection & w/o layer selection & 0.496 & \(+0.7\) & 0.463 & \(+0.1\) & 1075.03 \\
\cmidrule(lr){1-7}
\multirow{3}{*}{Variable pooling} & w/o conflict pooling & \underline{0.493} & \(+0.1\) & 0.464 & \(+0.2\) & 902.63 \\
 & w/o anchor pooling & 0.498 & \(+1.0\) & 0.467 & \(+0.8\) & 915.37 \\
 & w/o variable pooling & 0.494 & \(+0.4\) & 0.465 & \(+0.4\) & 822.81 \\
\cmidrule(lr){1-7}
\multirow{2}{*}{Direction surgery} & w/o magnitude decoupling & 0.498 & \(+1.1\) & \underline{0.463} & \(+0.0\) & 906.55 \\
 & w/o safe candidate selection & 0.497 & \(+0.8\) & 0.464 & \(+0.3\) & 919.78 \\
\midrule
\multicolumn{2}{c}{\textbf{PV-Surgery (full)}} & \textbf{0.493} & -- & \textbf{0.463} & -- & 857.04 \\
\bottomrule
\end{tabular*}
\par\vspace{0.1em}
{\raggedright\scriptsize\textsuperscript{a} Removing the output-cache proxy also disables cache-dependent layer selection, pooling, and candidate comparison.\par}
\end{table}

\begin{table}[p]
\centering
\caption{Core component ablation on ETTm2, averaged over 20 matched backbone--prediction-length settings. \(\Delta\) is the relative change against the full configuration. Bold indicates the best score and underlining the second best using unrounded averages.}
\label{tab:core_ablation_ettm2}
\vspace{1mm}
\footnotesize
\renewcommand{\arraystretch}{0.5}
\begin{tabular*}{\linewidth}{@{\extracolsep{\fill}}cl>{\fontsize{8}{9.6}\selectfont}r>{\scriptsize}r>{\fontsize{8}{9.6}\selectfont}r>{\scriptsize}r>{\fontsize{8}{9.6}\selectfont}c@{}}
\toprule
\multirow{2}{*}{\textbf{Category}} & \multirow{2}{*}{\textbf{Module}} & \multicolumn{2}{c}{\textbf{MSE}} & \multicolumn{2}{c}{\textbf{MAE}} & \multirow{2}{*}{\textbf{Time (s)}} \\
\cmidrule(lr){3-4}\cmidrule(lr){5-6}
 & & \textbf{value} & \(\boldsymbol{\Delta\%}\) & \textbf{value} & \(\boldsymbol{\Delta\%}\) & \\
\midrule
Gradient proxy & w/o output-cache proxy\textsuperscript{a} & 0.226 & \(+1.4\) & 0.308 & \(+1.6\) & 925.66 \\
\cmidrule(lr){1-7}
Layer selection & w/o layer selection & 0.223 & \(+0.2\) & 0.303 & \(+0.1\) & 1126.85 \\
\cmidrule(lr){1-7}
\multirow{3}{*}{Variable pooling} & w/o conflict pooling & \underline{0.222} & \(-0.7\) & \underline{0.302} & \(-0.4\) & 984.73 \\
 & w/o anchor pooling & 0.228 & \(+2.2\) & 0.308 & \(+1.6\) & 1077.18 \\
 & w/o variable pooling & 0.223 & \(-0.0\) & 0.304 & \(+0.3\) & 968.47 \\
\cmidrule(lr){1-7}
\multirow{2}{*}{Direction surgery} & w/o magnitude decoupling & 0.222 & \(-0.5\) & 0.303 & \(-0.1\) & 1105.34 \\
 & w/o safe candidate selection & \textbf{0.221} & \(-0.9\) & \textbf{0.301} & \(-0.6\) & 1000.83 \\
\midrule
\multicolumn{2}{c}{\textbf{PV-Surgery (full)}} & 0.223 & -- & 0.303 & -- & 938.45 \\
\bottomrule
\end{tabular*}
\par\vspace{0.1em}
{\raggedright\scriptsize\textsuperscript{a} Removing the output-cache proxy also disables cache-dependent layer selection, pooling, and candidate comparison.\par}
\end{table}

\begin{table}[p]
\centering
\caption{Core component ablation on Weather, averaged over 20 matched backbone--prediction-length settings. \(\Delta\) is the relative change against the full configuration. Bold indicates the best score and underlining the second best using unrounded averages.}
\label{tab:core_ablation_weather}
\vspace{1mm}
\footnotesize
\renewcommand{\arraystretch}{0.5}
\begin{tabular*}{\linewidth}{@{\extracolsep{\fill}}cl>{\fontsize{8}{9.6}\selectfont}r>{\scriptsize}r>{\fontsize{8}{9.6}\selectfont}r>{\scriptsize}r>{\fontsize{8}{9.6}\selectfont}c@{}}
\toprule
\multirow{2}{*}{\textbf{Category}} & \multirow{2}{*}{\textbf{Module}} & \multicolumn{2}{c}{\textbf{MSE}} & \multicolumn{2}{c}{\textbf{MAE}} & \multirow{2}{*}{\textbf{Time (s)}} \\
\cmidrule(lr){3-4}\cmidrule(lr){5-6}
 & & \textbf{value} & \(\boldsymbol{\Delta\%}\) & \textbf{value} & \(\boldsymbol{\Delta\%}\) & \\
\midrule
Gradient proxy & w/o output-cache proxy\textsuperscript{a} & 0.301 & \(+2.8\) & 0.327 & \(+1.8\) & 1138.48 \\
\cmidrule(lr){1-7}
Layer selection & w/o layer selection & \textbf{0.291} & \(-0.6\) & \textbf{0.321} & \(-0.4\) & 1887.39 \\
\cmidrule(lr){1-7}
\multirow{3}{*}{Variable pooling} & w/o conflict pooling & 0.298 & \(+1.6\) & 0.326 & \(+1.3\) & 1212.13 \\
 & w/o anchor pooling & 0.307 & \(+4.8\) & 0.332 & \(+3.1\) & 1094.43 \\
 & w/o variable pooling & 0.308 & \(+4.9\) & 0.332 & \(+3.1\) & 993.80 \\
\cmidrule(lr){1-7}
\multirow{2}{*}{Direction surgery} & w/o magnitude decoupling & 0.301 & \(+2.7\) & 0.326 & \(+1.3\) & 1103.28 \\
 & w/o safe candidate selection & \underline{0.292} & \(-0.4\) & \underline{0.321} & \(-0.2\) & 1132.23 \\
\midrule
\multicolumn{2}{c}{\textbf{PV-Surgery (full)}} & 0.293 & -- & 0.322 & -- & 1286.21 \\
\bottomrule
\end{tabular*}
\par\vspace{0.1em}
{\raggedright\scriptsize\textsuperscript{a} Removing the output-cache proxy also disables cache-dependent layer selection, pooling, and candidate comparison.\par}
\end{table}

\begin{table}[p]
\centering
\caption{Core component ablation on Exchange, averaged over 20 matched backbone--prediction-length settings. \(\Delta\) is the relative change against the full configuration. Bold indicates the best score and underlining the second best using unrounded averages.}
\label{tab:core_ablation_exchange}
\vspace{1mm}
\footnotesize
\renewcommand{\arraystretch}{0.5}
\begin{tabular*}{\linewidth}{@{\extracolsep{\fill}}cl>{\fontsize{8}{9.6}\selectfont}r>{\scriptsize}r>{\fontsize{8}{9.6}\selectfont}r>{\scriptsize}r>{\fontsize{8}{9.6}\selectfont}c@{}}
\toprule
\multirow{2}{*}{\textbf{Category}} & \multirow{2}{*}{\textbf{Module}} & \multicolumn{2}{c}{\textbf{MSE}} & \multicolumn{2}{c}{\textbf{MAE}} & \multirow{2}{*}{\textbf{Time (s)}} \\
\cmidrule(lr){3-4}\cmidrule(lr){5-6}
 & & \textbf{value} & \(\boldsymbol{\Delta\%}\) & \textbf{value} & \(\boldsymbol{\Delta\%}\) & \\
\midrule
Gradient proxy & w/o output-cache proxy\textsuperscript{a} & 0.528 & \(+22.6\) & 0.486 & \(+7.8\) & 106.10 \\
\cmidrule(lr){1-7}
Layer selection & w/o layer selection & 0.461 & \(+7.0\) & 0.461 & \(+2.2\) & 172.68 \\
\cmidrule(lr){1-7}
\multirow{3}{*}{Variable pooling} & w/o conflict pooling & 0.516 & \(+19.8\) & 0.484 & \(+7.2\) & 97.14 \\
 & w/o anchor pooling & 0.436 & \(+1.2\) & 0.452 & \(+0.1\) & 101.68 \\
 & w/o variable pooling & \underline{0.434} & \(+0.7\) & \textbf{0.450} & \(-0.2\) & 101.73 \\
\cmidrule(lr){1-7}
\multirow{2}{*}{Direction surgery} & w/o magnitude decoupling & 0.448 & \(+3.9\) & 0.461 & \(+2.2\) & 113.14 \\
 & w/o safe candidate selection & 0.440 & \(+2.1\) & 0.457 & \(+1.3\) & 106.53 \\
\midrule
\multicolumn{2}{c}{\textbf{PV-Surgery (full)}} & \textbf{0.431} & -- & \underline{0.451} & -- & 115.19 \\
\bottomrule
\end{tabular*}
\par\vspace{0.1em}
{\raggedright\scriptsize\textsuperscript{a} Removing the output-cache proxy also disables cache-dependent layer selection, pooling, and candidate comparison.\par}
\end{table}

\begin{table}[p]
\centering
\caption{Core component ablation on ILI, averaged over 20 matched backbone--prediction-length settings. \(\Delta\) is the relative change against the full configuration. Bold indicates the best score and underlining the second best using unrounded averages.}
\label{tab:core_ablation_ili}
\vspace{1mm}
\footnotesize
\renewcommand{\arraystretch}{0.5}
\begin{tabular*}{\linewidth}{@{\extracolsep{\fill}}cl>{\fontsize{8}{9.6}\selectfont}r>{\scriptsize}r>{\fontsize{8}{9.6}\selectfont}r>{\scriptsize}r>{\fontsize{8}{9.6}\selectfont}c@{}}
\toprule
\multirow{2}{*}{\textbf{Category}} & \multirow{2}{*}{\textbf{Module}} & \multicolumn{2}{c}{\textbf{MSE}} & \multicolumn{2}{c}{\textbf{MAE}} & \multirow{2}{*}{\textbf{Time (s)}} \\
\cmidrule(lr){3-4}\cmidrule(lr){5-6}
 & & \textbf{value} & \(\boldsymbol{\Delta\%}\) & \textbf{value} & \(\boldsymbol{\Delta\%}\) & \\
\midrule
Gradient proxy & w/o output-cache proxy\textsuperscript{a} & \underline{3.05} & \(-0.4\) & \textbf{1.18} & \(-0.7\) & 12.80 \\
\cmidrule(lr){1-7}
Layer selection & w/o layer selection & \textbf{3.04} & \(-0.8\) & \underline{1.19} & \(-0.5\) & 19.48 \\
\cmidrule(lr){1-7}
\multirow{3}{*}{Variable pooling} & w/o conflict pooling & 3.08 & \(+0.4\) & 1.20 & \(+0.4\) & 15.07 \\
 & w/o anchor pooling & 3.09 & \(+0.7\) & 1.20 & \(+0.3\) & 15.03 \\
 & w/o variable pooling & 3.09 & \(+0.7\) & 1.20 & \(+0.3\) & 14.43 \\
\cmidrule(lr){1-7}
\multirow{2}{*}{Direction surgery} & w/o magnitude decoupling & 3.13 & \(+2.0\) & 1.22 & \(+2.0\) & 13.79 \\
 & w/o safe candidate selection & 3.09 & \(+0.7\) & 1.20 & \(+0.6\) & 14.12 \\
\midrule
\multicolumn{2}{c}{\textbf{PV-Surgery (full)}} & 3.07 & -- & 1.19 & -- & 15.21 \\
\bottomrule
\end{tabular*}
\par\vspace{0.1em}
{\raggedright\scriptsize\textsuperscript{a} Removing the output-cache proxy also disables cache-dependent layer selection, pooling, and candidate comparison.\par}
\end{table}

\clearpage
\section{Multi-Task Gradient Operators}
\label{app:mtl_comparison}
Multi-task optimization offers several ways to combine gradients that disagree across objectives. We apply five operators discussed in Section~\ref{sec:related_work} to the variable-wise proxies constructed by PV-Surgery. These methods normally act on a small set of task gradients exposed directly by a multi-task model, while a forecaster combines \(D\) variable objectives into one scalar loss. This comparison asks whether native multi-task operators can use the reconstructed variable-wise rows without PV-specific pooling or rescaling. Its purpose is to test transferability rather than establish a definitive ranking.

\paragraph{Native-operator protocol.}
Each native operator receives the same unpooled variable-wise gradient proxies on the selected subspace. We hold output-cache reconstruction, effective-coverage layer selection, selected-layer concatenation, and full-sum assembly fixed. Pooling, candidate selection, magnitude restoration, and PV-specific wrappers or rescaling are disabled, so each operator retains its native aggregation rule. Table~\ref{tab:mtl_micn} reports the \(28\) matched MICN settings from seven datasets and four prediction lengths using seed \(42\). Standard MSE training and the unchanged full PV-Surgery configuration provide the two reference rows.

\paragraph{The operators.}
\textbf{MGDA} treats the objectives as a multi-objective problem and takes the minimum-norm point of
the convex hull of their gradients, \(\min_{\alpha\in\Delta}\|\sum_i\alpha_ig_i\|_2^2\), which descends on every objective at once unless that minimum is zero, in which case the current point is already Pareto stationary \citep{sener2018multi}. \textbf{PCGrad} works pair by
pair. For each objective, it compares its gradient with the others in random order and, whenever the cosine is negative,
projects the gradient onto the normal plane of the conflicting one, leaving non-conflicting pairs
untouched \citep{yu2020gradient}. \textbf{GradVac} generalizes PCGrad's rule from ``repair negative cosines'' to ``reach a target cosine'', tracking an exponential moving average of each pair's
observed similarity and using it as the target, so PCGrad is the special case of a zero target
applied only under conflict \citep{wang2021gradient}. \textbf{CAGrad} maximizes the worst-case
objective improvement inside a ball of radius \(c\|\bar g\|\) around the average gradient
\(\bar g=D^{-1}\sum_i g_i\), which keeps
the update anchored to the mean-loss direction, recovers plain descent at \(c=0\), and approaches MGDA as
\(c\to\infty\) \citep{liu2021conflict}. We use \(c=0.5\). \textbf{Aligned-MTL} reads the condition number of the gradient
matrix as a joint measure of conflict and dominance, eigendecomposes the objective-by-objective Gram
matrix, and rescales its singular values to the smallest one, leaving principal components orthogonal
and equal in magnitude \citep{senushkin2023independent}. 

\paragraph{Native-operator settings.}
PCGrad and GradVac visit peers in random order, and GradVac uses an exponential-moving-average coefficient of \(0.01\). MGDA runs at most \(250\) minimum-norm iterations with tolerance \(10^{-5}\). CAGrad uses \(c=0.5\), rescale mode 1, and SLSQP with at most \(250\) iterations and tolerance \(10^{-10}\). Aligned-MTL uses the smallest positive singular value and has no tuned hyperparameter.

The operators differ in both direction and aggregation scale. PCGrad and GradVac sum their corrected rows, MGDA keeps its minimum-norm convex-combination scale, CAGrad uses its mean-anchored native rescaling, and Aligned-MTL uses its spectral scale. We preserve these native rules, so this is a method-level comparison rather than a direction-only ablation. Our common-direction operator restores the norm of each surgery input after changing its direction. The inputs are variable rows in the unpooled branch and pool rows in the pooled branch, so this restoration does not preserve every individual variable-row norm after pooling.

\begin{table}[H]
\centering
\caption{Native multi-task gradient operators on MICN over seven benchmarks and four
prediction lengths. Each native row receives the same unpooled variable-wise proxies on the
selected subspace and retains its own aggregation rule. PV-Surgery uses the unchanged full
configuration. The MSE row denotes standard MSE training. Best results are bolded and second-best
results are underlined using unrounded scores.}
\label{tab:mtl_micn}
\setlength{\tabcolsep}{2.6pt}
\renewcommand{\arraystretch}{1.02}
\scriptsize
\resizebox{\textwidth}{!}{%
\begin{tabular}{@{}c|c|rrrr|rrrr|rrrr|rrrr|rrrr|rrrr|rrrr@{}}
\toprule
\multicolumn{2}{c|}{Dataset} & \multicolumn{4}{c|}{ETTh1} & \multicolumn{4}{c|}{ETTh2} & \multicolumn{4}{c|}{ETTm1} & \multicolumn{4}{c|}{ETTm2} & \multicolumn{4}{c|}{Weather} & \multicolumn{4}{c|}{Exchange} & \multicolumn{4}{c}{ILI} \\
\midrule
\multicolumn{2}{c|}{Forecast length} & 96 & 192 & 336 & \multicolumn{1}{c|}{720} & 96 & 192 & 336 & \multicolumn{1}{c|}{720} & 96 & 192 & 336 & \multicolumn{1}{c|}{720} & 96 & 192 & 336 & \multicolumn{1}{c|}{720} & 96 & 192 & 336 & \multicolumn{1}{c|}{720} & 96 & 192 & 336 & \multicolumn{1}{c|}{720} & 24 & 36 & 48 & \multicolumn{1}{c}{60} \\
\midrule
\multirow{2}{*}{\textbf{MSE}} & MSE & 0.552 & 0.594 & \textbf{0.688} & \underline{0.753} & 0.238 & 0.315 & 0.452 & 0.672 & 0.445 & 0.495 & 0.567 & 0.627 & 0.166 & 0.202 & 0.266 & 0.389 & 0.398 & 0.403 & \textbf{0.380} & \underline{0.496} & 0.123 & 0.219 & 0.498 & 2.931 & \textbf{3.169} & 2.996 & 3.269 & \underline{3.273} \\
 & MAE & 0.530 & 0.553 & \textbf{0.623} & \underline{0.671} & 0.333 & 0.388 & 0.472 & 0.583 & 0.472 & 0.511 & 0.570 & 0.596 & 0.276 & 0.303 & 0.355 & 0.434 & 0.442 & 0.447 & \textbf{0.422} & \underline{0.501} & 0.262 & 0.352 & 0.525 & 1.400 & \textbf{1.237} & 1.228 & 1.284 & \underline{1.267} \\
\midrule
\multirow{2}{*}{\textbf{MGDA}} & MSE & 0.497 & 0.641 & 0.718 & 0.790 & 0.236 & 0.295 & \textbf{0.349} & \textbf{0.436} & 0.447 & 0.473 & \textbf{0.495} & \textbf{0.551} & 0.151 & \textbf{0.189} & 0.237 & 0.320 & 0.355 & \textbf{0.375} & \underline{0.385} & 0.555 & 0.122 & 0.226 & 0.401 & \textbf{0.857} & 3.454 & 2.991 & \textbf{3.224} & 3.394 \\
 & MAE & \underline{0.496} & 0.599 & 0.641 & 0.697 & 0.324 & 0.370 & \underline{0.413} & \textbf{0.475} & 0.472 & 0.487 & \underline{0.494} & \textbf{0.538} & 0.253 & \textbf{0.280} & \underline{0.315} & 0.383 & 0.398 & \textbf{0.424} & \underline{0.426} & 0.539 & 0.261 & 0.361 & 0.491 & \textbf{0.743} & 1.300 & \textbf{1.215} & \textbf{1.271} & 1.286 \\
\midrule
\multirow{2}{*}{\textbf{PCGrad}} & MSE & 0.497 & 0.559 & 0.770 & 0.773 & \underline{0.230} & \underline{0.291} & 0.365 & 0.603 & 0.427 & 0.475 & 0.517 & 0.599 & \underline{0.147} & 0.193 & 0.237 & \textbf{0.295} & \underline{0.312} & 0.446 & 0.430 & 0.523 & 0.108 & 0.217 & 0.386 & 1.143 & 3.483 & \underline{2.983} & 3.234 & 3.324 \\
 & MAE & 0.502 & 0.542 & 0.671 & 0.686 & \underline{0.321} & \textbf{0.366} & 0.424 & 0.562 & 0.454 & 0.489 & 0.522 & 0.582 & 0.251 & 0.291 & 0.324 & \textbf{0.361} & \underline{0.378} & 0.471 & 0.459 & 0.525 & 0.242 & 0.352 & \underline{0.475} & 0.822 & 1.306 & 1.221 & 1.274 & 1.271 \\
\midrule
\multirow{2}{*}{\textbf{GradVac}} & MSE & 0.495 & \underline{0.556} & 0.767 & \textbf{0.728} & 0.233 & 0.296 & 0.369 & 0.600 & 0.434 & \underline{0.468} & 0.513 & 0.587 & 0.148 & 0.195 & 0.233 & \underline{0.296} & \textbf{0.304} & 0.422 & 0.424 & 0.512 & \underline{0.108} & \underline{0.215} & \underline{0.385} & 1.142 & 3.477 & \textbf{2.980} & 3.232 & 3.321 \\
 & MAE & 0.500 & \underline{0.540} & 0.670 & \textbf{0.656} & 0.325 & 0.371 & 0.427 & 0.561 & 0.462 & \underline{0.485} & 0.519 & 0.573 & 0.253 & 0.290 & 0.319 & \textbf{0.361} & \textbf{0.370} & 0.449 & 0.458 & 0.518 & \underline{0.241} & \underline{0.350} & 0.479 & 0.814 & 1.305 & 1.220 & 1.273 & 1.272 \\
\midrule
\multirow{2}{*}{\textbf{CAGrad}} & MSE & 0.510 & \textbf{0.555} & 0.770 & 0.769 & 0.234 & 0.318 & 0.386 & 0.647 & \underline{0.419} & 0.484 & 0.515 & 0.604 & 0.151 & 0.209 & 0.243 & 0.303 & 0.341 & 0.447 & 0.437 & 0.512 & 0.115 & 0.220 & 0.427 & 1.399 & 3.281 & 2.985 & 3.233 & \textbf{3.185} \\
 & MAE & 0.506 & \textbf{0.539} & 0.671 & 0.682 & 0.325 & 0.392 & 0.451 & 0.587 & \underline{0.448} & 0.498 & 0.521 & 0.588 & 0.256 & 0.305 & 0.329 & \underline{0.372} & 0.402 & 0.470 & 0.464 & 0.517 & 0.251 & 0.354 & 0.491 & 0.870 & 1.261 & 1.219 & 1.274 & \textbf{1.240} \\
\midrule
\multirow{2}{*}{\textbf{Aligned-MTL}} & MSE & \textbf{0.482} & 0.587 & 0.721 & 0.810 & 0.232 & 0.295 & 0.360 & 0.601 & \textbf{0.417} & \textbf{0.447} & \underline{0.496} & 0.578 & 0.148 & \underline{0.189} & \textbf{0.228} & 0.393 & 0.401 & \underline{0.378} & 0.464 & 0.513 & \textbf{0.105} & 0.245 & 0.436 & 2.729 & 3.430 & 2.984 & 3.251 & 3.429 \\
 & MAE & \textbf{0.487} & 0.566 & 0.645 & 0.708 & 0.321 & 0.374 & 0.418 & 0.563 & \textbf{0.448} & \textbf{0.458} & \textbf{0.493} & \underline{0.558} & \underline{0.250} & \underline{0.283} & \textbf{0.310} & 0.445 & 0.428 & \underline{0.426} & 0.483 & 0.518 & \textbf{0.236} & 0.378 & 0.501 & 1.367 & 1.293 & \underline{1.216} & 1.278 & 1.299 \\
\midrule
\multirow{2}{*}{\shortstack{\textbf{PV-Surgery}\\\textbf{(Ours)}}} & MSE & \underline{0.492} & 0.561 & \underline{0.700} & 0.798 & \textbf{0.230} & \textbf{0.291} & \underline{0.352} & \underline{0.513} & 0.430 & 0.470 & 0.508 & \underline{0.576} & \textbf{0.146} & 0.190 & \underline{0.231} & 0.331 & 0.318 & 0.435 & 0.436 & \textbf{0.479} & 0.109 & \textbf{0.212} & \textbf{0.378} & \underline{1.106} & \underline{3.190} & 2.985 & \underline{3.228} & 3.364 \\
 & MAE & 0.497 & 0.543 & \underline{0.630} & 0.700 & \textbf{0.320} & \underline{0.369} & \textbf{0.412} & \underline{0.520} & 0.460 & 0.487 & 0.513 & 0.562 & \textbf{0.248} & 0.285 & 0.316 & 0.389 & 0.383 & 0.460 & 0.468 & \textbf{0.493} & 0.242 & \textbf{0.348} & \textbf{0.470} & \underline{0.793} & \underline{1.243} & 1.220 & \underline{1.272} & 1.283 \\
\bottomrule
\end{tabular}%
}
\end{table}

All five native operators reduce MSE on average relative to standard training over the \(28\) settings. The mean per-setting reductions are \(7.59\%\) for MGDA, \(6.87\%\) for PCGrad, \(7.55\%\) for GradVac, \(4.82\%\) for CAGrad, and \(3.51\%\) for Aligned-MTL. The unchanged PV-Surgery configuration gives the largest average reduction at \(8.50\%\). The native operators outperform standard training in \(19\), \(21\), \(22\), \(19\), and \(19\) settings, respectively, but each also outperforms PV-Surgery in between \(5\) and \(11\) settings. The comparison therefore does not establish a uniform ranking. Instead, the average gains of all five operators show that the reconstructed variable-wise proxies provide a meaningful training signal that methods designed for explicit task gradients can use, while the per-setting differences show that the way this signal is aggregated still affects the outcome.

\section{Mechanism Diagnostics}
\label{app:mechanism}
Test error alone does not show whether the internal steps of PV-Surgery behave as intended. We therefore examine the variable-wise gradient proxies reconstructed from cached layer signals and the optimizer update assembled from these proxies. The first analysis measures proxy fidelity by comparing the reconstructed proxy rows with exact gradients obtained by separate per-variable backward passes. The second measures pairwise conflict and descent violations before and after surgery. A descent violation occurs when the update passed to the optimizer has a negative inner product with a reconstructed variable gradient, indicating that a small step along the negative update would increase that variable's loss to first order. A separate test-set analysis then asks whether PV-Surgery narrows the oracle gaps observed under standard shared training.

\paragraph{Measurement protocol.}
We run two diagnostic versions of the \(140\) settings in Table~\ref{tab:backbone_pv}. Both use the full PV-Surgery configuration and the same training protocol as the reported runs. They enable different measurements at fixed intervals, but the diagnostic results are never used to form the optimizer update. For proxy fidelity, which measures directional agreement between a proxy and its exact gradient, we compute exact per-variable gradients every twenty optimizer steps and compare them with the reconstructed proxy rows using cosine similarity. These additional backward passes provide exact reference gradients only for measurement, while training continues to use the proxy-based update. We report fidelity at the output boundary, over all hooked parameters \(\Omega_{\mathcal{H}}\), and on the selected subspace \(\Omega_{\mathcal{S}}\). A separate set of runs records conflict mass, the average negative-cosine magnitude across reconstructed variable pairs, and descent violations every fifty steps.

At \(12{,}256\) of the \(45{,}897\) logged steps, the selected-slice proxy rows have zero norm and the conditional gate skips the intervention. Cosine similarity is undefined for these zero vectors. Our primary selected-subspace statistic is therefore conditional on a nonzero selected-slice proxy. Consequently, five runs have no defined selected-subspace cosine after conditioning, leaving \(135\) runs for the primary statistic. If the zero entries are instead assigned a cosine of zero, the selected-subspace average is \(0.739\) over all \(140\) runs. The output-boundary and all-parameter statistics use all runs without this conditioning.

\begin{figure}[H]
\centering
\includegraphics[width=0.92\linewidth]{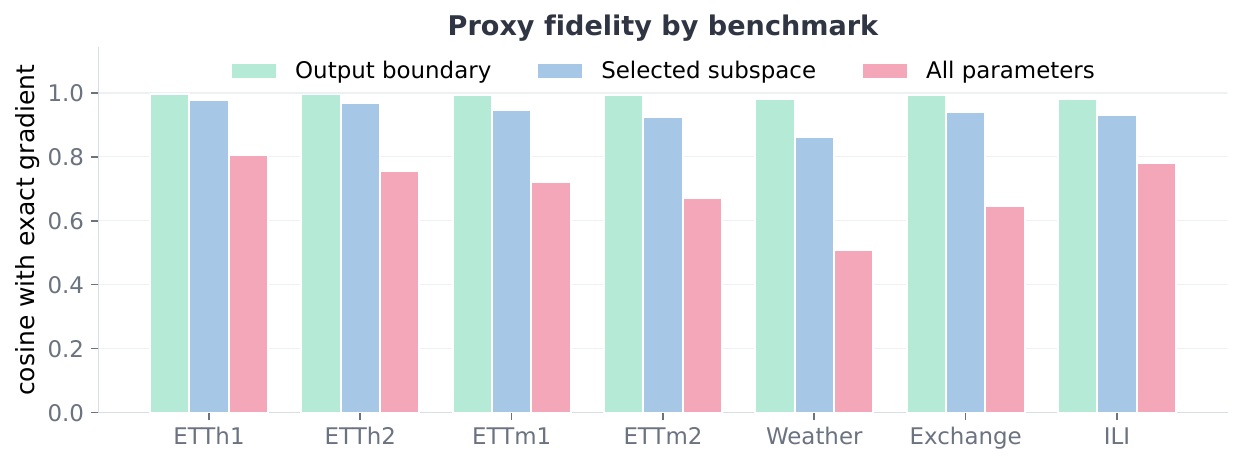}
\caption{Cosine similarity between reconstructed proxy rows and exact per-variable gradients for each benchmark. Each bar first averages logged steps within a run and then averages the available runs for that benchmark. Each benchmark contributes \(20\) runs from five backbones and four prediction lengths. Output-boundary and all-parameter bars use all \(20\) runs. Selected-subspace bars use \(17\) Exchange runs, \(18\) ILI runs, and all \(20\) runs elsewhere after conditioning on nonzero proxy entries. The smaller counts occur when every logged selected-slice proxy in a run has zero norm, leaving no defined selected-subspace cosine.}
\label{fig:fidelity_dataset}
\end{figure}

\paragraph{Proxy fidelity.}
The average cosine with the exact gradient is \(0.990\) at the output boundary, \(0.936\) on the selected subspace when its proxy is nonzero, and \(0.698\) over all parameters. Figure~\ref{fig:fidelity_dataset} shows the same conditional comparison for each benchmark. Output-boundary fidelity ranges from \(0.981\) to \(0.996\). The gap is larger inside the backbone, where variable-mixing operations make the cache an approximation. Weather has the lowest all-parameter fidelity at \(0.508\), which rises to \(0.862\) after layer selection. ETTh1 rises from \(0.806\) to \(0.977\). The conditional selected-subspace value exceeds the all-parameter value on every benchmark. Layer selection therefore concentrates active interventions on parameters for which the reconstructed gradients are more faithful, although the selected rows are not exact in every model.

\begin{figure}[H]
\centering
\setlength{\tabcolsep}{1.5pt}
\begin{tabular}{@{}c@{\hspace{0.02\linewidth}}c@{}}
\includegraphics[width=0.47\linewidth]{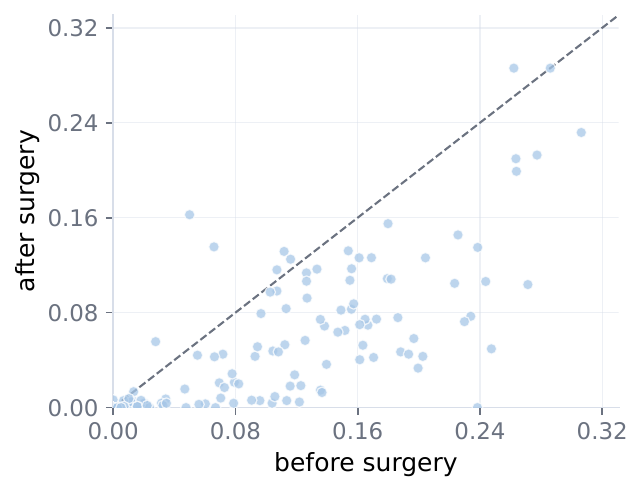}
&
\includegraphics[width=0.47\linewidth]{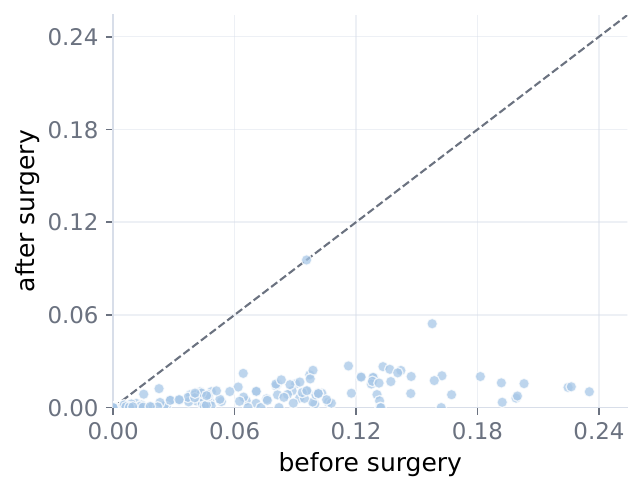}
\\[-1mm]
{\small (a) Descent-violation ratio.}
&
{\small (b) Conflict mass.}
\end{tabular}
\caption{Descent violations and conflict mass before and after surgery over the \(140\) matched settings. (a) shows the descent-violation ratio, and (b) shows conflict mass. Each point is one run, with the horizontal and vertical axes giving the values before and after surgery. Points below the diagonal indicate a reduction.}
\label{fig:mechanism_scatter}
\end{figure}

\paragraph{Effect on the reconstructed gradients.}
Figure~\ref{fig:mechanism_scatter} compares the sum-loss update with the final update in each of the \(140\) mechanism runs. Conflict mass is the mean of \(\max(-\cos_{ij},0)\) over all variable pairs, so a cosine of \(-0.4\) contributes \(0.4\) while a nonnegative cosine contributes zero. A larger conflict mass therefore indicates stronger overall directional opposition among the reconstructed variable gradients. The descent violation ratio is the fraction of reconstructed variable gradients that have a negative inner product with the update. A ratio of \(0.2\), for example, means that a small step along the negative update would increase the losses of two out of ten reconstructed variable objectives to first order. These are optimization diagnostics defined on the reconstructed rows. They should not be read as direct measurements of test-set harm.

Mean conflict mass falls from \(0.075\) to \(0.009\). It decreases in \(129\) runs, remains unchanged in \(11\), and increases in none. This one-sided shift directly verifies the intended operation of PV-Surgery. The mean descent violation ratio falls from \(0.097\) to \(0.049\). It decreases in \(103\) runs, remains unchanged in \(29\), and increases in \(8\). Descent violations are a stricter downstream test because they also depend on how the corrected rows combine into the final update. Their mean is nearly halved and they decrease in most runs, showing that conflict correction usually carries through to an update that is better aligned with the reconstructed variable objectives.

\begin{figure}[H]
\centering
\setlength{\tabcolsep}{1.5pt}
\begin{tabular}{@{}c@{\hspace{0.02\linewidth}}c@{}}
\includegraphics[width=0.54\linewidth]{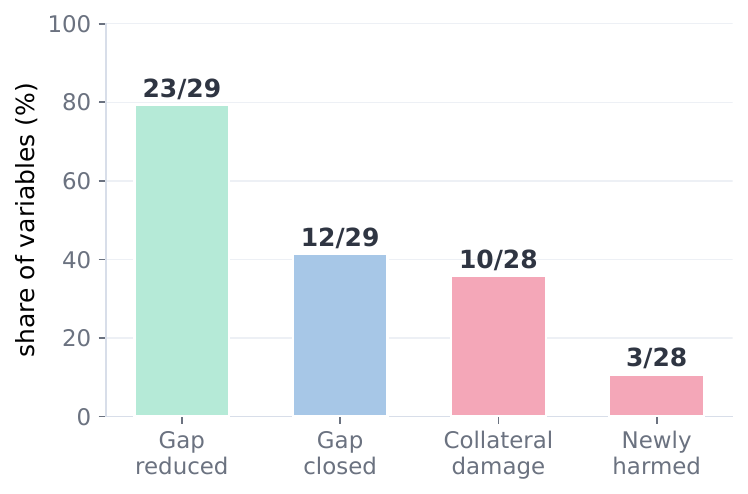}
&
\includegraphics[width=0.40\linewidth]{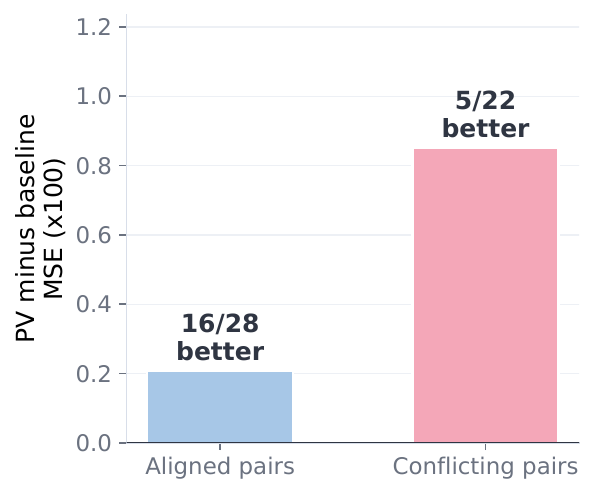}
\\[-1mm]
{\small (a) Effect on the Section 3 oracle gap.}
&
{\small (b) Frozen variable-pair subsets.}
\end{tabular}
\caption{Test-set evidence for the effect of PV-Surgery. (a) reports changes in the oracle gap. (b) reports PV-Surgery MSE minus baseline MSE on fixed variable-pair subsets. Positive values in (b) favor the baseline.}
\label{fig:oracle_gap}
\end{figure}

\paragraph{Test-set oracle gap.}
Figure~\ref{fig:oracle_gap}(a) returns to the oracle gap from Section~\ref{sec:diagnosis}. The comparison contains \(57\) iTransformer variables from the six benchmarks with matched baseline, PV-Surgery, and full-input single-target oracle runs. Shared training harms \(29\) of these variables. PV-Surgery reduces the gap for \(23\) and closes it for \(12\). Among the remaining \(28\) variables, whose baseline gap is not positive, PV-Surgery increases test MSE for \(10\) and creates a positive oracle gap for \(3\). Across all \(57\) variables, the mean gap decreases by \(0.0040\) MSE. The aggregate improvement therefore includes both repaired gaps and a smaller number of regressions.

\paragraph{Frozen variable-pair subsets.}
Figure~\ref{fig:oracle_gap}(b) tests a stricter hypothesis using separately trained fixed variable-pair subsets. PV-Surgery improves test MSE on only \(5\) of \(22\) conflicting pairs, compared with \(16\) of \(28\) aligned pairs. Its mean MSE change relative to the baseline is also unfavorable for both groups, at \(+0.0085\) for conflicting pairs and \(+0.0021\) for aligned pairs. Persistent pairwise conflict is therefore not a reliable marker of the subsets on which surgery will help. This negative result agrees with Section~\ref{sec:diagnosis}, where conflict identifies disagreement but not harm under shared training.

\section{Qualitative Case Studies}
\label{app:qualitative}
Aggregate MSE does not show how two forecasts differ within a window. We therefore inspect four settings with prediction length \(96\), namely MICN on ETTm2, iTransformer on ETTm2, MICN on ETTh1, and DLinear on ETTh2. Together they cover three backbones and two ETT benchmarks. Each figure shows the six load variables HUFL, HULL, MUFL, MULL, LUFL, and LULL together with oil temperature OT.

\paragraph{Protocol.}
For each setting, the baseline and PV-Surgery runs use the same data split and test windows. Their saved ground-truth arrays are identical. Since the variables span different vertical ranges, we summarize the error in each panel using a range-normalized mean absolute error. For window \(n\) and variable \(d\), let \(\hat y^{\mathrm{MSE}}\) and \(\hat y^{\mathrm{PV}}\) denote the forecasts from standard MSE training and PV-Surgery. For either forecast \(\hat y\), the panel distance is
\begin{equation}
\label{eq:panel_distance}
    \delta_{n,d}(\hat y)
    =\frac{\frac{1}{H}\sum_{h=1}^{H}\bigl|\hat y_{n,h,d}-y_{n,h,d}\bigr|}
          {\max\{y,\hat y^{\mathrm{MSE}},\hat y^{\mathrm{PV}}\}_{n,\cdot,d}
           -\min\{y,\hat y^{\mathrm{MSE}},\hat y^{\mathrm{PV}}\}_{n,\cdot,d}}.
\end{equation}
The denominator is the range spanned by the ground truth and both forecasts within the same panel. A value of \(0.10\) means that the average absolute error is one tenth of this range. The plotted vertical axes retain the standardized target and forecast values. We use \(\delta_{n,d}\) only as a scale-normalized scalar summary to compare the two forecasts within the same window and variable. The smaller of \(\delta_{n,d}(\hat y^{\mathrm{MSE}})\) and \(\delta_{n,d}(\hat y^{\mathrm{PV}})\) identifies the forecast with the lower average vertical error relative to their common displayed range. We also report the cosine between the predicted and true trajectories after subtracting their respective horizon means. This direction cosine measures temporal shape independently of the mean level. A negative value indicates that the predicted trajectory moves in the opposite direction from the target.

\paragraph{Window selection.}
We manually select one representative example from each setting in which the effect of PV-Surgery is clearly visible. Together, the examples illustrate level correction, reduced drift, and cases in which panel distance improves without a higher direction cosine. They are descriptive and do not enter any aggregate estimate.

\begin{figure}[H]
\centering
\includegraphics[width=\linewidth]{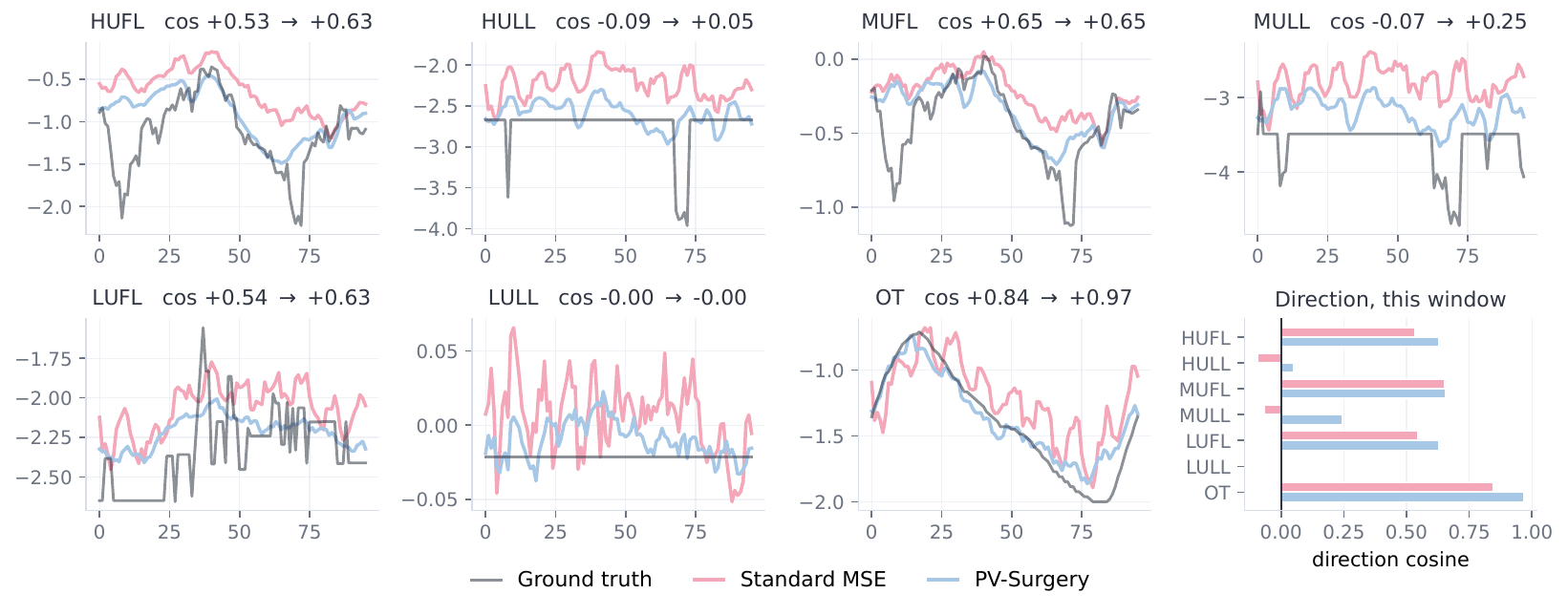}
\caption{MICN on ETTm2. Panel titles report the direction cosine for standard MSE training and PV-Surgery. The final panel collects the same values.}
\label{fig:qual_micn_ettm2}
\end{figure}

\paragraph{MICN on ETTm2.}
Across the full test set, PV-Surgery lowers MSE by \(12.52\%\) and raises the mean direction cosine from \(0.422\) to \(0.510\). Window-level MSE improves in \(19{,}377\) of \(23{,}409\) windows. The number of variables with negative direction cosine decreases in \(7{,}726\) windows and increases in \(2{,}534\). In Figure~\ref{fig:qual_micn_ettm2}, the baseline forecast is generally shifted above the ground truth, while PV-Surgery reduces this offset. The panel distance falls from \(0.25\) to \(0.13\) for HUFL, from \(0.24\) to \(0.10\) for HULL, and from \(0.35\) to \(0.18\) for MULL. HULL and MULL also change from negative to positive direction cosine. OT has the smallest corrected distance at \(0.08\), and its direction cosine rises to \(0.97\).

\begin{figure}[H]
\centering
\includegraphics[width=\linewidth]{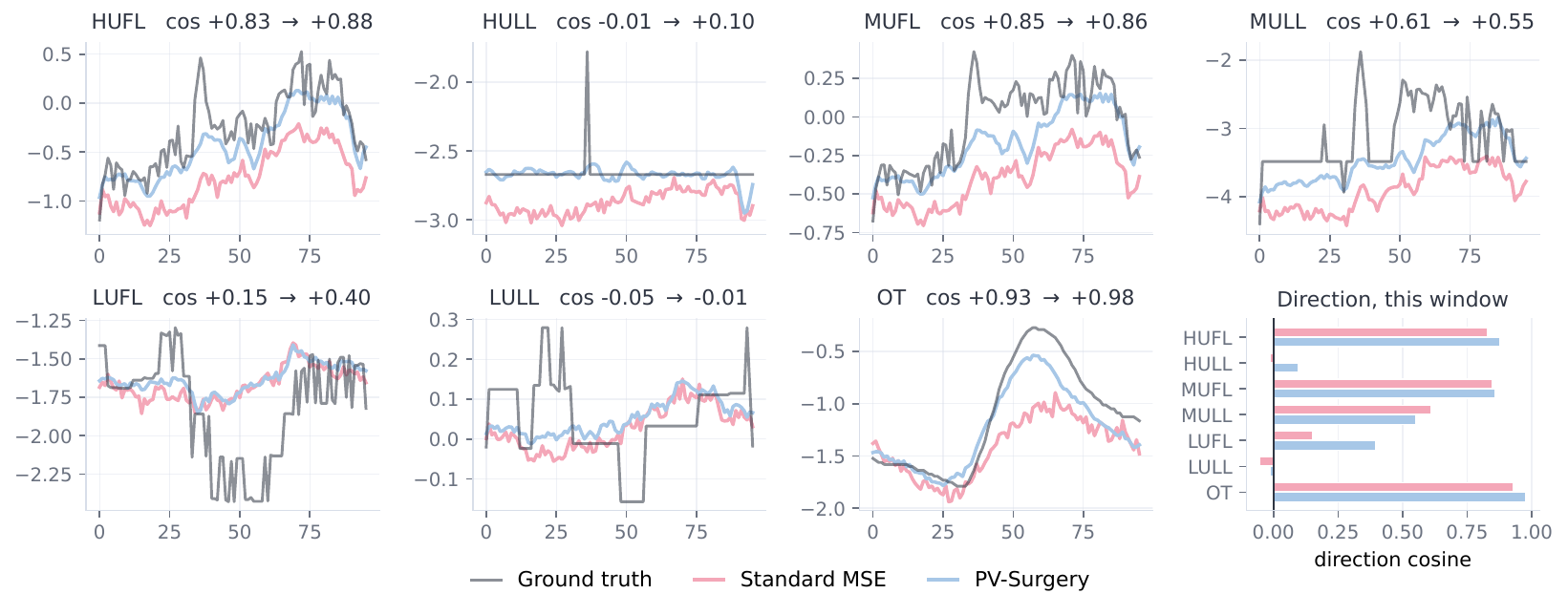}
\caption{iTransformer on ETTm2. Panel titles report the direction cosine for standard MSE training and PV-Surgery. The final panel collects the same values.}
\label{fig:qual_itr_ettm2}
\end{figure}

\paragraph{iTransformer on ETTm2.}
On the same dataset, iTransformer obtains a smaller MSE reduction of \(4.61\%\). Window-level MSE improves in \(15{,}752\) of \(23{,}409\) windows. Negative direction cosines become less frequent in \(5{,}020\) windows and more frequent in \(4{,}671\). Figure~\ref{fig:qual_itr_ettm2} again shows a substantial level correction. The panel distance falls from \(0.28\) to \(0.11\) for HUFL, from \(0.32\) to \(0.14\) for MUFL, and from \(0.17\) to \(0.03\) for HULL. The improvement is small for LUFL and LULL. MULL provides a useful counterexample. Its distance decreases from \(0.31\) to \(0.16\), while its direction cosine decreases from \(0.61\) to \(0.55\). A forecast can therefore move closer in level without matching the temporal shape more closely.

\begin{figure}[H]
\centering
\includegraphics[width=\linewidth]{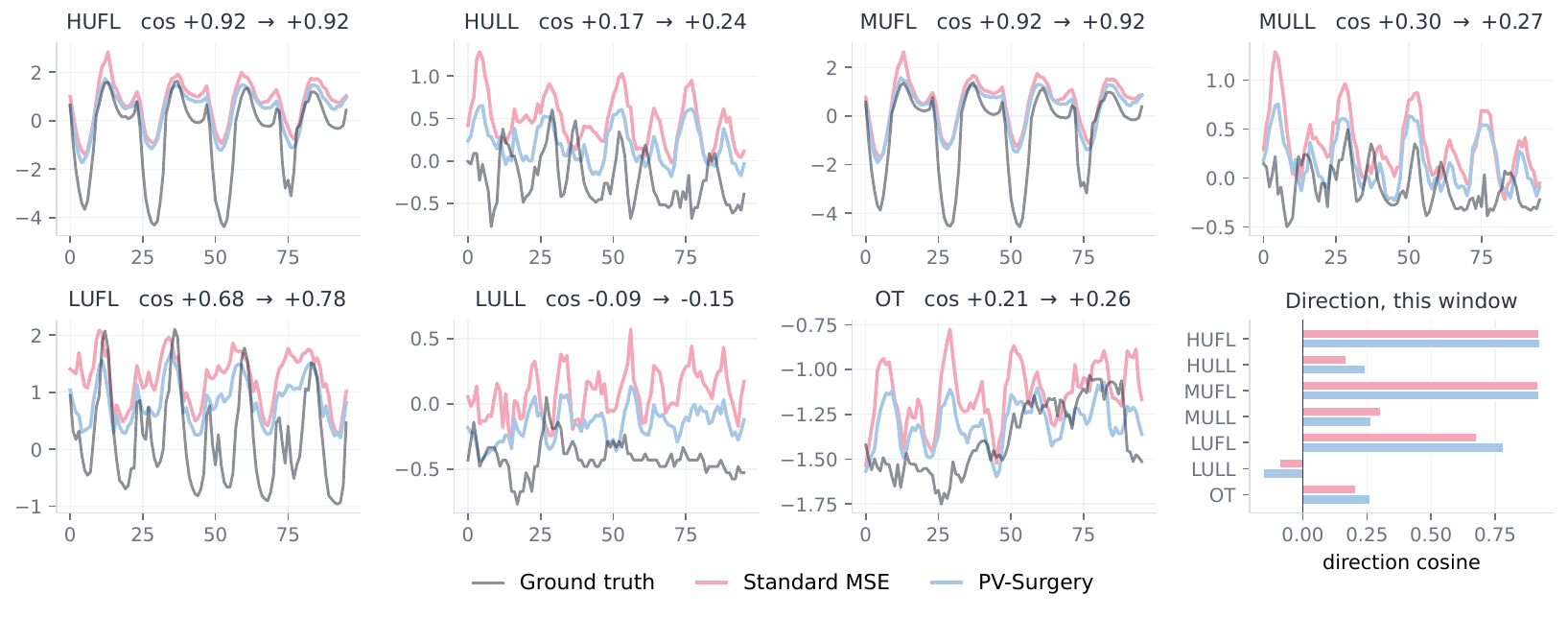}
\caption{MICN on ETTh1. Panel titles report the direction cosine for standard MSE training and PV-Surgery. The final panel collects the same values.}
\label{fig:qual_micn_etth1}
\end{figure}

\paragraph{MICN on ETTh1.}
PV-Surgery lowers MSE by \(10.85\%\) for MICN on ETTh1. It improves window-level MSE in \(4{,}188\) of \(5{,}709\) windows. The number of negative direction cosines decreases in \(1{,}156\) windows and increases in \(221\). In Figure~\ref{fig:qual_micn_etth1}, the baseline tends to overshoot the target and PV-Surgery moves the forecasts downward. The largest distance reductions occur for LULL, from \(0.37\) to \(0.21\), and LUFL, from \(0.35\) to \(0.24\). HUFL and MUFL retain direction cosines near \(0.92\). Direction does not improve for every variable. MULL falls from \(0.30\) to \(0.27\), and LULL becomes more negative, even though both panel distances decrease.

\begin{figure}[H]
\centering
\includegraphics[width=\linewidth]{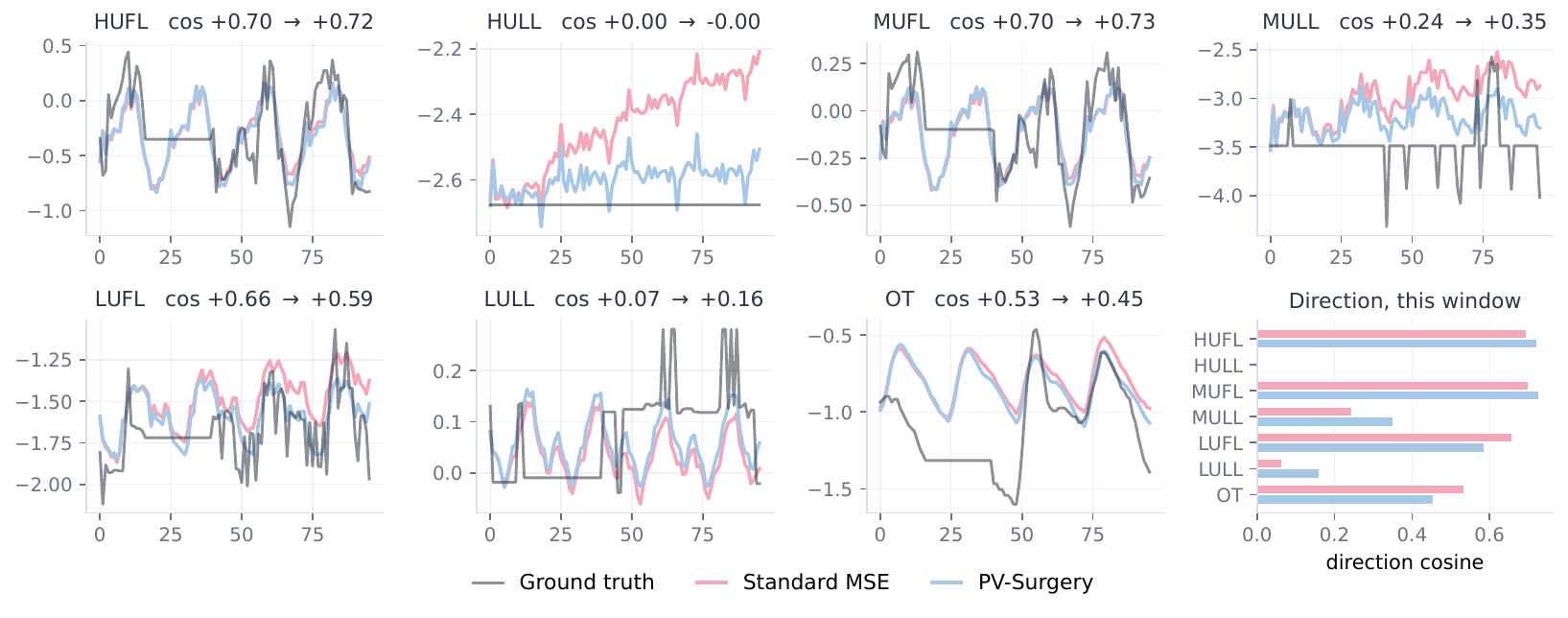}
\caption{DLinear on ETTh2. Panel titles report the direction cosine for standard MSE training and PV-Surgery. The final panel collects the same values.}
\label{fig:qual_dlinear_etth2}
\end{figure}

\paragraph{DLinear on ETTh2.}
PV-Surgery lowers MSE by \(4.55\%\) in the DLinear setting. Window-level MSE decreases in \(4{,}174\) of \(5{,}709\) windows. Negative direction cosines become less frequent in \(1{,}073\) windows and more frequent in \(264\). The clearest change in Figure~\ref{fig:qual_dlinear_etth2} occurs for HULL. The baseline drifts away from an almost flat target, while the PV-Surgery forecast remains closer throughout the horizon. Its panel distance falls from \(0.45\) to \(0.15\). MULL shows a smaller reduction from \(0.28\) to \(0.18\). LUFL and OT again show that the two diagnostics need not agree. Their panel distances decrease, but their direction cosines fall from \(0.66\) to \(0.59\) and from \(0.53\) to \(0.45\).

Across these four settings, the mean direction cosine changes by \(0.088\), \(0.034\), \(0.041\), and \(0.028\). The selected windows show larger changes in level and amplitude than in temporal shape. They also show why the mechanism results in Appendix~\ref{app:mechanism} should not be interpreted as a guarantee for every variable. PV-Surgery can reduce aggregate MSE and pairwise conflict while the direction cosine of an individual variable remains unchanged or becomes worse. These examples add a window-level view of the aggregate results and show that lower error does not always come with a higher direction cosine.

\section{Computational Cost and Training Time}
\label{app:training_time}
PV-Surgery leaves the forecasting architecture unchanged but requires additional computation at each training step. We describe this additional computation and measure its wall-clock cost over the main experimental grid.

\paragraph{Analytical cost.}
Like standard MSE training, PV-Surgery uses one backward pass for each optimizer step. Its additional cost comes from reconstructing and processing the variable-wise gradient rows. Let \(N_{\mathcal{H}}\) denote the number of parameters in the hooked layers and \(N_{\mathcal{S}}\) the number in the selected layers. Dense zero-extended proxy storage and later row-wise processing scale as \(\mathcal{O}(DN_{\mathcal{H}})\). Computing the observed weight outer products costs \(\mathcal{O}(\sum_{\ell\in\mathcal{H}_{\mathrm{run}}}BDM_\ell F_{\mathrm{in},\ell}F_{\mathrm{out},\ell})\), where \(F_{\mathrm{in},\ell}\) and \(F_{\mathrm{out},\ell}\) are the layer widths and the bias term is lower order. Pairwise cosine similarities on the selected subspace cost \(\mathcal{O}(D^2N_{\mathcal{S}})\), and evaluating the unpooled candidate plus a pooled candidate with \(P\) pools costs \(\mathcal{O}((D+P)N_{\mathcal{S}})\).

Layer selection cannot reduce the cache-construction term because the cache is needed to score the layers. It does reduce the pairwise and candidate computations because \(N_{\mathcal{S}}\) is no larger than \(N_{\mathcal{H}}\). Candidate selection adds another direction evaluation when pooling is active. These operations affect training only. The forward pass used at inference is identical to that of the original backbone.

\paragraph{Protocol.}
We pair each PV-Surgery run with the standard MSE run from the same backbone, benchmark, and prediction length. All \(140\) pairs use seed \(42\), the same batch size, and the same GPU type. Early stopping gives different training lengths, so total time alone mixes computational overhead with the number of completed epochs. We divide each run's logged training time by its completed epoch count and form the PV-Surgery to MSE ratio within each pair. The baseline completes \(22.1\) epochs on average and PV-Surgery completes \(23.3\). These timings cover complete training epochs, including data loading, model computation, optimizer steps, and logging. They therefore measure practical training time rather than the isolated cost of the PV-Surgery operator.

\begin{table}[H]
\centering
\caption{Training cost of PV-Surgery against standard MSE training over the \(140\) matched settings
of Table~\ref{tab:backbone_pv}. Epochs and seconds per epoch are means over the \(28\) settings of
each backbone. The ratio columns are computed per setting and then aggregated, so they are not the
quotient of the two preceding columns.}
\label{tab:training_time}
\footnotesize
\renewcommand{\arraystretch}{0.9}
\begin{tabular*}{\linewidth}{@{\extracolsep{\fill}}lrrrrrrc@{}}
\toprule
\multirow{2}{*}{\textbf{Backbone}} & \multicolumn{2}{c}{\textbf{Epochs}}
  & \multicolumn{2}{c}{\textbf{Seconds / epoch}} & \multicolumn{2}{c}{\textbf{Per-epoch ratio}}
  & \multirow{2}{*}{\textbf{Slower}} \\
\cmidrule(lr){2-3}\cmidrule(lr){4-5}\cmidrule(lr){6-7}
& \textbf{MSE} & \textbf{+PV} & \textbf{MSE} & \textbf{+PV} & \textbf{median} & \textbf{mean} & \\
\midrule
DLinear & 27.2 & 23.4 & 6.2 & 5.5 & 0.88 & 1.15 & 6/28 \\
MICN & 17.6 & 17.8 & 46.1 & 62.1 & 1.35 & 1.37 & 28/28 \\
SCINet & 28.5 & 27.7 & 18.7 & 28.5 & 1.52 & 1.51 & 28/28 \\
iTransformer & 15.6 & 25.4 & 6.4 & 15.3 & 2.09 & 2.29 & 28/28 \\
TimeXer & 21.8 & 22.3 & 6.6 & 14.8 & 2.22 & 2.24 & 28/28 \\
\midrule
\textbf{All} & 22.1 & 23.3 & 16.8 & 25.3 & \textbf{1.54} & \textbf{1.71} & \textbf{118/140} \\
\bottomrule
\end{tabular*}
\end{table}

\paragraph{Measured overhead.}
Table~\ref{tab:training_time} reports the paired results. Across all settings, the per-epoch ratio has a median of \(1.54\) and a mean of \(1.71\). PV-Surgery is slower in \(118\) of the \(140\) pairs. Summing the logged time over the grid gives \(12.84\) hours for standard MSE training and \(20.56\) hours for PV-Surgery, a total-time ratio of \(1.60\). On the six benchmarks with prediction lengths \(96\), \(192\), \(336\), and \(720\), the median per-epoch ratio remains between \(1.51\) and \(1.54\). We do not observe a systematic increase with forecast length in this grid.

The overhead varies more across backbones. The median ratio is \(2.09\) for iTransformer and \(2.22\) for TimeXer, both of which expose several cache-compatible linear layers. MICN has a lower median ratio of \(1.35\), although its baseline epoch is much slower in absolute time. DLinear has a median ratio below one at \(0.88\), but its mean ratio is \(1.15\) and six of its twenty-eight settings are slower with PV-Surgery. The DLinear result should therefore be read as low overhead with substantial timing variation, not as evidence that surgery generally accelerates training.

\paragraph{Implementation considerations.}
The measured ratios depend on the hardware and current implementation and should not be treated as architecture-independent constants. Caching a separate variable axis and, on some execution paths, materializing dense \(D\times N_{\mathcal H}\) proxy rows increase memory traffic, while evaluating both unpooled and pooled candidates adds computation. Compact slice-wise assembly and reuse of intermediate quantities could reduce this overhead. Reducing the number of cached layers offers another cost-coverage trade-off. PV-Surgery changes only training-time gradient construction, so inference remains unchanged.

\end{document}